%% file: iclr2027_conference.tex
\documentclass{article} %
\usepackage{iclr2027_conference,times}

\input{math_commands.tex}

\usepackage{hyperref}
\usepackage{url}

\usepackage{xspace}
\usepackage{minitoc}
\usepackage{titletoc}
\usepackage{algorithm}
\usepackage{algpseudocode}
\usepackage{amssymb}
\usepackage{caption}
\usepackage{placeins}
\usepackage{wrapfig}
\usepackage{needspace}
\usepackage{mathtools}
\usepackage{algorithm}
\usepackage{booktabs}
\usepackage{svg}
\usepackage{multirow}
\usepackage{adjustbox}
\usepackage{booktabs}
\usepackage{xcolor}
\usepackage{colortbl}
\usepackage[normalem]{ulem}
\usepackage[bottom]{footmisc}
\usepackage{pifont}
\usepackage{fancyvrb}
\usepackage{tikz}
\usetikzlibrary{arrows.meta}
\newcommand{\cmark}{\ding{51}}
\newcommand{\xmark}{\ding{55}}

\usepackage{xcolor}
\usepackage{listings}

\definecolor{revisiongreen}{rgb}{0,0.31,0.71}
\definecolor{revisionblack}{rgb}{0,0,0}
\providecommand{\ybedit}[1]{{\color{revisionblack}#1}}

\DeclarePairedDelimiter{\norm}{\lVert}{\rVert}

\newcommand{\denoiser}{D_{\theta}}
\newcommand{\corrsym}{C_{\phi}}
\newcommand{\corrector}{\ensuremath{\corrsym}\xspace}

\usepackage{xcolor}
\newcommand{\yb}[1]{\textcolor{black}{#1}}
\definecolor{ybncolor}{HTML}{000000}
\definecolor{dhncolor}{HTML}{000000}

\newcommand{\dho}[1]{{\color{black}#1}}
\newcommand{\efros}[1]{{\color{black}#1}}
\newcommand{\SiTREPA}{\ensuremath{\text{SiT}_{\mathrm{REPA}}}}
\input{generated/results.tex}

\input{generated/unconditional_matched50k_preview.tex}
\input{generated/human_evaluation_macros.tex}
\input{generated/jit_interval_macros.tex}
\title{Correction Training for Diffusion Models}
\iclrfinalcopy

\author{
Donghoon Ahn$^{1}$,
Runqian Wang$^{1}$,
Alexei A. Efros$^{1}$,
Taesung Park$^{2}$,
Yash Belhe$^{2}$ \\
$^{1}$University of California, Berkeley
\qquad
$^{2}$Reve
}

\begin{document}
\definecolor{heatgreen}{RGB}{46,150,90}

\maketitle
\fancyhead[L]{}
\AddToShipoutPictureFG*{%
  \AtPageLowerLeft{%
    \put(\LenToUnit{1.5in},\LenToUnit{0.55in}){%
      \makebox(0,0)[lb]{\footnotesize\textbf{Project page:} \url{https://sunovivid.github.io/correction-training/}}%
    }%
  }%
}
\input{generated/teaser_curated}

\begin{abstract}
\efros{Iterative sampling in diffusion models during inference can lead to error accumulation due to exposure bias. The denoiser is trained on real training images interpolated with noise, but at inference its inputs are} \dho{constructed from} \efros{its own imperfect predictions from the earlier steps, something it has never seen at training time. This might be one reason diffusion models often struggle without additional help, such as CFG.} 
In this paper, we introduce a correction step for diffusion models which attempts to reduce this training-inference gap. The corrector takes the base model’s own prediction as input and is trained against the same ground-truth target as the base model. This makes correction an explicit learning task, with training inputs
provided by the base model. During sampling, at each denoising step, we iteratively refine
the denoiser’s clean-sample prediction. This allows us to correct predictions at intermediate denoising steps, preventing early artifacts from persisting into the final sample. We find that applying correction beyond the single step supervised during training is key to the significant quality gains. This simple framework leads to substantial improvement in sample quality while largely preserving diversity. We extensively evaluate our method across widely used architectures, including DiT, SiT, and JiT. Our method substantially improves upon previously reported FID scores for \textit{unguided generation} and  \textit{unconditional generation}. Our method achieves performance competitive with classifier-free guidance (CFG), and combining correction with CFG further improves diversity and reduces oversaturation typical of CFG. Finally, we show that the additional inference cost of correction can be amortized by distilling the refined predictions back into the base diffusion model, retaining the quality gains without additional inference-time overhead. 

\end{abstract}

\input{sections/intro}

\input{sections/method}

\input{sections/results}

\input{sections/discussion_new}

\input{sections/related_works}

\subsubsection*{Acknowledgments}
We would like to thank Konpat Preechakul for valuable insights and feedback, as well as Michael Gharbi for insightful discussions, and Nicolas Dufour for providing detailed feedback on the manuscript. We are grateful to A. Sophia Koepke, Anne Harrington, Amil Dravid, Jorge Diaz Chao, Isabella Yu, Yutong Bai, Yong-Hyun Park, Sangyoon Lee, Sihyun Yu, Susung Hong, Joonghyuk Shin, Gyuseong Lee, and Sanghyun Lee for helpful discussions. This work was supported, in part, by NSF IIS-2403305 and ONR MURI grants.

\input{sections/appendix}

\end{document}

%% file: math_commands.tex
\usepackage{amsmath,amsfonts,bm}

\def\eqref#1{equation~\ref{#1}}

\def\1{\bm{1}}

\DeclareMathAlphabet{\mathsfit}{\encodingdefault}{\sfdefault}{m}{sl}
\SetMathAlphabet{\mathsfit}{bold}{\encodingdefault}{\sfdefault}{bx}{n}

%% file: generated/results.tex
\providecommand{\UncondBaseSamplerSteps}{250}
\providecommand{\UncondCorrectionOneSamplerSteps}{125}
\providecommand{\UncondCorrectionBestSamplerSteps}{63}
\providecommand{\UncondAGBestSamplerSteps}{125}

\providecommand{\RepaMatchedBaseFID}{5.53}
\providecommand{\RepaMatchedBasePrecision}{0.708}
\providecommand{\RepaMatchedBaseRecall}{0.739}
\providecommand{\RepaMatchedCorrectionFID}{1.31}
\providecommand{\RepaMatchedCorrectionPrecision}{0.797}
\providecommand{\RepaMatchedCorrectionRecall}{0.703}
\providecommand{\RepaCfgFID}{1.35}
\providecommand{\RepaCfgPrecision}{0.803}
\providecommand{\RepaCfgRecall}{0.699}

\providecommand{\JitLBasePrecision}{0.607}
\providecommand{\JitLBaseRecall}{0.672}
\providecommand{\JitLCfgPrecision}{0.780}
\providecommand{\JitLCfgRecall}{0.606}
\providecommand{\JitLIntervalPrecision}{0.766}
\providecommand{\JitLIntervalRecall}{0.601}
\providecommand{\JitLIntervalFID}{2.96}
\providecommand{\JitLBaseFID}{14.86}
\providecommand{\JitLCfgFID}{2.35}

\providecommand{\DitFineTuneControlFID}{9.58}
\providecommand{\DitCorrectionBestFID}{2.07}
\providecommand{\DitCorrectionBestPrecision}{0.811}
\providecommand{\DitCorrectionBestRecall}{0.603}

\providecommand{\DitBasePrecision}{0.668}
\providecommand{\DitBaseRecall}{0.676}

\providecommand{\SitBaseFID}{8.55}

\providecommand{\SitBasePrecision}{0.687}
\providecommand{\SitBaseRecall}{0.669}

\providecommand{\SitCorrectionBestFID}{1.92}

\providecommand{\SitCorrectionBestPrecision}{0.812}
\providecommand{\SitCorrectionBestRecall}{0.608}

\providecommand{\UncondBaseNFE}{250}
\providecommand{\UncondBaseFID}{24.26}
\providecommand{\UncondBasePrecision}{0.550}
\providecommand{\UncondBaseRecall}{0.711}
\providecommand{\UncondCorrectionOneNFE}{250}
\providecommand{\UncondCorrectionOneFID}{18.29}
\providecommand{\UncondCorrectionOnePrecision}{0.601}
\providecommand{\UncondCorrectionOneRecall}{0.721}
\providecommand{\UncondCorrectionBestNFE}{252}
\providecommand{\UncondCorrectionBestFID}{5.09}
\providecommand{\UncondCorrectionBestPrecision}{0.785}
\providecommand{\UncondCorrectionBestRecall}{0.652}

\providecommand{\UncondBaseStep}{100K}
\providecommand{\UncondCorrectionOneStep}{100K}
\providecommand{\UncondCorrectionBestStep}{100K}

\providecommand{\DitCfgMatchedFID}{2.51}

\providecommand{\DitCfgMatchedPrecision}{0.818}
\providecommand{\DitCfgMatchedRecall}{0.581}

\providecommand{\SitCfgMatchedFID}{2.11}

\providecommand{\SitCfgMatchedPrecision}{0.820}
\providecommand{\SitCfgMatchedRecall}{0.595}

\providecommand{\DistillTableBaselineFID}{8.62}
\providecommand{\DistillKTwoStudentMatchedFID}{2.04}
\providecommand{\DistillKTwoStudentMatchedSteps}{250}
\providecommand{\DistillKTwoStudentMatchedNFE}{250}
\providecommand{\DistillKTwoTeacherFID}{1.92}
\providecommand{\DistillTeacherK}{2}
\providecommand{\DistillKTwoStudentFID}{2.26}
\providecommand{\DistillKTwoStudentTrainingSteps}{100K}
\providecommand{\DistillTableBaselineSteps}{250}
\providecommand{\DistillTableBaselineNFE}{250}
\providecommand{\DistillKTwoTeacherSteps}{83}
\providecommand{\DistillKTwoTeacherNFE}{249}
\providecommand{\DistillKTwoStudentSteps}{83}
\providecommand{\DistillKTwoStudentNFE}{83}

\providecommand{\UncondAGBestFID}{8.28}
\providecommand{\UncondAGBestPrecision}{0.700}
\providecommand{\UncondAGBestRecall}{0.679}
\providecommand{\UncondAGBestNFE}{250}
\providecommand{\UncondAGBestW}{6}

%% file: generated/unconditional_matched50k_preview.tex
\providecommand{\UncondBaseFID}{}\renewcommand{\UncondBaseFID}{24.82}
\providecommand{\UncondBasePrecision}{}\renewcommand{\UncondBasePrecision}{0.571}
\providecommand{\UncondBaseRecall}{}\renewcommand{\UncondBaseRecall}{0.699}
\providecommand{\UncondBaseStep}{}\renewcommand{\UncondBaseStep}{50K}
\providecommand{\UncondBaseNFE}{}\renewcommand{\UncondBaseNFE}{250}
\providecommand{\UncondBaseSamplerSteps}{}\renewcommand{\UncondBaseSamplerSteps}{250}
\providecommand{\UncondCorrectionOneFID}{}\renewcommand{\UncondCorrectionOneFID}{17.80}
\providecommand{\UncondCorrectionOnePrecision}{}\renewcommand{\UncondCorrectionOnePrecision}{0.620}
\providecommand{\UncondCorrectionOneRecall}{}\renewcommand{\UncondCorrectionOneRecall}{0.679}
\providecommand{\UncondCorrectionOneStep}{}\renewcommand{\UncondCorrectionOneStep}{50K}
\providecommand{\UncondCorrectionOneNFE}{}\renewcommand{\UncondCorrectionOneNFE}{250}
\providecommand{\UncondCorrectionOneSamplerSteps}{}\renewcommand{\UncondCorrectionOneSamplerSteps}{125}
\providecommand{\UncondCorrectionBestFID}{}\renewcommand{\UncondCorrectionBestFID}{5.37}
\providecommand{\UncondCorrectionBestPrecision}{}\renewcommand{\UncondCorrectionBestPrecision}{0.782}
\providecommand{\UncondCorrectionBestRecall}{}\renewcommand{\UncondCorrectionBestRecall}{0.595}
\providecommand{\UncondCorrectionBestStep}{}\renewcommand{\UncondCorrectionBestStep}{50K}
\providecommand{\UncondCorrectionBestNFE}{}\renewcommand{\UncondCorrectionBestNFE}{250}
\providecommand{\UncondCorrectionBestSamplerSteps}{}\renewcommand{\UncondCorrectionBestSamplerSteps}{50}
\providecommand{\UncondCorrectionBestK}{}\renewcommand{\UncondCorrectionBestK}{4}
\providecommand{\UncondAGBestFID}{}\renewcommand{\UncondAGBestFID}{9.37}
\providecommand{\UncondAGBestPrecision}{}\renewcommand{\UncondAGBestPrecision}{0.699}
\providecommand{\UncondAGBestRecall}{}\renewcommand{\UncondAGBestRecall}{0.630}
\providecommand{\UncondAGBestNFE}{}\renewcommand{\UncondAGBestNFE}{250}
\providecommand{\UncondAGBestSamplerSteps}{}\renewcommand{\UncondAGBestSamplerSteps}{125}
\providecommand{\UncondAGBestW}{}\renewcommand{\UncondAGBestW}{7}

%% file: generated/human_evaluation_macros.tex
\providecommand{\HumanStudyPromptCount}{80}
\providecommand{\HumanStudyVotesPerPrompt}{10}
\providecommand{\HumanStudyVotesPerMetric}{800}

\providecommand{\HumanStudyBaselineCFG}{6}
\providecommand{\HumanStudyCFG}{4}
\providecommand{\HumanStudyK}{8}

\providecommand{\HumanStudyQualityInstruction}{Which image has better visual quality?}
\providecommand{\HumanStudyDiversityInstruction}{Which group is more diverse? Ignore image quality.}

%% file: generated/teaser_curated.tex
\begin{figure}[H]
\centering
\vspace{-20pt}
\includegraphics[width=\textwidth]{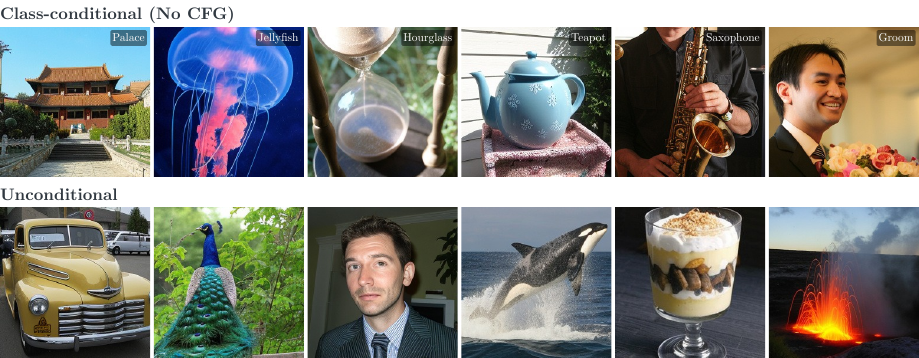}
\vspace{-15pt}
\caption{\textbf{ImageNet samples with correction.}{ \textit{Without the use of guidance}, our correction method can achieve high-quality  class-conditional (top) and \textit{unconditional} (bottom) image samples.}}
\label{fig:teaser}
\end{figure}

%% file: sections/intro.tex
\input{figures/method_figure}

\section{Introduction}

\dho{Diffusion models generate images through a sequence of denoising predictions~\citep{ho2020denoising,song2020score}. At each sampling step, a neural network estimates clean data from a noisy input, and this estimate is used to update the sample. But one curious property of diffusion models is the training-inference gap arising from what is known as exposure bias~\citep{ning2023input, ning2024elucidating, li2024alleviating, yu2025frequency}. The denoiser is trained on real training images interpolated with noise. At inference time, however, it receives noisy inputs that depend on \textit{its own imperfect predictions} from earlier steps. These inputs can differ from the noisy training inputs, creating a mismatch between training and inference. As noise decreases along the denoising trajectory, prediction errors can accumulate, often producing visual artifacts and incoherent structures in the generated images. A common way to improve sample quality is classifier-free guidance (CFG)~\citep{ho2022classifier}, which generates better quality samples, but at the cost of reduced diversity and oversaturation.}

In this paper, we take a more direct approach: correcting the \textit{model's own prediction}. We introduce a \textit{corrector} that takes the base model's prediction as input and is trained against the same ground-truth target, as illustrated in Fig.~\ref{fig:correction}. This makes correction an explicit learning task, with training inputs provided by the base model. During sampling, at each denoising step, we therefore iteratively correct the denoiser's clean-sample prediction. Although the corrector is trained only on predictions from the base denoiser, we find that it can be applied repeatedly to its own outputs at inference time without additional training. 
This iterative correction further improves sample quality, producing sharper images with more coherent structures. \dho{We hypothesize that repeated correction helps compensate for errors accumulated over earlier denoising steps, reducing the denoiser's training--inference gap and contributing to the observed improvement (see Sec.~\ref{sec:how-iterative-correction-changes} for details).}

We evaluate our method across class-conditional, unconditional, and text-to-image generation, using established models including DiT~\citep{peebles2023scalable}, SiT~\citep{ma2024sit}, REPA~\citep{yu2024representation}, and JiT~\citep{li2026back}. Correction improves unguided sample quality across these settings with a relatively small amount of additional training. The gain remains at matched numbers of function evaluations (NFEs), suggesting that introducing correction steps can make more effective use of inference compute than denoising alone.
In unconditional generation, correction also provides substantial improvements without any guidance signal. Comparisons with CFG on DiT and SiT show better quality at matched diversity, as measured by FID~\citep{heusel2017gans} and recall~\citep{kynkaanniemi2019improved}. Qualitatively, correction shows less oversaturation and greater visual realism than CFG. Moreover, adding correction to CFG reduces the stereotypical ``AI-generated'' look of its samples. We present conditional and unconditional samples using correction in Fig.~\ref{fig:teaser}.

Finally, we show that the corrector can be made as efficient as standard diffusion models while retaining its quality advantage. We distill corrected predictions back into the base diffusion model. The student learns to match predictions obtained through repeated correction and requires only one forward pass per sampling step. The resulting model retains much of the quality improvement while preserving the base model's architecture, parameter count, and inference cost. This allows the benefits of correction to be incorporated into the base model without additional correction passes during generation. We hope that the effectiveness, simplicity, and efficiency of corrector can help to bridge the long-standing gap between unconditional and conditional generation.

%% file: figures/method_figure.tex
\begin{figure}[!tbp]
\vspace{-20pt}
    \centering
\includegraphics[width=\linewidth]{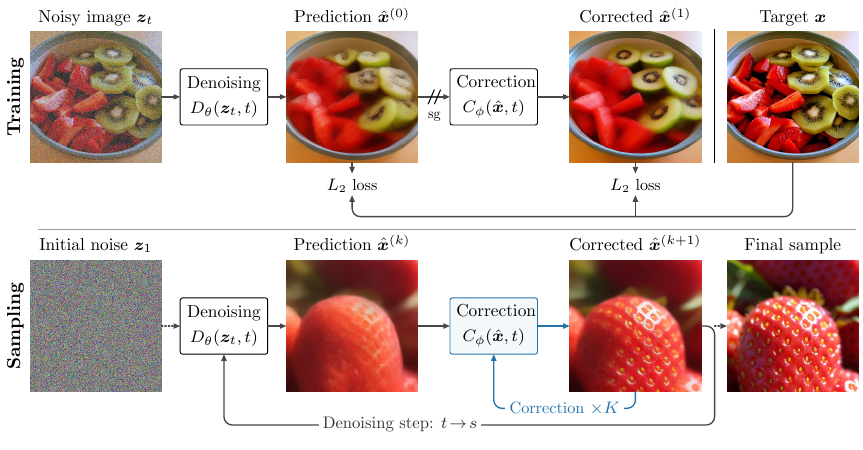}
\vspace{-15pt}
    \caption{\textbf{Denoising and prediction correction.} During training, the denoiser and corrector predict the same clean target. Stop-gradient blocks the correction loss through the denoiser's prediction. During sampling, each denoising prediction is refined $K$ times before the next sampling step.}
    \vspace{-15pt}

    \label{fig:correction}
\end{figure}

%% file: sections/method.tex
\section{\ybedit{Learning to Correct Denoising Predictions}}
\subsection{\ybedit{Background: Denoising and Sampling}}
\ybedit{Diffusion models learn to recover data from noisy observations~\citep{sohl2015deep,song2019generative,ho2020denoising,song2020score}. Flow matching~\citep{lipman2022flow,liu2022flow,albergo2025stochastic} extends the diffusion paradigm by using a linear interpolation between noise and image. Specifically, given a clean sample ${\bm{x}}\sim p_{\mathrm{data}}$ and independent noise $\bm{\epsilon}\sim\mathcal{N}(0,I)$, the training input is}
\begin{equation}
\color[rgb]{0,0,0} %
{\bm{z}_t}=(1-t){\bm{x}}+t\bm{\epsilon},\qquad t\in[0,1],
\label{eq:forward-interpolation}
\end{equation}
\ybedit{where $t=0$ denotes data and $t=1$ denotes noise. An {$\bm{x}$-prediction} network $\denoiser$ estimates the clean sample from $({\bm{z}_t},t)$. Its denoising objective is}
\begin{equation}
\color{revisionblack}
\mathcal{L}_{\mathrm{denoise}}
=\mathbb{E}_{{\bm{x}},\bm{\epsilon},t}\!\left[\norm{\denoiser({\bm{z}_t},t)-{\bm{x}}}_2^2\right].
\label{eq:denoising-loss}
\end{equation}
At each sampling step, the denoiser predicts the clean image, and the sampler uses this prediction to construct the next noisy state. For example, an Euler step from $t$ to $s<t$ gives
\begin{equation}
\bm{z}_s=\frac{s}{t}\bm{z}_t+
\left(1-\frac{s}{t}\right)\hat{\bm{x}}^{(0)},
\qquad
\hat{\bm{x}}^{(0)}=\denoiser(\bm{z}_t,t).
\end{equation}
The updated state \(\bm{z}_s\) then becomes the denoiser’s next input. During training, however, inputs are constructed by interpolating real images with independent noise (Eq.~\ref{eq:forward-interpolation}). Sampling inputs therefore depend on the model’s own earlier predictions and may differ from the inputs encountered during training, creating a training–inference gap~\citep{ning2023input,ning2024elucidating,li2024alleviating}.

\input{algorithm/training-sampling}

\subsection{Prediction Error During Sampling}
\begin{figure}[tbp]
    \centering
    \vspace{-15pt}
    \includegraphics[width=1.0\linewidth]{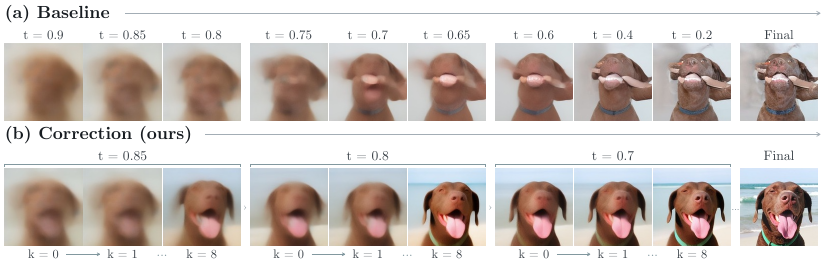}
    \vspace{-20pt}
    \caption{\textbf{Prediction correction improves structural coherence during sampling.}
    We show clean-sample predictions $\hat{\bm{x}}$ along denoising trajectories without (a) and with (b) correction. In (a), early structural errors persist as later denoising steps add details. In (b), iterative correction revises the prediction at each $t$, producing more coherent object structures before the next denoising step. Each triplet shows this refinement within one denoising step. Additional visualizations are shown in Appendix~\ref{app:correction-visualizations} (Fig.~\ref{fig:correction-full-dog}).}
    \label{fig:iterative-correction}
\end{figure}

Fig.~\ref{fig:iterative-correction} (a) illustrates how structural errors can persist along the sampling trajectory. Early predictions contain incoherent object structures. Later denoising steps add texture and facial details such as teeth, eyes, and a tongue while largely preserving the incorrect object structure. The clean targets used for training, however, are real images with coherent object structures. The denoiser learns to recover these structures from noise, but is not explicitly trained to further refine its own predictions at the same noise level.

One widely used way to obtain sharper, more coherent samples is classifier-free guidance (CFG)~\citep{ho2022classifier}. CFG strengthens conditioning by amplifying the difference between conditional and unconditional predictions. However, it can reduce sample diversity and produce oversaturated images.

In this paper, we make prediction correction an explicit learning task: we train a corrector $\corrector$ that takes the denoiser's own prediction as input. During sampling, we apply this learned mapping repeatedly to improve sample quality. As we show in Sec.~\ref{sec:discussion}, repeated correction can improve generation even when it increases reconstruction error on training inputs. We investigate this behavior and its relationship to the training–inference gap.

\subsection{Learning to Correct Model Predictions}
Standard denoising training learns the mapping $\bm{z}_t\rightarrow\bm{x}$ by reconstructing images corrupted by a prescribed noise process. Given a noisy observation $\bm{z}_t$, the denoiser produces an initial prediction $\hat{\bm{x}}^{(0)}$, which may still contain inaccuracies. We introduce \emph{prediction correction} as a reconstruction task that uses this prediction as its input. The corrector takes $\hat{\bm{x}}^{(0)}$ and predicts the same clean target $\bm{x}$, learning the mapping $\hat{\bm{x}}^{(0)}\rightarrow\bm{x}$ (Fig.~\ref{fig:correction}).

When $\hat{\bm{x}}^{(0)}$ already captures much of the structure in $\bm{x}$, the initial denoising pass has already performed much of the reconstruction. The corrector can then build on this estimate and focus on the remaining inaccuracies.
{We refine $\hat{\bm{x}}^{(0)}$ by applying corrector $\corrector$ again at the original time $t$:
$
\hat{\bm{x}}^{(1)}
=\corrector(\hat{\bm{x}}^{(0)},t).
$
The corrected prediction directly estimates the clean sample $\bm{x}$. We train it with
\begin{equation}
\mathcal{L}_{\mathrm{corr}}
=\mathbb{E}_{\bm{x},\bm{\epsilon},t}\!\left[
  \norm{\corrector(\operatorname{sg}(\hat{\bm{x}}^{(0)}),t)-\bm{x}}_2^2
\right],
\label{eq:correction-loss}
\end{equation}
where $\operatorname{sg}$ is the stop-gradient operation.}

{Starting from a pretrained denoiser, we jointly optimize
$\mathcal{L}_{\mathrm{total}}
=\mathcal{L}_{\mathrm{denoise}}
+\lambda_{\mathrm{corr}}\mathcal{L}_{\mathrm{corr}},$
where $\lambda_{\mathrm{corr}}$ weights the correction task. The denoising term supervises the initial prediction, while the correction term supervises its refinement. The two tasks are trained jointly, with corrector inputs generated online by the current model. Algorithm~\ref{alg:training} summarizes the loss computation in Python-style pseudocode.}

\subsection{{Iterative Correction at Inference}}

Training supervises only a single correction step, from the initial denoising prediction $\hat{\bm{x}}^{(0)}$ to the refined prediction $\hat{\bm{x}}^{(1)}$.
At inference, however, the model can repeatedly refine its own outputs, further improving sample quality. The output $\hat{\bm{x}}^{(k)}$ at the $k$-th correction iteration is another estimate of the same clean image. This estimate may still contain inaccuracies, motivating further applications of the learned correction.

{Starting from the denoising prediction, we apply $K$ correction passes:
\begin{equation}
\begin{aligned}
\hat{\bm{x}}^{(0)}
  &=\denoiser(\bm{z}_t,t),\\
\hat{\bm{x}}^{(k+1)}
  &=\corrector(\hat{\bm{x}}^{(k)},t),
  \qquad k=0,\ldots,K-1.
\end{aligned}
\label{eq:iterative-correction}
\end{equation}}
Fig.~\ref{fig:iterative-correction} illustrates how successive corrections refine intermediate predictions throughout the sampling trajectory. This provides a way to spend additional inference compute on improving the predictions at each time. At matched total inference cost, we find this more effective than simply increasing the number of denoising steps.

%% file: algorithm/training-sampling.tex
\begin{figure}[!t]
\vspace{-15pt}

\definecolor{codekeyword}{HTML}{8959A8}
\definecolor{codefunction}{HTML}{4271AE}
\definecolor{codecomment}{HTML}{718096}
\definecolor{codemodel}{HTML}{A63D40}

\lstdefinestyle{correctionpython}{
    language=Python,
    basicstyle=\ttfamily\small\color{black},
    keywordstyle=\color{codekeyword}\bfseries,
    commentstyle=\color{codecomment}\itshape,
    stringstyle=\color{codefunction},
    morekeywords=[2]{sample_time,randn_like,randn,stopgrad,mse,zip,range},
    keywordstyle=[2]\color{codefunction},
    emph={D,C},
    emphstyle=\color{codemodel}\bfseries,
    columns=fullflexible,
    keepspaces=true,
    showstringspaces=false,
    breaklines=true,
    tabsize=4,
    aboveskip=0.25em,
    belowskip=0.25em,
    lineskip=1pt
}

\newcommand{\PaperCode}[2]{#2}
\newcommand{\PaperCodeZsh}{}

\makeatletter
\def\PaperCode@reset{\let\PaperCode@it=\relax \let\PaperCode@bf=\relax%
    \let\PaperCode@ul=\relax \let\PaperCode@tc=\relax%
    \let\PaperCode@bc=\relax \let\PaperCode@ff=\relax}
\def\PaperCode@tok#1{\csname PaperCode@tok@#1\endcsname}
\def\PaperCode@toks#1+{\ifx\relax#1\empty\else%
    \PaperCode@tok{#1}\expandafter\PaperCode@toks\fi}
\def\PaperCode@do#1{\PaperCode@bc{\PaperCode@tc{\PaperCode@ul{%
    \PaperCode@it{\PaperCode@bf{\PaperCode@ff{#1}}}}}}}
\def\PaperCode#1#2{\PaperCode@reset\PaperCode@toks#1+\relax+\PaperCode@do{#2}}

\@namedef{PaperCode@tok@}{\def\PaperCode@tc##1{\textcolor[rgb]{0.00,0.00,0.00}{##1}}}
\@namedef{PaperCode@tok@c}{\def\PaperCode@tc##1{\textcolor[rgb]{0.00,0.47,0.50}{##1}}}
\@namedef{PaperCode@tok@k}{\let\PaperCode@bf=\textbf\def\PaperCode@tc##1{\textcolor[rgb]{0.82,0.00,0.31}{##1}}}
\@namedef{PaperCode@tok@n}{\def\PaperCode@tc##1{\textcolor[rgb]{0.00,0.00,0.00}{##1}}}
\@namedef{PaperCode@tok@nb}{\let\PaperCode@bf=\textbf\def\PaperCode@tc##1{\textcolor[rgb]{0.82,0.00,0.31}{##1}}}
\@namedef{PaperCode@tok@m}{\def\PaperCode@tc##1{\textcolor[rgb]{0.00,0.31,0.82}{##1}}}
\@namedef{PaperCode@tok@s}{\def\PaperCode@tc##1{\textcolor[rgb]{0.70,0.30,0.00}{##1}}}
\@namedef{PaperCode@tok@o}{\def\PaperCode@tc##1{\textcolor[rgb]{0.00,0.00,0.00}{##1}}}
\@namedef{PaperCode@tok@p}{\def\PaperCode@tc##1{\textcolor[rgb]{0.00,0.00,0.00}{##1}}}
\@namedef{PaperCode@tok@w}{\def\PaperCode@tc##1{\textcolor[rgb]{0.00,0.00,0.00}{##1}}}
\@namedef{PaperCode@tok@kc}{\let\PaperCode@bf=\textbf\def\PaperCode@tc##1{\textcolor[rgb]{0.82,0.00,0.31}{##1}}}
\@namedef{PaperCode@tok@kd}{\let\PaperCode@bf=\textbf\def\PaperCode@tc##1{\textcolor[rgb]{0.82,0.00,0.31}{##1}}}
\@namedef{PaperCode@tok@kn}{\let\PaperCode@bf=\textbf\def\PaperCode@tc##1{\textcolor[rgb]{0.82,0.00,0.31}{##1}}}
\@namedef{PaperCode@tok@kp}{\let\PaperCode@bf=\textbf\def\PaperCode@tc##1{\textcolor[rgb]{0.82,0.00,0.31}{##1}}}
\@namedef{PaperCode@tok@kr}{\let\PaperCode@bf=\textbf\def\PaperCode@tc##1{\textcolor[rgb]{0.82,0.00,0.31}{##1}}}
\@namedef{PaperCode@tok@kt}{\let\PaperCode@bf=\textbf\def\PaperCode@tc##1{\textcolor[rgb]{0.82,0.00,0.31}{##1}}}
\@namedef{PaperCode@tok@na}{\def\PaperCode@tc##1{\textcolor[rgb]{0.00,0.00,0.00}{##1}}}
\@namedef{PaperCode@tok@bp}{\let\PaperCode@bf=\textbf\def\PaperCode@tc##1{\textcolor[rgb]{0.82,0.00,0.31}{##1}}}
\@namedef{PaperCode@tok@nc}{\def\PaperCode@tc##1{\textcolor[rgb]{0.00,0.00,0.00}{##1}}}
\@namedef{PaperCode@tok@no}{\def\PaperCode@tc##1{\textcolor[rgb]{0.00,0.00,0.00}{##1}}}
\@namedef{PaperCode@tok@nd}{\def\PaperCode@tc##1{\textcolor[rgb]{0.00,0.00,0.00}{##1}}}
\@namedef{PaperCode@tok@ni}{\def\PaperCode@tc##1{\textcolor[rgb]{0.00,0.00,0.00}{##1}}}
\@namedef{PaperCode@tok@ne}{\def\PaperCode@tc##1{\textcolor[rgb]{0.00,0.00,0.00}{##1}}}
\@namedef{PaperCode@tok@nf}{\def\PaperCode@tc##1{\textcolor[rgb]{0.00,0.00,0.00}{##1}}}
\@namedef{PaperCode@tok@fm}{\def\PaperCode@tc##1{\textcolor[rgb]{0.00,0.00,0.00}{##1}}}
\@namedef{PaperCode@tok@py}{\def\PaperCode@tc##1{\textcolor[rgb]{0.00,0.00,0.00}{##1}}}
\@namedef{PaperCode@tok@nl}{\def\PaperCode@tc##1{\textcolor[rgb]{0.00,0.00,0.00}{##1}}}
\@namedef{PaperCode@tok@nn}{\def\PaperCode@tc##1{\textcolor[rgb]{0.00,0.00,0.00}{##1}}}
\@namedef{PaperCode@tok@nx}{\def\PaperCode@tc##1{\textcolor[rgb]{0.00,0.00,0.00}{##1}}}
\@namedef{PaperCode@tok@nt}{\def\PaperCode@tc##1{\textcolor[rgb]{0.00,0.00,0.00}{##1}}}
\@namedef{PaperCode@tok@nv}{\def\PaperCode@tc##1{\textcolor[rgb]{0.00,0.00,0.00}{##1}}}
\@namedef{PaperCode@tok@vc}{\def\PaperCode@tc##1{\textcolor[rgb]{0.00,0.00,0.00}{##1}}}
\@namedef{PaperCode@tok@vg}{\def\PaperCode@tc##1{\textcolor[rgb]{0.00,0.00,0.00}{##1}}}
\@namedef{PaperCode@tok@vi}{\def\PaperCode@tc##1{\textcolor[rgb]{0.00,0.00,0.00}{##1}}}
\@namedef{PaperCode@tok@vm}{\def\PaperCode@tc##1{\textcolor[rgb]{0.00,0.00,0.00}{##1}}}
\@namedef{PaperCode@tok@sa}{\def\PaperCode@tc##1{\textcolor[rgb]{0.70,0.30,0.00}{##1}}}
\@namedef{PaperCode@tok@sb}{\def\PaperCode@tc##1{\textcolor[rgb]{0.70,0.30,0.00}{##1}}}
\@namedef{PaperCode@tok@sc}{\def\PaperCode@tc##1{\textcolor[rgb]{0.70,0.30,0.00}{##1}}}
\@namedef{PaperCode@tok@dl}{\def\PaperCode@tc##1{\textcolor[rgb]{0.70,0.30,0.00}{##1}}}
\@namedef{PaperCode@tok@sd}{\def\PaperCode@tc##1{\textcolor[rgb]{0.70,0.30,0.00}{##1}}}
\@namedef{PaperCode@tok@s2}{\def\PaperCode@tc##1{\textcolor[rgb]{0.70,0.30,0.00}{##1}}}
\@namedef{PaperCode@tok@se}{\def\PaperCode@tc##1{\textcolor[rgb]{0.70,0.30,0.00}{##1}}}
\@namedef{PaperCode@tok@sh}{\def\PaperCode@tc##1{\textcolor[rgb]{0.70,0.30,0.00}{##1}}}
\@namedef{PaperCode@tok@si}{\def\PaperCode@tc##1{\textcolor[rgb]{0.70,0.30,0.00}{##1}}}
\@namedef{PaperCode@tok@sx}{\def\PaperCode@tc##1{\textcolor[rgb]{0.70,0.30,0.00}{##1}}}
\@namedef{PaperCode@tok@sr}{\def\PaperCode@tc##1{\textcolor[rgb]{0.70,0.30,0.00}{##1}}}
\@namedef{PaperCode@tok@s1}{\def\PaperCode@tc##1{\textcolor[rgb]{0.70,0.30,0.00}{##1}}}
\@namedef{PaperCode@tok@ss}{\def\PaperCode@tc##1{\textcolor[rgb]{0.70,0.30,0.00}{##1}}}
\@namedef{PaperCode@tok@mb}{\def\PaperCode@tc##1{\textcolor[rgb]{0.00,0.31,0.82}{##1}}}
\@namedef{PaperCode@tok@mf}{\def\PaperCode@tc##1{\textcolor[rgb]{0.00,0.31,0.82}{##1}}}
\@namedef{PaperCode@tok@mh}{\def\PaperCode@tc##1{\textcolor[rgb]{0.00,0.31,0.82}{##1}}}
\@namedef{PaperCode@tok@mi}{\def\PaperCode@tc##1{\textcolor[rgb]{0.00,0.31,0.82}{##1}}}
\@namedef{PaperCode@tok@il}{\def\PaperCode@tc##1{\textcolor[rgb]{0.00,0.31,0.82}{##1}}}
\@namedef{PaperCode@tok@mo}{\def\PaperCode@tc##1{\textcolor[rgb]{0.00,0.31,0.82}{##1}}}
\@namedef{PaperCode@tok@ow}{\def\PaperCode@tc##1{\textcolor[rgb]{0.00,0.00,0.00}{##1}}}
\@namedef{PaperCode@tok@pm}{\def\PaperCode@tc##1{\textcolor[rgb]{0.00,0.00,0.00}{##1}}}
\@namedef{PaperCode@tok@ch}{\def\PaperCode@tc##1{\textcolor[rgb]{0.00,0.47,0.50}{##1}}}
\@namedef{PaperCode@tok@cm}{\def\PaperCode@tc##1{\textcolor[rgb]{0.00,0.47,0.50}{##1}}}
\@namedef{PaperCode@tok@cp}{\def\PaperCode@tc##1{\textcolor[rgb]{0.00,0.47,0.50}{##1}}}
\@namedef{PaperCode@tok@cpf}{\def\PaperCode@tc##1{\textcolor[rgb]{0.00,0.47,0.50}{##1}}}
\@namedef{PaperCode@tok@c1}{\def\PaperCode@tc##1{\textcolor[rgb]{0.00,0.47,0.50}{##1}}}
\@namedef{PaperCode@tok@cs}{\def\PaperCode@tc##1{\textcolor[rgb]{0.00,0.47,0.50}{##1}}}

\def\PaperCodeZbs{\char`\\}
\def\PaperCodeZus{\char`\_}
\def\PaperCodeZob{\char`\{}
\def\PaperCodeZcb{\char`\}}
\def\PaperCodeZca{\char`\^}
\def\PaperCodeZam{\char`\&}
\def\PaperCodeZlt{\char`\<}
\def\PaperCodeZgt{\char`\>}
\def\PaperCodeZsh{\char`\#}
\def\PaperCodeZpc{\char`\%}
\def\PaperCodeZdl{\char`\$}
\def\PaperCodeZhy{\char`\-}
\def\PaperCodeZsq{\char`\'}
\def\PaperCodeZdq{\char`\"}
\def\PaperCodeZti{\char`\~}
\def\PaperCodeZat{@}
\def\PaperCodeZlb{[}
\def\PaperCodeZrb{]}
\makeatother

\centering
\captionsetup{hypcap=false}
\setlength{\intextsep}{0pt}
\begin{minipage}[t]{0.485\textwidth}
\begin{algorithm}[H]
\caption{Training loss}
\label{alg:training}
\begin{Verbatim}[commandchars=\\\{\},fontsize=\small,baselinestretch=1.0,formatcom=\color{black}]
\PaperCode{c+c1}{\PaperCodeZsh{} D: denoiser; C: corrector}
\PaperCode{c+c1}{\PaperCodeZsh{} weight: correction loss weight}
\PaperCode{n}{t} \PaperCode{o}{=} \PaperCode{n+nb}{sample\PaperCodeZus{}time}\PaperCode{p}{(}\PaperCode{n}{x}\PaperCode{p}{)}  \PaperCode{c+c1}{\PaperCodeZsh{} per image}
\PaperCode{n}{e} \PaperCode{o}{=} \PaperCode{n+nb}{randn\PaperCodeZus{}like}\PaperCode{p}{(}\PaperCode{n}{x}\PaperCode{p}{)}
\PaperCode{n}{z} \PaperCode{o}{=} \PaperCode{p}{(}\PaperCode{l+m+mi}{1} \PaperCode{o}{\PaperCodeZhy{}} \PaperCode{n}{t}\PaperCode{p}{)} \PaperCode{o}{*} \PaperCode{n}{x} \PaperCode{o}{+} \PaperCode{n}{t} \PaperCode{o}{*} \PaperCode{n}{e}

\PaperCode{n}{pred} \PaperCode{o}{=} \PaperCode{n}{D}\PaperCode{p}{(}\PaperCode{n}{z}\PaperCode{p}{,} \PaperCode{n}{t}\PaperCode{p}{)}
\PaperCode{n}{refined} \PaperCode{o}{=} \PaperCode{n}{C}\PaperCode{p}{(}\PaperCode{n+nb}{stopgrad}\PaperCode{p}{(}\PaperCode{n}{pred}\PaperCode{p}{)}\PaperCode{p}{,} \PaperCode{n}{t}\PaperCode{p}{)}

\PaperCode{n}{loss} \PaperCode{o}{=} \PaperCode{n+nb}{mse}\PaperCode{p}{(}\PaperCode{n}{pred}\PaperCode{p}{,} \PaperCode{n}{x}\PaperCode{p}{)}
\PaperCode{n}{loss} \PaperCode{o}{+}\PaperCode{o}{=} \PaperCode{n}{weight} \PaperCode{o}{*} \PaperCode{n+nb}{mse}\PaperCode{p}{(}\PaperCode{n}{refined}\PaperCode{p}{,} \PaperCode{n}{x}\PaperCode{p}{)}
\end{Verbatim}
\end{algorithm}
\end{minipage}
\hfill
\begin{minipage}[t]{0.485\textwidth}
\vspace{0pt}
\begin{algorithm}[H]
\caption{Sampling with correction}
\label{alg:inference}
\begin{Verbatim}[commandchars=\\\{\},fontsize=\small,baselinestretch=1.0,formatcom=\color{black}]
\PaperCode{c+c1}{\PaperCodeZsh{} D: denoiser; C: corrector}
\PaperCode{c+c1}{\PaperCodeZsh{} K: corrections per step}
\PaperCode{n}{z} \PaperCode{o}{=} \PaperCode{n+nb}{randn}\PaperCode{p}{(}\PaperCode{n}{shape}\PaperCode{p}{)}
\PaperCode{k}{for} \PaperCode{n}{t}\PaperCode{p}{,} \PaperCode{n}{s} \PaperCode{o+ow}{in} \PaperCode{n+nb}{zip}\PaperCode{p}{(}\PaperCode{n}{times}\PaperCode{p}{,} \PaperCode{n}{times}\PaperCode{p}{[}\PaperCode{l+m+mi}{1}\PaperCode{p}{:}\PaperCode{p}{]}\PaperCode{p}{)}\PaperCode{p}{:}
    \PaperCode{n}{pred} \PaperCode{o}{=} \PaperCode{n}{D}\PaperCode{p}{(}\PaperCode{n}{z}\PaperCode{p}{,} \PaperCode{n}{t}\PaperCode{p}{)}
    \PaperCode{c+c1}{\PaperCodeZsh{}  correction steps}
    \PaperCode{k}{for} \PaperCode{n}{\PaperCodeZus{}} \PaperCode{o+ow}{in} \PaperCode{n+nb}{range}\PaperCode{p}{(}\PaperCode{n}{K}\PaperCode{p}{)}\PaperCode{p}{:}
        \PaperCode{n}{pred} \PaperCode{o}{=} \PaperCode{n}{C}\PaperCode{p}{(}\PaperCode{n}{pred}\PaperCode{p}{,} \PaperCode{n}{t}\PaperCode{p}{)}
    \PaperCode{n}{v} \PaperCode{o}{=} \PaperCode{p}{(}\PaperCode{n}{z} \PaperCode{o}{\PaperCodeZhy{}} \PaperCode{n}{pred}\PaperCode{p}{)} \PaperCode{o}{/} \PaperCode{n}{t}
    \PaperCode{n}{z} \PaperCode{o}{=} \PaperCode{n}{z} \PaperCode{o}{+} \PaperCode{p}{(}\PaperCode{n}{s} \PaperCode{o}{\PaperCodeZhy{}} \PaperCode{n}{t}\PaperCode{p}{)} \PaperCode{o}{*} \PaperCode{n}{v}
\PaperCode{n}{sample} \PaperCode{o}{=} \PaperCode{n}{z}
\end{Verbatim}
\end{algorithm}
\end{minipage}
\vspace{-5pt}
\end{figure}

%% file: sections/results.tex
\section{Results}
\label{sec:results}
{We evaluate correction on class-conditional and unconditional generation (Sec.~\ref{subsec:imagenet-generation}) and text-to-image generation (Sec.~\ref{subsec:text-to-image-generation}).
We also demonstrate that the base and correction models can share most parameters and that additional correction passes can be eliminated with distillation (Sec.~\ref{subsec:distillation}).} We detail our experimental settings in Appendix~\ref{app:imp}.

\subsection{ImageNet Generation}
\label{subsec:imagenet-generation}

We evaluate our method on widely used transformer-based architectures, including DiT~\citep{peebles2023scalable}, SiT~\citep{ma2024sit}, and JiT~\citep{li2026back}. We initialize both the base denoiser and a separate corrector from publicly released checkpoints and jointly fine-tune them following Algorithm~\ref{alg:training}. Unless otherwise specified, we follow each baseline's training recipe and evaluation protocol. Number of function evaluations (NFE) under correction measures both denoiser steps and correction steps. We approximately match the total NFE by reducing the number of denoising steps as the number of correction iterations $K$ increases.

\begin{figure}[!tbp]
    \centering
    \vspace{-10pt}
    \includegraphics[width=\linewidth]{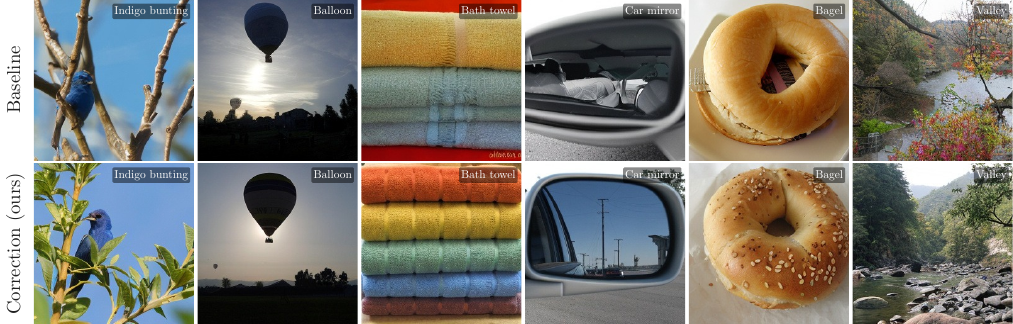}
    \caption{\textbf{Correction improves class-conditional sample quality.} We compare baseline (top) and correction (bottom), starting from the same noise in each column. Correction gives more coherent and detailed images for different classes and noises compared against the baseline.}
    \vspace{-8pt}
    \label{fig:conditional-paired}
\end{figure}
\input{tables/main_quan}

\par
\vspace{\parskip}
\begin{wrapfigure}{r}{0.35\textwidth}
  \centering
    \vspace{-30pt}
  \includegraphics[width=\linewidth]{
      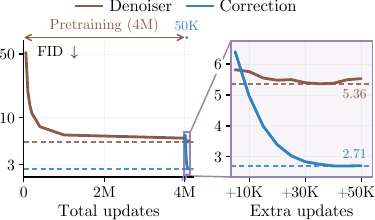
  }
  \captionsetup{font=footnotesize,skip=3pt}
    \caption{\textbf{Efficiency of correction training.}
    FID converges within relatively few corrector training steps.}
  \label{fig:pretraining-progress}
  \vspace{-25pt}
\end{wrapfigure}
\noindent

\vspace{-20pt}

\noindent\textbf{Class-conditional generation.}\label{subsec:class-conditional-generation}
We present class conditional image samples in Fig.~\ref{fig:conditional-paired}.
Correction produces sharp, realistic, and diverse images, all \textit{without applying any guidance}.
The resulting samples exhibit realism competitive with that achieved by guidance-based methods such as classifier-free guidance~\citep{ho2022classifier}, perturbation guidance~\citep{ahn2024self} and autoguidance~\citep{karras2024guiding}.
See Appendix~\ref{app:conditional-samples} for more samples.

\begin{figure}[!tbp]
    \centering
    \vspace{-15pt}
    \includegraphics[width=\linewidth]{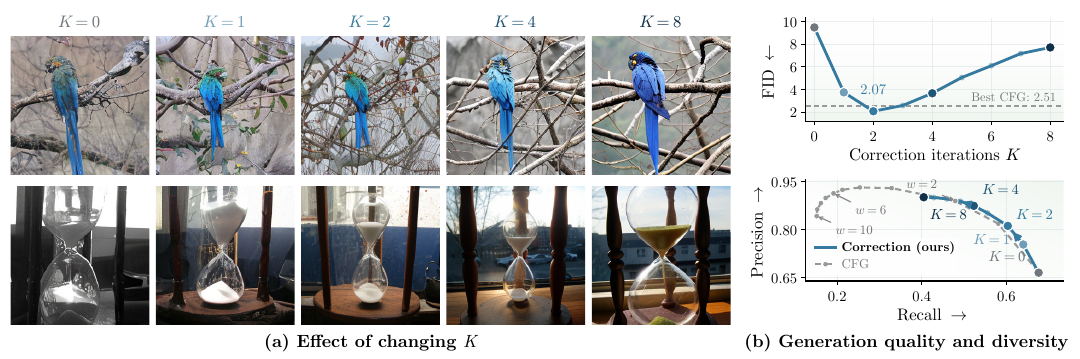}
    \vspace{-17pt}
    \caption{\textbf{Iterative correction provides a controllable refinement trajectory on ImageNet.} Left: Evolvement of the same samples over more correction steps. The main content remains the same while quality improves. Right: FID and precision--recall across $K$. Correction gives a controllable tradeoff between coverage and quality and lower minimum FID. }
    \label{fig:controllable-refinement}
\end{figure}

In Tab.~\ref{tab:imagenet-results}, we evaluate generation using FID, precision, and recall~\citep{heusel2017gans,kynkaanniemi2019improved}.
Correction \textit{significantly reduces FID at matched NFE} (DiT: \DitFineTuneControlFID{} $\to$ \textbf{\DitCorrectionBestFID},
SiT: \SitBaseFID{} $\to$ \textbf{\SitCorrectionBestFID},
and SiT$_{\mathrm{REPA}}$: \RepaMatchedBaseFID{} $\to$ \textbf{\RepaMatchedCorrectionFID}.).
This suggests that correction is a better allocation of inference compute than simply increasing denoising steps.  These gains require little additional training: for SiT$_{\mathrm{REPA}}$, 50K updates amount to just 1.25\% of its 4M pretraining updates. Fig.~\ref{fig:pretraining-progress} compares FID convergence with correction versus continued denoising. Fig.~\ref{fig:controllable-refinement} shows how the number of correction iterations $K$ affects the final samples and their fidelity--diversity trade-off. Correction effectively suppresses artifacts and improves object coherence, yielding lower FID and higher precision at the cost of some recall. We further analyze the power spectra of the generated samples in Appendix~\ref{app:power-spectrum-analysis}.

\begin{figure}[!tbp]
    \centering
    \includegraphics[width=\linewidth]{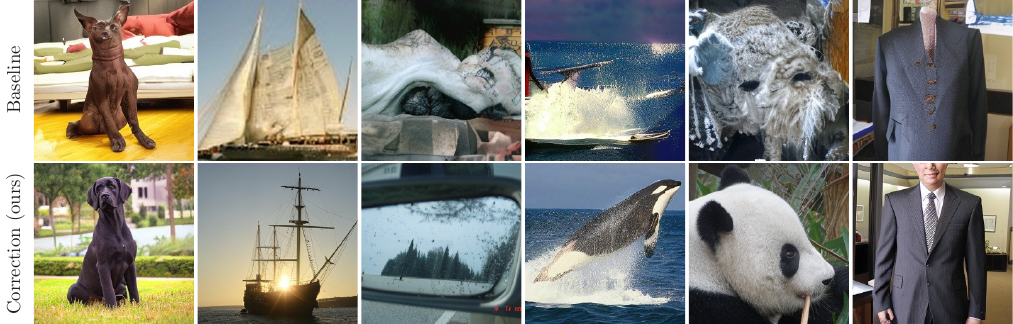}
    \caption{\textbf{Correction improves \textit{unconditional} sample quality on ImageNet.} Without correction, generated samples exhibit artifacts and structural inconsistencies (top), while corrective sampling produces diverse and high-quality images (bottom). The samples in each column share the same initial noise.}
    \label{fig:unconditional-paired}
    \vspace{-10pt}
\end{figure}

\paragraph{Unconditional generation.}
\label{subsec:unconditional-generation}
Unconditional generation remains a challenging task since no class or text conditioning is available to constrain the outputs~\citep{li2024return}. We show that for unconditional generation, our prediction correction significantly improves image quality and coherence, all without requiring guidance. We present qualitative comparisons of corrected and uncorrected samples in Fig.~\ref{fig:unconditional-paired}. Correction removes structural inconsistencies and artifacts from the baseline samples while maintaining the main subject.

\begin{figure}[!tbp]
\centering
\raisebox{7.98bp}{%
\begin{minipage}[c]{0.38\linewidth}
\input{tables/uncond_generation}
\end{minipage}}\hfill
\begin{minipage}[c]{0.58\linewidth}
\includegraphics[width=\linewidth]{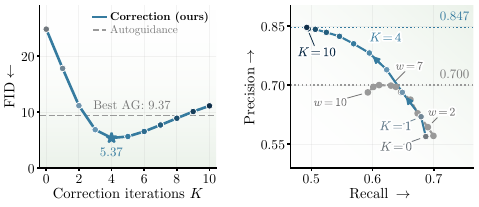}
\end{minipage}
\par
\begin{minipage}[t]{0.40\linewidth}
\vspace{0pt}
\captionsetup{type=table,font=footnotesize,skip=5pt,justification=raggedright,singlelinecheck=false}
\caption{\textbf{Correction lowers unconditional FID without guidance.} \UncondBaseStep{} training updates, \UncondBaseNFE{} NFE, autoguidance $w=\UncondAGBestW$.}
\label{tab:unconditional}
\end{minipage}\hfill
\begin{minipage}[t]{0.56\linewidth}
\vspace{0pt}
\captionsetup{type=figure,font=footnotesize,skip=5pt}
\caption{\textbf{Correction reaches lower FID and higher peak precision than autoguidance.} Dashed: best autoguidance FID. Dotted: each method's observed peak precision.}
\label{fig:unconditional}
\end{minipage}
\vspace{-6pt}
\end{figure}

Quantitatively, we compare correction and autoguidance~\citep{karras2024guiding} in Tab.~\ref{tab:unconditional} on ImageNet unconditional generation. Correction with $K=\UncondCorrectionBestK$ substantially improves FID over the baseline and the best evaluated autoguidance scale ($w \in \{1,2,\dots,10\}$, with the released pretrained model as the guide), while larger $K$ trades recall for higher precision (Fig.~\ref{fig:unconditional}).


\paragraph{Diversity preservation.}
\label{app:cfg-comparison}

\begin{figure}[!t]
\centering
\includegraphics[width=\linewidth]{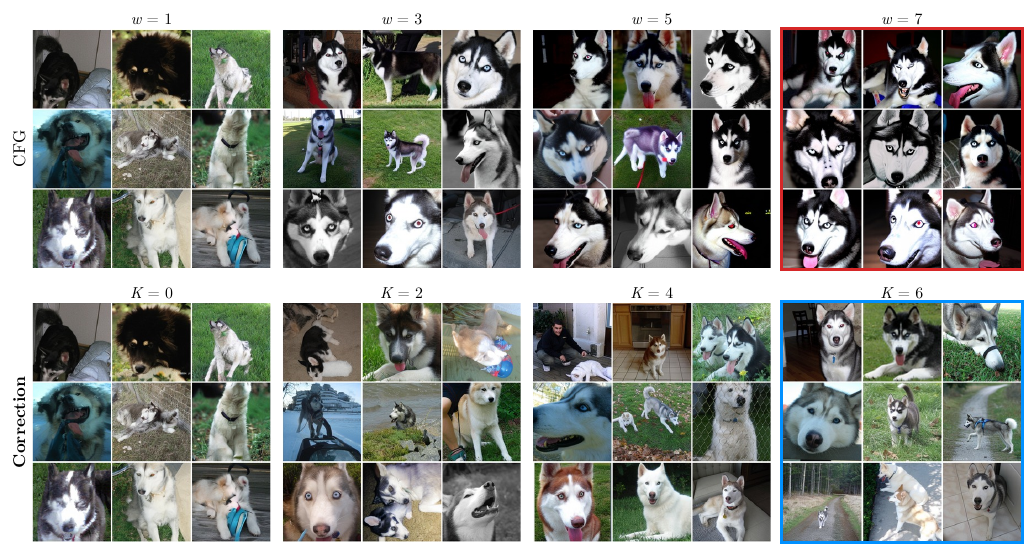}
\caption{\textbf{Correction preserves more diversity and produces less oversaturation than CFG.} Samples from SiT$_{\mathrm{REPA}}$-XL/2 with increasing correction depth $K$ (bottom) or CFG scale $w$ (top), at approximately matched NFE, same initial noise.}
\label{fig:correction-cfg-comparison}
\end{figure}

Although comparison with CFG is not our main goal, our method addresses two key limitations of CFG: reduced diversity and oversaturation. Fig.~\ref{fig:correction-cfg-comparison} shows samples from the same ImageNet class as we vary the CFG scale $w$ and the number of correction iterations $K$. At higher CFG scales, the samples become less diverse, converging on canonical close-up portraits, while their colors become increasingly saturated. With correction, the dogs appear in more varied poses, framings, and backgrounds while retaining sharp details and realism. Additional class-wise comparisons are shown in
Figs.~\ref{fig:correction-guidance-class-96}
and~\ref{fig:correction-guidance-class-1} in the appendix.

\subsection{Text-to-Image Generation}
\label{subsec:text-to-image-generation}
To assess our method on a more complex, practically relevant task, we evaluate text-to-image generation using NanoGen~\citep{diffusionbench}, a unified training framework for diffusion transformers. We use a pretrained DiT~\citep{peebles2023scalable} with a DDT head~\citep{wang2026ddt}, operating in the latent space of the FLUX.2 VAE~\citep{flux-2-2025}. We use the JourneyDB~\citep{sun2023journeydb}, Short-Caption, and Long-Caption subsets of the BLIP-3o pretraining mixture~\citep{chen2025blip3}, with aspect-preserving preprocessing to $256\times256$. For the quantitative comparison, we jointly fine-tune the denoiser and corrector for 50K steps, adding a relatively small training cost compared with pretraining the base model.
\input{generated/diffusionbench_samples}
Fig.~\ref{fig:diffusionbench-samples} illustrates changes in object structure without guidance: correction separates the violin bodies, clarifies the guitar and car geometry, and organizes cluttered interiors.
Quantitatively, Tab.~\ref{tab:diffusionbench-t2i} shows higher GenEval~\citep{ghosh2023geneval} and DPG-Bench~\citep{hu2024ella} scores with correction at approximately matched NFE, first without CFG and then with correction applied before CFG. Both comparisons use the same-checkpoint baseline and DiffusionBench's native evaluation pipeline; sampling protocols are provided in Table~\ref{tab:main-settings}, and additional guidance scales in Appendix~\ref{app:figure-table-details}.
Fig.~\ref{fig:diversity-control} compares different combinations of CFG and correction. Stronger CFG alone can improve conditioning adherence and sample quality, but often at the cost of reduced diversity and oversaturated or stylized appearances. Lowering the guidance scale alleviates these effects, but may leave samples under-refined. We find that \textit{applying correction jointly with CFG alleviates the diversity issue} while preserving the quality refinement. We observe that CFG alone produces visually similar samples, while combining CFG and correction leads to higher diversity (Fig.~\ref{fig:diversity-control}).

\begin{figure}[!h]
\centering
\captionsetup{font=small,skip=4pt}
\input{generated/guidance_correction_matrix}
\end{figure}

\label{subsec:cfg-comparison}
\label{par:cfg-comparison}

\input{tables/human_evaluation}
\begin{samepage}
\par\end{samepage}

\subsection{Making correction efficient}
\paragraph{Parameter-efficient correction through weight sharing.}
The denoiser and corrector can share a backbone while retaining separate lightweight input embedders, which contribute only a small fraction of the total parameters. On SiT-XL/2, this configuration achieves performance comparable to separate networks with roughly half the total parameters (Tab.~\ref{tab:sit-shared-ladder}). In the shared configuration, the backbone receives gradients from both the denoising and correction losses.

We also compare against a larger single denoiser with roughly matched parameter count, and find that iterative correction yields substantially larger gains than simply doubling the denoiser depth (Sec.~\ref{sec:matched-baselines}, Tab.~\ref{tab:deeper-denoiser}).

\input{tables/efficiency_pair}
\label{subsec:distillation}

\paragraph{Distilling correction into the base model.}
Weight sharing reduces parameter overhead, but each correction iteration still requires an additional forward pass. We therefore distill corrected predictions into the original denoiser. The student is initialized from the teacher's base denoiser and trained to match its predictions after $k^*$ correction iterations:
\begin{equation}
    \mathcal{L}_{\mathrm{distill}}
    = w_{\mathrm{distill}}(t)\norm{D_{\theta'}(\bm{z}_t,t)-\hat{x}^{(k^*)}}_2^2,
\end{equation}
    where $k^*$ controls the number of teacher correction iterations. Tab.~\ref{tab:distillation} reports results for class-conditional ImageNet-256 generation using SiT-XL/2 with $k^*=2$. The student achieves comparable FID while reducing NFE by approximately 67\%. We use $w_{\mathrm{distill}}(t)=1/t^2$ for $t>0$, corresponding to velocity-space MSE.

%% file: tables/main_quan.tex
\providecommand{\MainTableRowColor}{black}
\begin{table}[t]
    \centering
    \small
    \setlength{\tabcolsep}{3pt}
    \renewcommand{\arraystretch}{1.12}
    \caption{\textbf{Correction improves unguided class-conditional generation.} We compare FID-50K, precision, and recall on ImageNet-256 class-conditional image generation across 4 base models. We report consistent improvement on FID and precision over unguided baseline and it is competitive with CFG. All methods are evaluated at approximately 250 NFEs. See Tab.~\ref{tab:main-settings} for evaluation details.}
    \label{tab:imagenet-results}

    \begin{minipage}[t]{0.48\linewidth}
    \centering
    \begin{tabular*}{\linewidth}{
        @{\extracolsep{\fill}}
        >{\color{\MainTableRowColor}}l
        *{3}{>{\color{\MainTableRowColor}}r}@{}
    }
        \toprule
        \textbf{DiT-XL/2} & \textbf{FID} $\downarrow$ & \textbf{Prec.} $\uparrow$ & \textbf{Rec.} $\uparrow$ \\
        \midrule

        Unguided (baseline) & \DitFineTuneControlFID & \DitBasePrecision & \textbf{\DitBaseRecall} \\

        \textbf{Corrective sampling} & \textbf{\DitCorrectionBestFID} & \textbf{\DitCorrectionBestPrecision} & \DitCorrectionBestRecall \\

        \cmidrule(lr){1-4}
        \noalign{\gdef\MainTableRowColor{black!50}}
        CFG (reference) & \DitCfgMatchedFID & \DitCfgMatchedPrecision & \DitCfgMatchedRecall \\
        \noalign{\gdef\MainTableRowColor{black}}

        \bottomrule
        \addlinespace[4pt]
        \toprule
        \textbf{SiT-XL/2} & \textbf{FID} $\downarrow$ & \textbf{Prec.} $\uparrow$ & \textbf{Rec.} $\uparrow$ \\
        \midrule

        Unguided (baseline) & \SitBaseFID & \SitBasePrecision & \textbf{\SitBaseRecall} \\

        \textbf{Corrective sampling} & \textbf{\SitCorrectionBestFID} & \textbf{\SitCorrectionBestPrecision} & \SitCorrectionBestRecall \\

        \cmidrule(lr){1-4}
        \noalign{\gdef\MainTableRowColor{black!50}}
        CFG (reference) & \SitCfgMatchedFID & \SitCfgMatchedPrecision & \SitCfgMatchedRecall \\
        \noalign{\gdef\MainTableRowColor{black}}

        \bottomrule
    \end{tabular*}
    \end{minipage}%
    \hfill
    \begin{minipage}[t]{0.48\linewidth}
    \centering
    \begin{tabular*}{\linewidth}{
        @{\extracolsep{\fill}}
        >{\color{\MainTableRowColor}}l
        *{3}{>{\color{\MainTableRowColor}}r}@{}
    }
        \toprule
        \textbf{SiT$_{\mathrm{REPA}}$-XL/2} & \textbf{FID} $\downarrow$ & \textbf{Prec.} $\uparrow$ & \textbf{Rec.} $\uparrow$ \\
        \midrule

        Unguided (baseline) & \RepaMatchedBaseFID & \RepaMatchedBasePrecision & \textbf{\RepaMatchedBaseRecall} \\
        \textbf{Corrective sampling} & \textbf{\RepaMatchedCorrectionFID} & \textbf{\RepaMatchedCorrectionPrecision} & \RepaMatchedCorrectionRecall \\

        \cmidrule(lr){1-4}
        \noalign{\gdef\MainTableRowColor{black!50}}
        CFG (reference) & \RepaCfgFID & \RepaCfgPrecision & \RepaCfgRecall \\
        \noalign{\gdef\MainTableRowColor{black}}

        \bottomrule
        \addlinespace[4pt]
        \toprule
        \textbf{JiT-L/16} & \textbf{FID} $\downarrow$ & \textbf{Prec.} $\uparrow$ & \textbf{Rec.} $\uparrow$ \\
        \midrule

        Unguided (baseline) & \JitLBaseFID & \JitLBasePrecision & \textbf{\JitLBaseRecall} \\
        \textbf{Corrective sampling} & \textbf{\JitLIntervalFID} & \textbf{\JitLIntervalPrecision} & \JitLIntervalRecall \\

        \cmidrule(lr){1-4}
        \noalign{\gdef\MainTableRowColor{black!50}}
        CFG (reference) & \JitLCfgFID & \JitLCfgPrecision & \JitLCfgRecall \\
        \noalign{\gdef\MainTableRowColor{black}}

        \bottomrule
    \end{tabular*}
    \end{minipage}
    \par\smallskip
    \vspace{-10pt}
\end{table}

%% file: tables/uncond_generation.tex
\centering
\fontsize{8.5}{10}\selectfont
\setlength{\tabcolsep}{1pt}
\renewcommand{\arraystretch}{1.12}
\begin{tabular*}{\linewidth}{@{\extracolsep{\fill}}lrrr@{}}
    \toprule
    \noalign{\vskip 2.5pt}
    Method & FID $\downarrow$ & Prec. $\uparrow$ & Rec. $\uparrow$ \\
    \midrule
    \noalign{\vskip 2pt}
    Baseline & \UncondBaseFID & \UncondBasePrecision & \textbf{\UncondBaseRecall} \\
    \textbf{Correction} ($K=1$) & \UncondCorrectionOneFID & \UncondCorrectionOnePrecision & \UncondCorrectionOneRecall \\
    \textbf{Correction} ($K=\UncondCorrectionBestK$) & \textbf{\UncondCorrectionBestFID} & \textbf{\UncondCorrectionBestPrecision} & \UncondCorrectionBestRecall \\
    \cmidrule{1-4}
    \textcolor[HTML]{777777}{Autoguidance} & \textcolor[HTML]{777777}{\UncondAGBestFID} & \textcolor[HTML]{777777}{\UncondAGBestPrecision} & \textcolor[HTML]{777777}{\UncondAGBestRecall} \\
    \noalign{\vskip 2.5pt}
    \bottomrule
\end{tabular*}

%% file: generated/diffusionbench_samples.tex
\begin{figure}[!t]
    \centering
        \vspace{-20pt}
    \includegraphics[width=\textwidth]{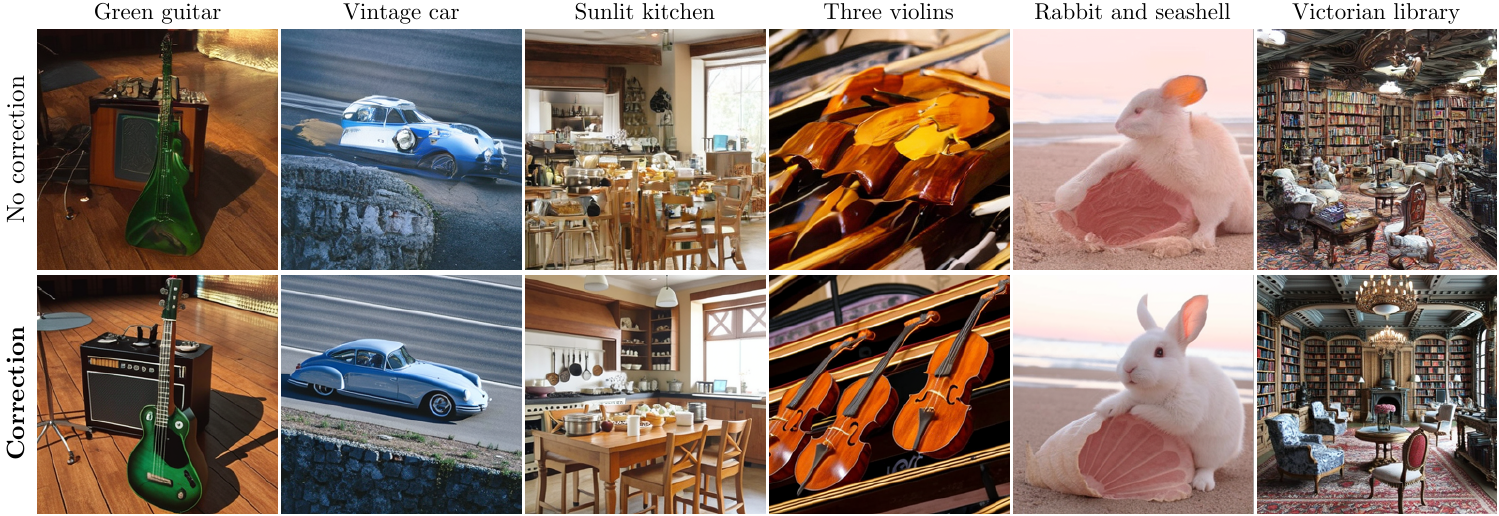}
    \caption{\textbf{Correction improves object structure in \textit{text-to-image} generation.} Compared with no correction (top), correction (bottom) separates the violin bodies, clarifies the guitar and car geometry, and organizes cluttered interiors. All samples are without CFG; full prompts are in Appendix~\ref{app:diffusionbench-full-prompts}.}
    \label{fig:diffusionbench-samples}
    \par\medskip
    \begingroup
    \captionsetup{type=table}
\input{generated/diffusionbench_t2i}
    \endgroup
\end{figure}

%% file: generated/diffusionbench_t2i.tex
\begingroup
\centering
\fontsize{8.5}{10}\selectfont
\setlength{\tabcolsep}{1pt}
\renewcommand{\arraystretch}{1.12}
\begin{minipage}[t]{0.485\linewidth}
\vspace{0pt}\centering
\textit{Without CFG}\par\vspace{2pt}
\begin{tabular*}{\linewidth}{@{\extracolsep{\fill}}lrrr@{}}
\toprule
Method & NFE & GenEval $\uparrow$ & DPG-Bench $\uparrow$ \\
\midrule
Baseline & 250 & 28.30 & 70.71 \\
\textbf{Correction} ($K=6$) & 245 & \textbf{50.09} & \textbf{76.82} \\
\bottomrule
\end{tabular*}
\end{minipage}\hfill%
\begin{minipage}[t]{0.485\linewidth}
\vspace{0pt}\centering
\textit{With CFG}\par\vspace{2pt}
\begin{tabular*}{\linewidth}{@{\extracolsep{\fill}}lrrr@{}}
\toprule
Method & NFE & GenEval $\uparrow$ & DPG-Bench $\uparrow$ \\
\midrule
Baseline & 250 & 68.08 & 83.38 \\
\textbf{Correction} ($K=2$) & 246 & \textbf{70.48} & \textbf{83.69} \\
\bottomrule
\end{tabular*}
\end{minipage}
\par
\captionsetup{font=footnotesize,skip=5pt,justification=raggedright,singlelinecheck=false}
\caption{\textbf{Correction improves text-to-image alignment with and without CFG.} We observe consistent improvements on GenEval and DPG-Bench by applying correction over the baseline, both with and without CFG, evaluated at approximately matched NFE. See Table~\ref{tab:main-settings} for detailed settings.}
\vspace{-10pt}
\label{tab:diffusionbench-t2i}
\endgroup

%% file: generated/guidance_correction_matrix.tex
\vspace{-5pt}
\includegraphics[width=\linewidth]{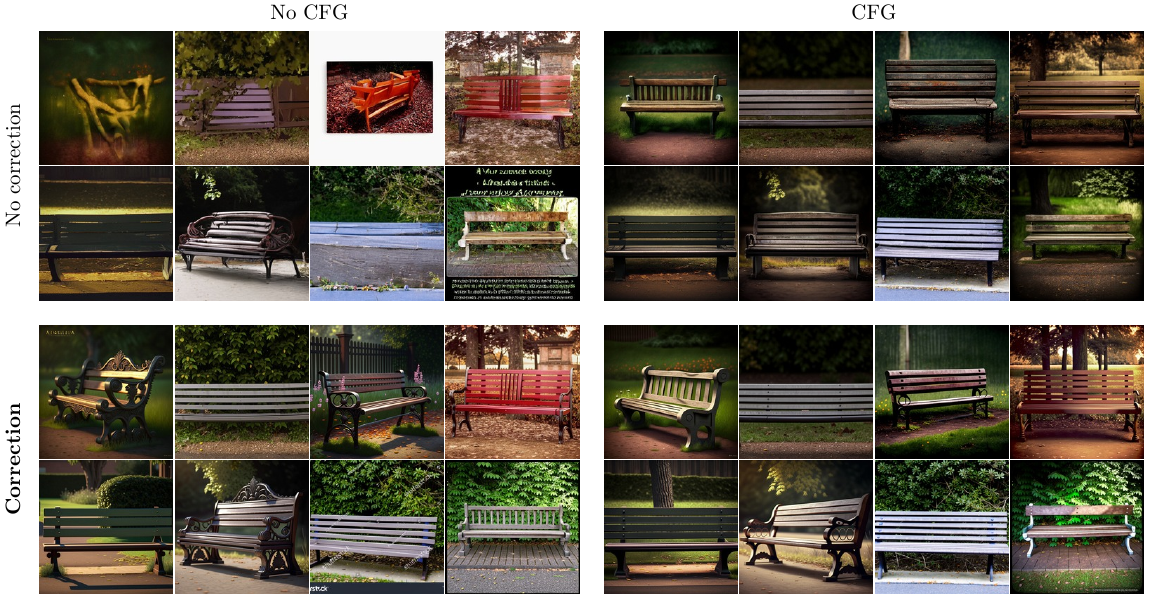}
\captionof{figure}{\textbf{Correction improves the diversity of CFG samples while preserving quality.} Columns compare no guidance ($w=1$) and CFG ($w=6$); rows compare no correction ($K=0$) and correction ($K=6$). All settings use the same eight seeds and number of denoising steps. Corrected samples show more varied bench shapes and viewpoints, with less oversaturation and a more natural appearance. Prompt: ``a photo of a bench''.}
\label{fig:diversity-control}

%% file: tables/human_evaluation.tex
\paragraph{Unconditional generation on web images.}
Using the same model, we also present \textit{unconditional} generation results in Appendix~\ref{app:t2i-unconditional}. Surprisingly, correction improves realism while preserving both the diversity of web images and the overall structure of the corresponding uncorrected samples.

\begin{wraptable}{r}{0.43\textwidth}
\input{tables/human_preference_float}
\vspace{-10pt}
\end{wraptable}
\noindent\textbf{Human evaluation.}
To assess sample quality and diversity through human judgments, we conduct a preference study (Tab.~\ref{tab:human-evaluation}) using \HumanStudyPromptCount{} GenEval single-object prompts on the Rapidata platform\footnote{\scriptsize\url{https://www.rapidata.ai/}}. Raters answer ``\HumanStudyQualityInstruction'' for individual images and ``\HumanStudyDiversityInstruction'' for four-image groups. Raters prefer CFG $+$ Correction over the baseline using DiffusionBench's default CFG scale~\citep{diffusionbench} in more than 60\% of responses. Evaluation details are provided in Appendix~\ref{app:human-evaluation}.
\par

%% file: tables/human_preference_float.tex
\centering
\fontsize{8.5}{10}\selectfont
\setlength{\tabcolsep}{1pt}
\renewcommand{\arraystretch}{1.12}
\input{generated/human_evaluation_body}

\captionsetup{type=table,font=footnotesize,skip=5pt,justification=raggedright,singlelinecheck=false}
\caption{\textbf{Correction is preferred for quality and diversity.} Released CFG ($w=\HumanStudyBaselineCFG$) versus CFG followed by correction ($w=\HumanStudyCFG$, $K=\HumanStudyK$).}
\label{tab:human-evaluation}

%% file: generated/human_evaluation_body.tex
\begin{tabular*}{\linewidth}{@{\extracolsep{\fill}}lrr@{}}
\toprule
Method & Quality $\uparrow$ & Diversity $\uparrow$ \\
\midrule
Baseline & 39.25\% & 39.75\% \\
\textbf{Correction} & \textbf{60.75\%} & \textbf{60.25\%} \\
\bottomrule
\end{tabular*}

%% file: tables/efficiency_pair.tex
\begin{table}[!htbp]
\input{generated/sit_model_budgets}
\centering
\begin{minipage}[t]{0.48\linewidth}
\vspace{0pt}
\captionsetup{font=small}
\input{generated/sit_shared_summary_body}
\end{minipage}\hfill
\begin{minipage}[t]{0.48\linewidth}
\vspace{0pt}
\captionsetup{font=small}
    \centering
    \small
    \setlength{\tabcolsep}{3pt}
    \begin{tabular}{@{}lrrcrr@{}}
        \toprule
        Model & Params & FID $\downarrow$ & $K$ & Steps & NFE $\downarrow$ \\
        \midrule
        Base & \SitBaseParams{} & \DistillTableBaselineFID{} & -- & \DistillTableBaselineSteps{} & \DistillTableBaselineNFE{} \\
        Teacher & \SitSeparateParams{} & \textbf{\DistillKTwoTeacherFID{}} & \DistillTeacherK{} & \DistillKTwoTeacherSteps{} & \DistillKTwoTeacherNFE{} \\
        Student & \SitBaseParams{} & \DistillKTwoStudentMatchedFID{} & -- & \DistillKTwoStudentMatchedSteps{} & \DistillKTwoStudentMatchedNFE{} \\
        Student & \SitBaseParams{} & \DistillKTwoStudentFID{} & -- & \DistillKTwoStudentSteps{} & \textbf{\DistillKTwoStudentNFE{}} \\
        \bottomrule
    \end{tabular}
    \caption{\textbf{Distillation achieves similar FID under much smaller NFE.} Distilling a SiT-XL/2 correction model into the base model leads to a similar FID performance with 3 times less NFE on ImageNet.}
    \label{tab:distillation}
\end{minipage}
\vspace{-10pt}
\end{table}

%% file: generated/sit_model_budgets.tex
\providecommand{\SitBaseParams}{0.68B}
\providecommand{\SitSeparateParams}{1.35B}
\providecommand{\SitSharedParams}{0.68B}

%% file: generated/sit_shared_summary_body.tex
    \centering
    \small
    \setlength{\tabcolsep}{3pt}
    \begin{tabular}{@{}lrrrr@{}}
        \toprule
        Model & Params & FID $\downarrow$ & $K$ & NFE \\
        \midrule
        Baseline & 0.68B & 8.55 & 0 & 250 \\
        Separate corrector & 1.35B & \textbf{1.92} & 2 & 249 \\
        Shared corrector & 0.68B & 4.47 & 1 & 250 \\
        Shared corrector & 0.68B & 2.14 & 3 & 248 \\
        \bottomrule
    \end{tabular}
    \caption{\textbf{A shared backbone retains similar correction quality.} Using shared layers from SiT-XL/2 on ImageNet with 100K steps leads to similar FID, suggesting that correction's advantage does not primarily come from using more parameters.}
    \label{tab:sit-shared-ladder}

%% file: sections/discussion_new.tex
\section{Discussion}
\label{sec:discussion}
In this section, we investigate where the quality gains from correction might come from. We first compare a denoiser--corrector pair with a single deeper denoiser containing the same total number of transformer layers (Sec.~\ref{sec:matched-baselines}).
We then empirically examine the behavior of iterative correction  and its relationship to the training--inference gap (Sec.~\ref{sec:how-iterative-correction-changes}). Finally, we examine how the number of corrections seen during training and denoiser updates affect the quality gains from iterative correction (Sec.~\ref{sec:how-training-affects-iterative-correction}).

\newpage
\subsection{Parameter-matched baselines}
\label{sec:matched-baselines}

\paragraph{A single correction achieves similar FID to a monolithic denoiser.}
\input{generated/matched_baseline_allocation}
A natural question is whether a denoiser--corrector pair with a single correction ($K=1$) can outperform a monolithic denoiser with the same total number of transformer layers (Fig.~\ref{fig:matched-baseline-allocation}, first and second rows). We also use the same number of denoising steps for both configurations. More specifically, we compare a 28-layer denoiser ($D_{28}$) with a 14-layer denoiser and a 14-layer corrector ($D_{14}+C_{14}$), both trained from scratch for 80K updates using LightningDiT~\citep{yao2025reconstruction} on ImageNet 256$\times$256~\citep{ILSVRC15}. The two configurations have approximately the same parameter count (675M and 681M, respectively). With a single correction at each denoising step ($K=1$), the denoiser--corrector pair achieves similar FID to the monolithic denoiser (8.74 versus 8.93). Thus, splitting the model into a denoiser and a corrector gives little quality gain when trained from scratch.%

\paragraph{Iterative correction outperforms a monolithic denoiser.}
We then reuse the same corrector for $K=3$ iterations at each denoising step (Fig.~\ref{fig:matched-baseline-allocation}, third row). This lowers FID to 5.66, while halving the number of denoising steps keeps the total transformer-layer evaluation budget unchanged (Tab.~\ref{tab:deeper-denoiser}). These results suggest that \textit{iterative} correction makes effective use of inference compute.

Even when we match the effective depth within each denoising step, the denoiser--corrector pair still performs better. The shallow $D_7+C_7$ pair with $K=3$ achieves lower FID than $D_{28}$ (4.21 versus 4.88), using the same number of denoising steps and roughly half as many parameters (Tab.~\ref{tab:deeper-denoiser}. See Fig.~\ref{fig:matched-effective-depth-allocation} in Appendix). These results are from training from scratch. When starting from a pretrained model, as in our main experiments in Sec.~\ref{sec:results}, a single correction can outperform a deeper monolithic denoiser and converges much faster (Fig.~\ref{fig:convergence-comparison} in the Appendix).

\begin{table}[!htbp]
    \centering
    \captionsetup{font=footnotesize}
\input{generated/lightningdit_matched_baselines_table}
\end{table}

\Needspace{4\baselineskip}

\subsection{How iterative correction changes predictions and sampling}
\label{sec:how-iterative-correction-changes}
\input{generated/iterative_correction_qualitative}

\paragraph{Iterative correction sharpens the denoiser's predictions.}
The corrector is trained to reconstruct the clean target from the denoiser’s prediction. Its output is another clean-image prediction, allowing the same mapping to be applied repeatedly at a fixed noise level. We examine how this repetition changes the predictions. 
Fig.~\ref{fig:repeated-correction} shows this visually. We use noisy data $\bm{z}_t=(1-t)\bm{x}+t\bm{\epsilon}$ as input to JiT-L/16 and show its predictions before and after correction. As the correction count $K$ increases, object structures become increasingly coherent, producing realistic-looking, globally coherent images at $K=9$. For example, in (a), the blurry, ``average-like'' predictions of dogs, pandas, and cars become distinct, well-formed objects. In (b), the corrector sometimes makes overly coherent structures, such as disk-like pizza toppings or strawberry seeds arranged in a regular grid.

In our from-scratch experiments, a single correction yields similar FID to a deeper, parameter-matched denoiser (Sec.~\ref{sec:matched-baselines}). However, repeated application of the same corrector substantially improves sample quality at a matched inference budget. Our intuition is that splitting the computation into denoising and correction exposes a refinement operation that can be reused. The denoiser first maps the noisy input into a clean-image prediction, and the corrector learns to refine this prediction toward the original clean target. A single correction may provide refinement similar to that achieved by additional denoiser layers. Repeated correction, however, reapplies this learned mapping to its own outputs, beyond the single iteration used during training. This may reinforce coherent structures, sometimes excessively, producing overly regular patterns such as the disks and grids.

\input{generated/correction_training_loss}
\paragraph{Does iterative correction reduce denoising error?}
Iterative correction sharpens predictions, but the corrected outputs can differ substantially from the original clean image (Fig.~\ref{fig:repeated-correction}). Thus, we analyze whether these visual improvements also reduce denoising error. We denote the denoising error after correction by
\begin{equation}
\mathcal{L}_{\mathrm{denoise}}^{(k)}
=
\mathbb{E}_{\bm{x},\bm{\epsilon},t}\!\left[
  \norm{\hat{\bm{x}}^{(k)}-\bm{x}}_2^2
\right],
\label{eq:correction-loss-k}
\end{equation}
where $k$ is the number of correction iterations.

We compare this error with the base denoiser's error (Eq.~\ref{eq:denoising-loss}) by measuring
$\mathcal{L}_{\mathrm{denoise}}^{(k)}-\mathcal{L}_{\mathrm{denoise}}$.
Fig.~\ref{fig:increased-training-loss} shows positive differences for $k\geq3$, meaning that the denoising error \textit{increases} after correction.
Thus, even as predictions become sharper and more coherent, iterative correction can move them farther on average from the optimal denoising prediction.

\paragraph{Iterative correction reduces the training--inference gap.}
\label{sec:gap}
\input{generated/training_inference_gap}

How can iterative correction improve sample quality even when it increases denoising error? Denoising error is measured on noisy inputs drawn from the \textit{training  distribution}, whereas intermediate states encountered during sampling can follow a different distribution~\citep{song2025selective}. Each corrected prediction is used to construct the next noisy state, which becomes the denoiser's input. Thus, we hypothesize that correction may improve generation by bringing these states closer to the training distribution.
We measure the training--inference gap using a classifier two-sample test~\citep{lopez2016revisiting}. We train classifiers to distinguish intermediate states $\bm{z}_t$ from the training distribution, $\bm{z}_t=(1-t)\bm{x}+t\bm{\epsilon}$, from those along denoising trajectories at inference. 
Specifically, we experiment with SiT~\citep{ma2024sit} on 100K intermediate states, and report binary cross-entropy (BCE) and AUROC on held-out datasets. A smaller training-inference gap should make the two distributions harder to distinguish, bringing AUROC closer to $0.5$ (random chance).

As shown by the BCE in Fig.~\ref{fig:classification-test}, the classifier can distinguish training and inference states, with classification becoming easier as the noise level decreases. This trend is consistent with errors accumulating during sampling due to exposure bias~\citep{ning2024elucidating,ning2023input}. Notably, a single correction (gray lines) yields higher BCE and lower AUROC, suggesting that correction \textit{reduces the training--inference gap}. Iterative correction generally reduces the gap further (blue lines), which may contribute to better sample quality even when reconstruction error on training inputs increases. 

\subsection{How training affects iterative correction}
\label{sec:how-training-affects-iterative-correction}
\paragraph{Extrapolation is key to substantial quality gains.}
\input{generated/ablation_single_multi}

We denote the number of correction iterations applied during training by $K_{\mathrm{train}}$. In our default setting ($K_{\mathrm{train}}=1$), the corrector sees only the denoiser's initial prediction during training. We also train the corrector to predict the target after multiple correction iterations ($K_{\mathrm{train}}>1$). Increasing $K_{\mathrm{train}}$ shifts the best-performing inference correction count $K$ toward larger values, while the best achieved FID remains similar (Fig.~\ref{fig:repeated-correction-graph}, left). Notably, \textbf{the best performance comes from applying correction more times than supervised during training}. The right panel shows that increasing $K_{\mathrm{train}}$ lowers denoising error below the base denoiser's only for correction iterations included in training. For $K>K_{\mathrm{train}}$, the error remains above this baseline, even at the correction count with the best FID.

Our intuition is that the corrector learns a mapping toward the optimal denoising prediction. Supervising more correction iterations may simply spread this correction over more steps, so more inference iterations are needed to obtain similar gains from \textit{extrapolating} the learned mapping beyond the iterations used during training.

\input{generated/ablation_freeze_samples}
\paragraph{Denoiser updates affect iterative correction.}
\input{generated/joint_frozen_curves_wrap}
To examine how denoiser updates affect iterative correction, we compare (a) jointly training the denoiser and corrector, our default setting in Algorithm~\ref{alg:training}, with (b) freezing the denoiser and training only the corrector under the same correction framework. Interestingly, as shown in Fig.~\ref{fig:ablation-freeze-samples}, samples in (b) improve with correction but change more abruptly in structure and appearance, with slight oversaturation typical of CFG. Fig.~\ref{fig:ablation-freeze-curves} shows similar FID for both settings at $K=5$. With further correction, FID continues to decrease with joint training but increases again with a frozen denoiser despite visually improved samples, similar to the trend at higher CFG scales.
One possible interpretation is that, in (b), the corrector specializes in the errors of a particular denoiser checkpoint, making it less robust to iterative application. In (a), the denoiser continues to evolve during training, exposing the corrector to a broader range of errors. This may help the corrector generalize to its own outputs and remain effective across different correction counts $K$.

%% file: generated/matched_baseline_allocation.tex
\begin{figure}[!ht]
    \centering
    \vspace{-15pt}
    \includegraphics[width=\linewidth]{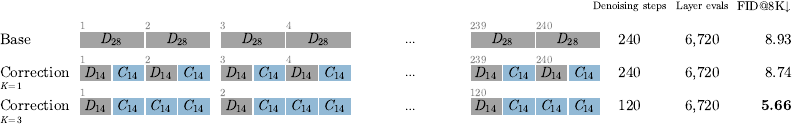}
    \caption{\textbf{Matched parameters.} Repeated correction outperforms a deeper denoiser at a comparable parameter count and matched inference budget.}
    \label{fig:matched-baseline-allocation}
\end{figure}

%% file: generated/lightningdit_matched_baselines_table.tex
\begingroup
\centering
\small
\setlength{\tabcolsep}{3pt}
\begin{tabular*}{\linewidth}{@{\extracolsep{\fill}}lrlrrrr@{}}
\toprule
Model & Params (M) & Sampling & Steps & Layer evals & FID@8K$\downarrow$ & FID@50K$\downarrow$ \\
\midrule
\multicolumn{7}{@{}l}{\textit{Matched parameters}} \\
\addlinespace[2pt]
$D_{28}$ & 675 & Unguided & 240 & 6,720 & 8.93 & -- \\
$D_{28}$ & 675 & CFG ($w=5$) & 120 & 6,720 & 6.17 & -- \\
$D_{14}$ + $C_{14}$ & 681 & $K=1$ & 240 & 6,720 & 8.74 & -- \\
$D_{14}$ + $C_{14}$ & 681 & $K=3$ & 120 & 6,720 & 5.66 & -- \\
$D_{14}$ + $C_{14}$ & 681 & $K=4$ & 96 & 6,720 & \textbf{5.56} & -- \\
\midrule
\multicolumn{7}{@{}l}{\textit{Matched effective depth}} \\
\addlinespace[2pt]
$D_{28}$ & 675 & Unguided & 240 & 6,720 & 8.19 & 4.88 \\
$D_7$ + $C_7$ & 346 & $K=3$ & 240 & 6,720 & \textbf{7.51} & \textbf{4.21} \\
\bottomrule
\end{tabular*}
\caption{\textbf{Iterative correction outperforms matched baselines.} Although a same-layer denoiser-corrector pair achieves similar performance with a large monolithic denoiser with a single correction ($D_{14}$ + $C_{14}$, $K=1$), iterative correction ($D_{14}$ + $C_{14}$, $K=3$) outperforms the monolithic baseline. Even the smaller $D_7+C_7$ pair outperforms the larger monolithic baseline with $K=3$. We select the best CFG scale from $w\in\{1,\ldots,10\}$.}
\label{tab:deeper-denoiser}
\label{tab:lightningdit-d7c7-d28}
\endgroup

%% file: generated/iterative_correction_qualitative.tex
\begin{figure}[!h]
    \centering
    \vspace{-15pt}
    \includegraphics[width=0.95\linewidth]{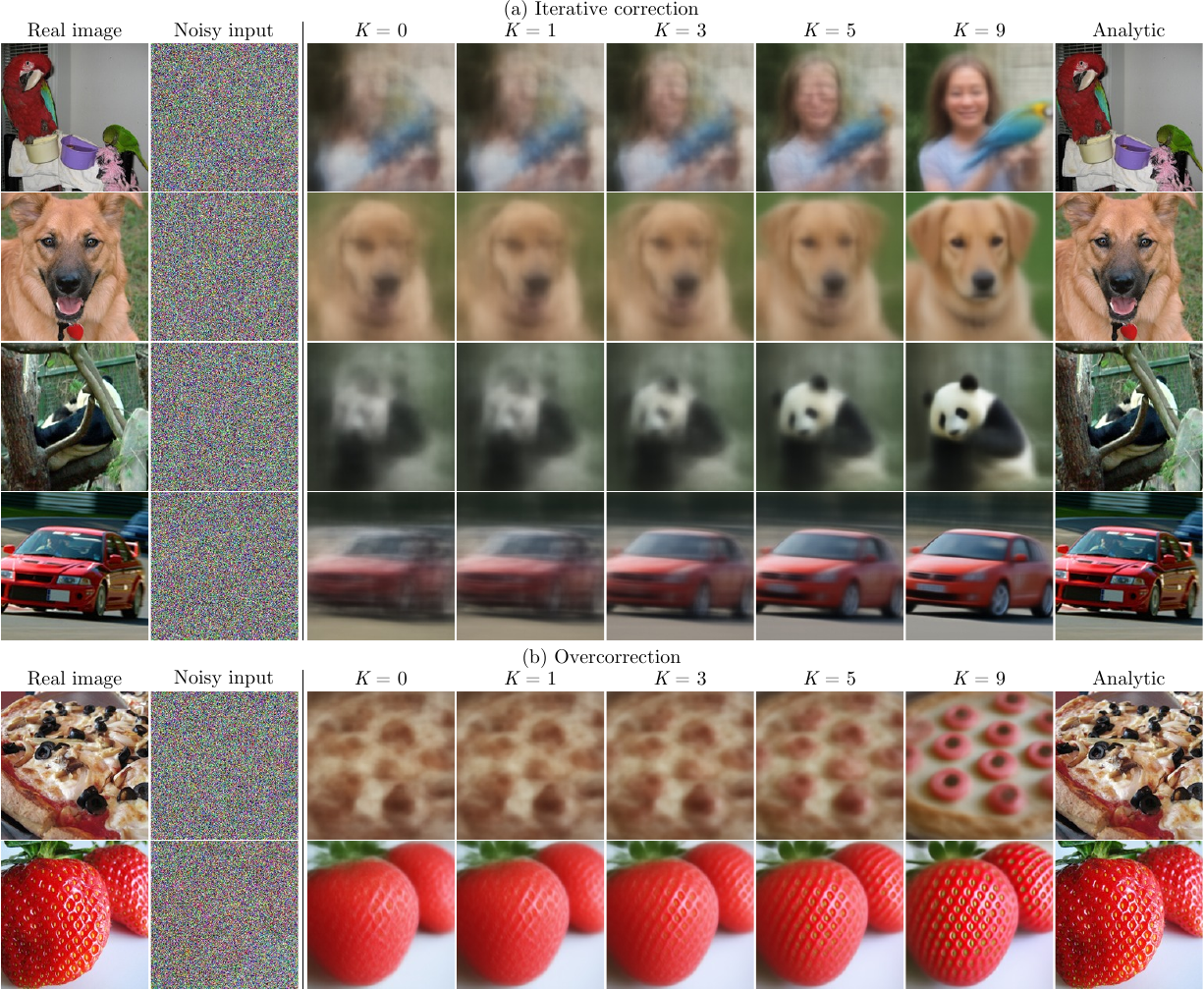}
    \caption{\textbf{Iterative correction sharpens predictions.}
    We show the effect of iterative correction on JiT-L/16.
    \textbf{(a)} The base denoiser's predictions ($K=0$) are often incoherent and contain artifacts. Repeated correction resolves ambiguous structures and makes faces, object contours, and other details more coherent.
    \textbf{(b)} In extreme cases, correction appears to push this coherence too far, turning ambiguous structures in the base prediction into overly regular geometric shapes. 
    Further selected examples appear in Appendix~\ref{app:correction-qualitative}. Analytic target is computed using Eq.~\ref{eq:analytic} in Appendix.}
    \label{fig:repeated-correction}
    \label{fig:refined-samples}
\vspace{-10pt}
\end{figure}

%% file: generated/correction_training_loss.tex
\begin{wrapfigure}[10]{r}{0.31\textwidth}\vspace{-6pt}
    \centering
    \vspace{-12pt}
    \captionsetup{font=footnotesize,skip=3pt,justification=raggedright,singlelinecheck=false}
    \includegraphics[height=1.45in,width=1.0\linewidth,keepaspectratio]{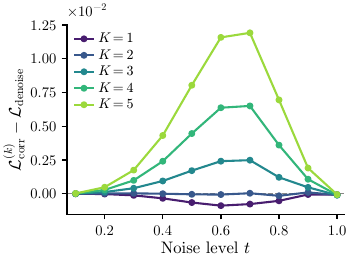}
    \caption{\textbf{Iterative correction can increase denoising error.}}
    \label{fig:increased-training-loss}
\vspace{-16pt}
\end{wrapfigure}

%% file: generated/training_inference_gap.tex
\begin{wrapfigure}{r}{0.42\textwidth}
    \centering
    \vspace{-11pt}
    \includegraphics[width=\linewidth]{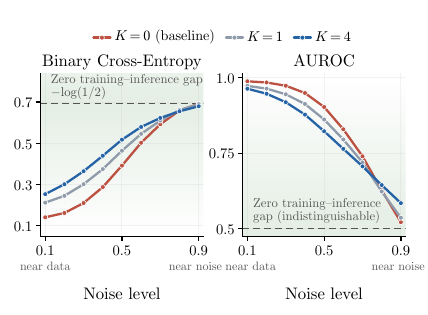}
    \vspace{-15pt}
    \caption{\textbf{Measuring the training--inference gap with classifier two-sample tests.} We train a classifier to distinguish samples from the training distribution and the inference distribution, and report AUROC and binary cross-entropy (BCE). Overall, correction makes classification harder, indicating a smaller gap between training and inference.
    \vspace{-15pt}
    }
    \label{fig:classification-test}
\end{wrapfigure}

%% file: generated/ablation_single_multi.tex
\begin{figure}[!htbp]
    \centering
    \captionsetup{font=footnotesize,skip=3pt}
    \includegraphics[width=\linewidth]{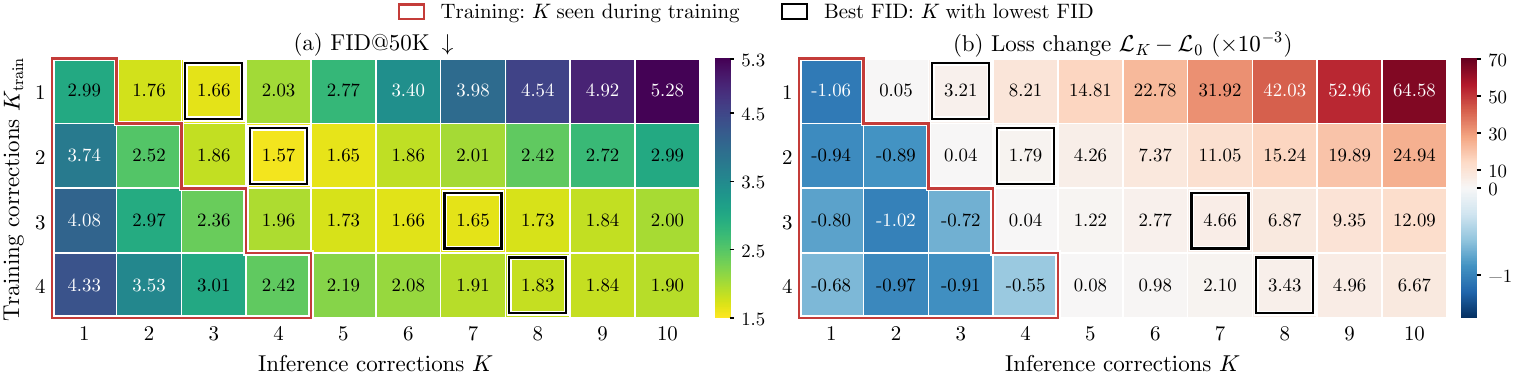}
    \caption{\textbf{Correction beyond the supervised training depth improves generation (lower FID).} Supervising more correction steps shifts the best-performing tested inference correction count $K$ toward larger values, often approximately twice the maximum supervised depth. This suggests that ``extrapolating'' the learned correction beyond its training inputs may play an important role in the observed gains.
    }
    \label{fig:repeated-correction-graph}
    \label{fig:ablation-single-multi}
\par
\end{figure}

%% file: generated/ablation_freeze_samples.tex
\begin{figure}[!t]
    \vspace{-6pt}
    \centering
    \includegraphics[width=\linewidth]{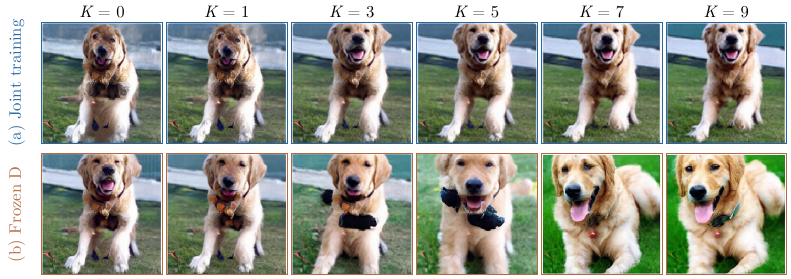}
    \caption{\textbf{A frozen denoiser can support high quality refinement, but iterative correction can also oversaturate samples and produce abrupt changes in the sample prediction.}}
    \label{fig:ablation-freeze-samples}
    \vspace{-10pt}
\end{figure}

%% file: generated/joint_frozen_curves_wrap.tex
\begin{wrapfigure}{r}{0.34\textwidth}
    \centering
    \vspace{-8pt}
    \captionsetup{font=footnotesize,skip=3pt,justification=raggedright,singlelinecheck=false}
    \includegraphics[width=\linewidth]{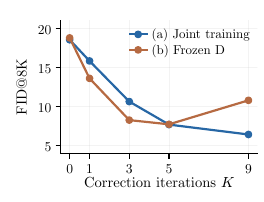}
    \caption{\textbf{Frozen-denoiser correction is competitive at small $K$ but deteriorates earlier.}}
    \label{fig:ablation-freeze-curves}
    \vspace{-8pt}
\end{wrapfigure}

%% file: sections/related_works.tex
\vspace{-2pt}
\section{Related work}
\vspace{-2pt}
Stacked generalization~\citep{wolpert1992stacked} trains a higher-level predictor on held-out predictions of base models to improve generalization. In diffusion models, self-conditioning~\citep{chen2022analog} uses previous clean-data predictions as additional denoiser inputs, and recent concurrent works extend this idea to iterative refinement in language modeling and structured reasoning~\citep{yoo2026self,helbling2026flow}. We train an explicit corrector on denoiser predictions and reuse it across correction iterations, without requiring the original noisy input. Idempotent Generative Networks~\citep{shocher2024idempotent} also learn iterative mappings through an explicit idempotency objective. Our method retains the denoising objective and applies correction at each noise level to refine predictions before the sampling update. The most relevant work to ours is \citet{schiff2026learn}, which trains a corrector on generated samples and applies it repeatedly during sampling in discrete diffusion for language modeling.

\vspace{-2pt}
\section{Conclusion} 
\vspace{-2pt}
We introduce correction training, which learns to correct a diffusion model's predictions using the original clean-data targets. Despite being trained for a single correction, the corrector supports repeated refinement at inference. Experiments on class-conditional and unconditional ImageNet generation and text-to-image generation demonstrate substantial improvements in sample quality. Correction can achieve competitive quality with guidance while better preserving diversity. Correction can also be distilled to remove inference overhead. Our analysis offers empirical insights into why correction works, but a theoretical understanding of its effectiveness and iterative behavior remains an open question for future research.

%% file: sections/appendix.tex
\clearpage
\appendix
\raggedbottom
\setlength{\textfloatsep}{10pt plus 2pt minus 2pt}
\setlength{\floatsep}{8pt plus 2pt minus 2pt}
\setlength{\intextsep}{8pt plus 2pt minus 2pt}
\makeatletter
\setlength{\@fptop}{0pt}
\setlength{\@fpsep}{10pt plus 2pt minus 2pt}
\makeatother

\section*{Appendix Contents}
\startcontents[appendix]
\printcontents[appendix]{}{1}{\setcounter{tocdepth}{2}}

\input{sections/appendix/ablation_study}

\FloatBarrier
\input{sections/appendix/implementation_details}

\input{sections/appendix/guidance_comparison}
\FloatBarrier
\input{sections/appendix/CFG}

\FloatBarrier
\input{sections/appendix/discussion}

\FloatBarrier
\input{sections/more_samples}

\clearpage

\input{generated/diffusionbench_unconditional}

%% file: sections/appendix/ablation_study.tex
\clearpage
\section{Design choices and alternative realizations}
\label{app:design-choices}

\paragraph{A practical realization of correction training.}
In this section, we examine the design choices in our framework. The default configuration used in the main paper is a simple implementation compatible with pretrained models: we use a separate corrector initialized from the base denoiser, omit the noisy input $\bm{z}_t$, retain the timestep-conditioning pathway to preserve the original architecture, and apply stop-gradient to prevent the correction loss from updating the denoiser. Many other combinations of these design choices are possible within our correction training framework. Instead of exhaustively evaluating all combinations, we vary one component at a time from the default configuration to understand how each choice affects performance.

\subsection{Corrector interface}
\label{app:corrector-interface}

\paragraph{Access to the noisy state $\bm{z}_t$ from corrector.}
\label{app:xt-conditioning}
Prior approaches~\citep{chen2022analog} to learning from model predictions often also provide the original input. In our diffusion setting, this corresponds to giving the corrector access to $\bm{z}_t$ in addition to the denoiser's prediction. Tab.~\ref{tab:k-ladder-xt} shows that, on JiT, correction without $\bm{z}_t$ achieves comparable FID to correction with $\bm{z}_t$, while requiring no architectural changes to the denoiser.

The SiT comparison in Fig.~\ref{fig:xt-conditioning-pr} further shows how this choice affects the precision--recall trade-off. Here, $p$-only denotes conditioning on the proposal $\hat{\bm{x}}$ alone, whereas $p+\bm{z}_t$ provides both the proposal and the noisy input. Proposal-only correction achieves higher precision at slightly lower recall. One possible explanation is that, without access to the original noisy input, the corrector is less constrained by it and can modify the proposal more aggressively.
\par
\input{tables/k_heatmap_xt}
\input{tables/ablation_xt_figure}
\paragraph{Timestep conditioning.}
By default, we provide the true timestep to reuse the pretrained model's timestep-conditioning pathway. However, Tab.~\ref{tab:ablation-timestep} shows that using the true timestep or a constant time embedding yields similar final FID in our residual-correction setting. This finding may not generalize to other settings, including correctors trained from scratch: the pretrained pathway associates $t$ with the noise level of the original noisy input, which does not directly describe the errors in the clean-sample prediction received by the corrector.
\par
\begin{figure}[!hb]
    \centering
    \includegraphics[width=\linewidth]{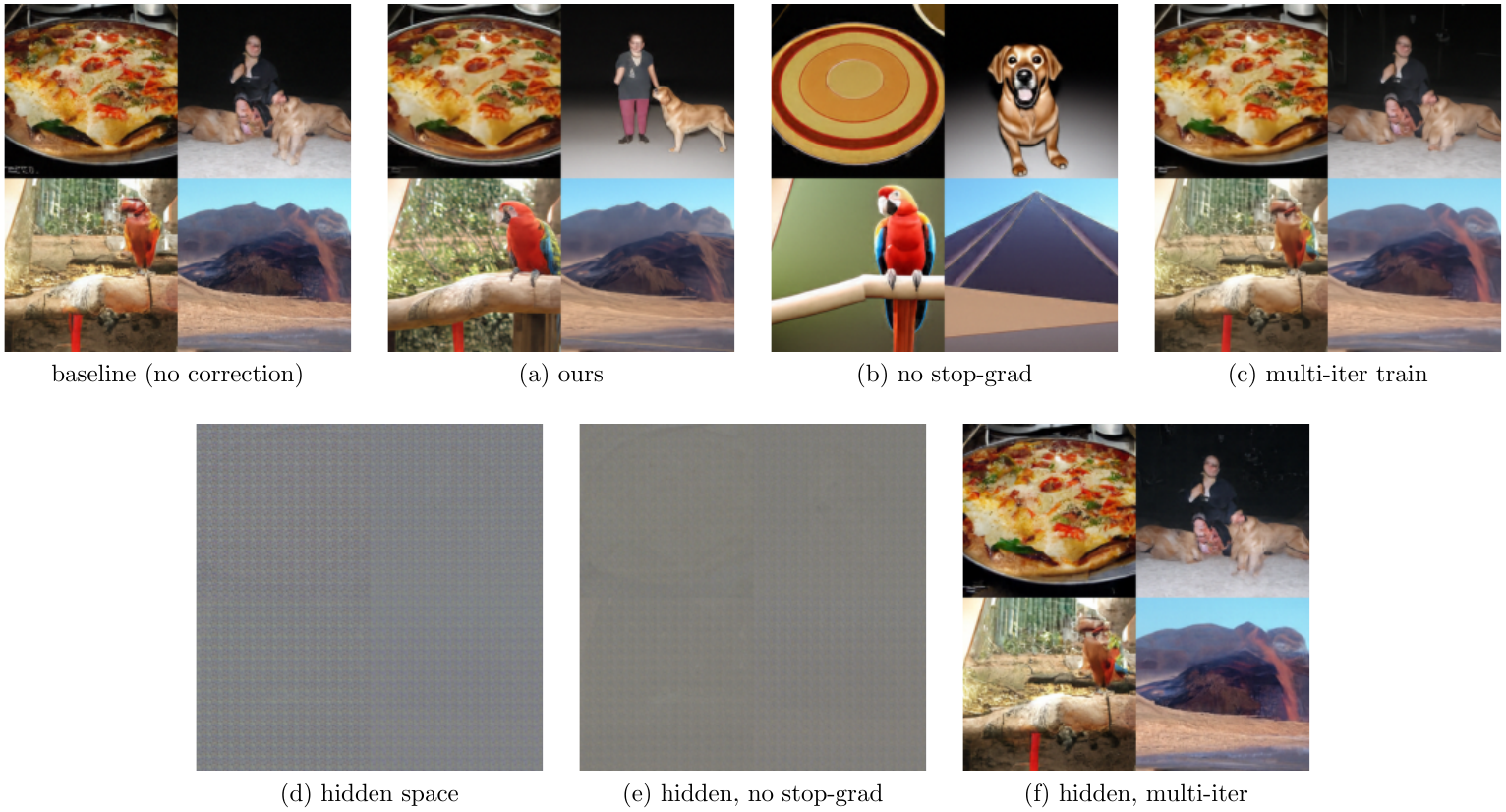}
    \caption{\textbf{Output-space correction remains effective under repeated application.} Qualitative comparisons of the component ablations in Tab.~\ref{tab:ablations} at $K=5$, using the same initial noise and class labels.}
    \label{fig:ablation-supercells}
\end{figure}

\input{generated/ablation_timestep}
\paragraph{Prediction-space versus hidden-state refinement.}
We compare correction in output space (pixels or latents) with refinement of the network's hidden states. Output-space correction benefits from repeated application, whereas the hidden-state variants do not achieve comparable gains (Tab.~\ref{tab:ablations}). Increasing the number of correction iterations during training does not improve sample quality in either setting. Fig.~\ref{fig:ablation-supercells} presents qualitative comparisons at $K=5$.

On the other hand, removing stop-gradient from the base denoiser's prediction achieves a lower best FID, but retaining it yields more consistent gains at larger $K$. Without stop-gradient, samples become visually simpler under repeated correction, as shown in Fig.~\ref{fig:ablation-supercells}(b).
\par

\input{tables/k_heatmap_ablation}

\subsection{Training configuration}
\label{app:correction-training-choices}

\paragraph{Stop-gradient versus end-to-end optimization.}
\label{app:stop-gradient}
\par
We further evaluate the effect of stop-gradient on additional pretrained models (Fig.~\ref{fig:ablation-gradient-paths}). Unlike the earlier comparison in Table~\ref{tab:ablations}, the variants achieve similar best FID here. Full stop-gradient (our default) nevertheless leads to less degradation at larger correction counts $K$. \input{generated/ablation_gradient_curves}
\subsection{Inference configuration}
\label{app:correction-inference-choices}

\paragraph{Correction count $K$.}
Increasing $K$ improves FID up to a setting-dependent optimum, after which further correction can degrade it (Tab.~\ref{tab:ablations}; Fig.~\ref{fig:ablation-gradient-paths}).

\paragraph{Correction interval.}
\label{app:correction-interval}
Restricting JiT correction to early steps and reallocating the saved calls to denoising improves FID at a fixed sampling budget (Tab.~\ref{tab:correction-interval}).
\par
\input{generated/jit_interval_heatmap.tex}
\input{generated/jit_interval_samples.tex}

%% file: tables/k_heatmap_xt.tex
\begin{table}[!htb]
    \centering
    \begingroup
    \small
    \setlength{\tabcolsep}{5pt}
    \begin{tabular}{@{}lcccccc@{}}
        \toprule
        Variant & $K{=}0$ & $K{=}1$ & $K{=}2$ & $K{=}3$ & $K{=}4$ & $K{=}5$ \\
        \midrule
        Proposal only ($\hat{x}$) & \cellcolor{heatgreen!20}24.63 & \cellcolor{heatgreen!33}17.96 & \cellcolor{heatgreen!43}14.07 & \cellcolor{heatgreen!45}\textbf{13.35} & \cellcolor{heatgreen!43}14.27 & \cellcolor{heatgreen!37}16.06 \\
        With $\bm{z}_t$ conditioning & \cellcolor{heatgreen!20}24.72 & \cellcolor{heatgreen!32}18.22 & \cellcolor{heatgreen!41}14.84 & \cellcolor{heatgreen!44}13.79 & \cellcolor{heatgreen!42}14.67 & \cellcolor{heatgreen!36}16.58 \\
        \bottomrule
    \end{tabular}
    \endgroup
\caption{\textbf{Proposal-only correction performs comparably to $\bm{z}_t$-conditioned correction.} Unguided FID50K on JiT-B/16 at approximately 250 NFE.}
    \label{tab:k-ladder-xt}
\end{table}

%% file: tables/ablation_xt_figure.tex
\begin{figure}[!htb]
    \centering
    \includegraphics[width=\linewidth]{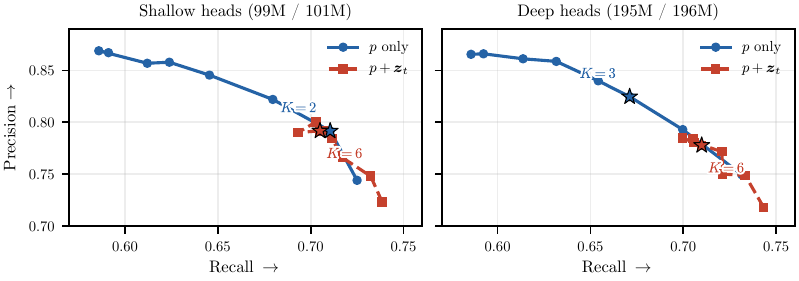}
    \caption{\textbf{Proposal-only correction reaches higher peak precision.} SiT on ImageNet with approximately capacity-matched correction heads, sweeping $K=1$--$8$. Stars mark each arm's lowest-FID setting.}
    \label{fig:xt-conditioning-pr}
\end{figure}

%% file: generated/ablation_timestep.tex
\begin{table}[!htbp]
\centering\small
\begin{tabular}{@{}lrrr@{}}
\toprule
Corrector time input & FID $\downarrow$ & Precision $\uparrow$ & Recall $\uparrow$ \\
\midrule
True timestep & 2.825 & 0.763 & 0.721 \\
Constant time embedding & 2.895 & 0.745 & 0.734 \\
\bottomrule
\end{tabular}
\caption{\textbf{Explicit time conditioning has little effect on FID in this setting.} We compare true and fixed timestep inputs to the corrector on SiT-REPA at $K=1$, using FID-50K at 250 NFE. The denoiser uses the true timestep in both cases.}
\label{tab:ablation-timestep}
\end{table}

%% file: tables/k_heatmap_ablation.tex
\begin{table}[!htb]
    \centering
    \begingroup
    \small
    \setlength{\tabcolsep}{4pt}
    \renewcommand{\arraystretch}{1.08}
    \begin{tabular}{@{}ccrrrrrr@{}}
        \toprule
        \multirow{2}{*}{\shortstack{Stop-\\gradient}}
        & \multirow{2}{*}{$K_{\mathrm{train}}$}
        & \multicolumn{6}{c}{Inference-time iterations $K$} \\
        \cmidrule(lr){3-8}
        & & $0$ & $1$ & $2$ & $3$ & $4$ & $5$ \\
        \midrule
    
        \multicolumn{8}{@{}l}{\textit{Correcting model predictions $\hat{x}$}} \\[1pt]
    
        \cmark & $1$
        & \cellcolor{heatgreen!16}24.63
        & \cellcolor{heatgreen!25}17.96
        & \cellcolor{heatgreen!33}14.07
        & \cellcolor{heatgreen!34}13.35
        & \cellcolor{heatgreen!33}14.27
        & \cellcolor{heatgreen!29}16.06 \\
    
        \xmark & $1$
        & \cellcolor{heatgreen!10}29.08
        & \cellcolor{heatgreen!26}17.38
        & \cellcolor{heatgreen!36}12.88
        & \cellcolor{heatgreen!23}19.25
        & \cellcolor{heatgreen!16}24.64
        & \cellcolor{heatgreen!10}28.92 \\
    
        \cmark & $1$--$5$
        & \cellcolor{heatgreen!15}24.66
        & \cellcolor{heatgreen!20}21.02
        & \cellcolor{heatgreen!20}21.68
        & \cellcolor{heatgreen!18}22.51
        & \cellcolor{heatgreen!17}23.30
        & \cellcolor{heatgreen!15}24.67 \\
    
        \addlinespace[3pt]
        \multicolumn{8}{@{}l}{\textit{Iterating over hidden states $h$}} \\[1pt]
    
        \cmark & $1$
        & \cellcolor{heatgreen!16}24.63
        & \cellcolor{heatgreen!15}24.78
        & \cellcolor{heatgreen!3}35.73
        & 102.84 & 261.61 & 339.93 \\
    
        \xmark & $1$
        & \cellcolor{heatgreen!16}24.19
        & \cellcolor{heatgreen!16}24.63
        & 39.85 & 104.94 & 233.81 & 326.03 \\
    
        \cmark & $1$--$5$
        & \cellcolor{heatgreen!15}24.66
        & \cellcolor{heatgreen!15}24.99
        & \cellcolor{heatgreen!15}25.01
        & \cellcolor{heatgreen!15}25.12
        & \cellcolor{heatgreen!14}25.35
        & \cellcolor{heatgreen!14}25.47 \\
    
        \bottomrule
    \end{tabular}
    \endgroup
\caption{\textbf{Output-space iteration enables effective iterative correction.} Component ablations on JiT-B/16, evaluated by unguided FID50K at approximately 250 NFE.}
\label{tab:ablations}
\end{table}

%% file: generated/ablation_gradient_curves.tex
%
\begin{figure}[!htbp]
    \centering
    \includegraphics[width=\linewidth]{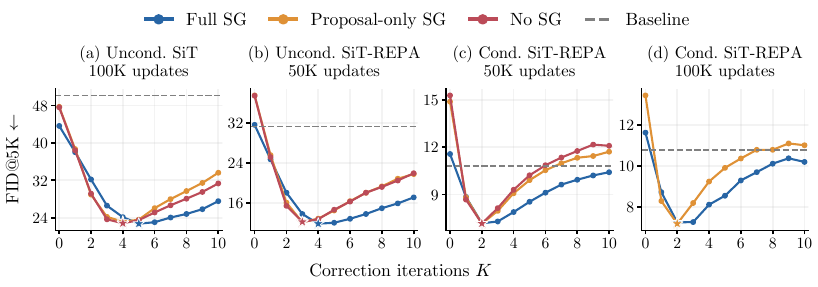}
    \caption{\textbf{Full stop-gradient degrades more slowly at large correction counts in these comparisons.}
    FID@5K versus $K$, using EMA weights, the full correction interval, and at most 250 NFE.
    SG denotes stop-gradient; full SG blocks both proposal and base-combination paths.}
    \label{fig:ablation-gradient-paths}
\end{figure}

%% file: generated/jit_interval_heatmap.tex
\begin{table}[!htbp]
\centering
\begingroup
\small
\definecolor{jitintervalblue}{RGB}{34,104,156}
\setlength{\tabcolsep}{4pt}
\renewcommand{\arraystretch}{1.05}
\begin{tabular}{@{}rccccc@{}}
\toprule
 & \multicolumn{5}{c}{End $r_{\rm end}$} \\
Start $r_{\rm start}$ & $0.8$ & $0.6$ & $0.4$ & $0.2$ & $0.0$ \\
\midrule
$1.0$ & \cellcolor{jitintervalblue!56}7.70 & \cellcolor{jitintervalblue!60}\textbf{6.79} & \cellcolor{jitintervalblue!56}7.72 & \cellcolor{jitintervalblue!51}9.05 & \cellcolor{jitintervalblue!40}11.85\textsuperscript{full} \\
$0.8$ & \textcolor{gray}{---} & \cellcolor{jitintervalblue!43}11.16 & \cellcolor{jitintervalblue!43}11.12 & \cellcolor{jitintervalblue!37}12.59 & \cellcolor{jitintervalblue!33}13.76 \\
$0.6$ & \textcolor{gray}{---} & \textcolor{gray}{---} & \cellcolor{jitintervalblue!30}14.54 & \cellcolor{jitintervalblue!27}15.09 & \cellcolor{jitintervalblue!16}17.94 \\
$0.4$ & \textcolor{gray}{---} & \textcolor{gray}{---} & \textcolor{gray}{---} & \cellcolor{jitintervalblue!20}17.10 & \cellcolor{jitintervalblue!8}20.05 \\
$0.2$ & \textcolor{gray}{---} & \textcolor{gray}{---} & \textcolor{gray}{---} & \textcolor{gray}{---} & \cellcolor{jitintervalblue!8}19.96 \\
\bottomrule
\end{tabular}
\par\smallskip{\scriptsize \textsuperscript{full}Full trajectory ($1\to0$).}
\endgroup
\captionsetup{font=small,skip=4pt}
\caption{\textbf{Allocating correction to early steps yields the lowest FID at a fixed sampling budget.} JiT-L/16 with $K=5$, evaluated by FID8K. Rows and columns specify the start and end noise levels, decreasing from noise to data.}
\label{tab:correction-interval}
\end{table}

%% file: generated/jit_interval_samples.tex
\clearpage
\begin{figure}[!htbp]
\centering
\captionsetup{font=footnotesize,skip=5pt}
{\small (a)}\par\vspace{2pt}
\includegraphics[width=.88\linewidth]{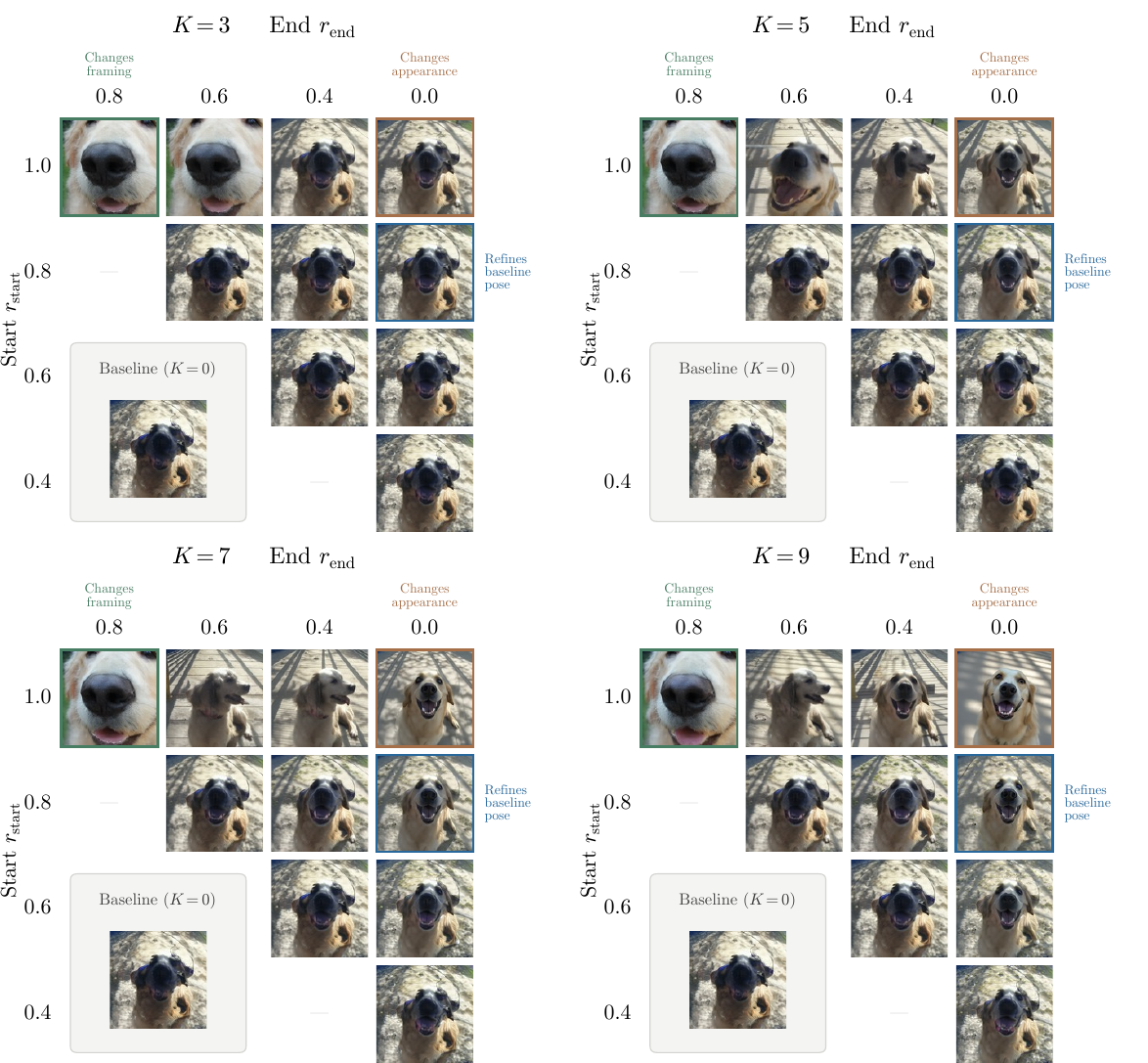}
\par\vspace{5pt}
{\small (b)}\par\vspace{2pt}
\begingroup
\setlength{\tabcolsep}{1.5pt}
\begin{tabular}{@{}cccccc@{}}
\multicolumn{2}{c}{$K=3$} & \multicolumn{2}{c}{$K=5$} & \multicolumn{2}{c}{$K=9$} \\
Full & Selected & Full & Selected & Full & Selected \\
{\scriptsize $1\to 0$} & {\scriptsize $1\to 0.4$} & {\scriptsize $1\to 0$} & {\scriptsize $1\to 0.6$} & {\scriptsize $1\to 0$} & {\scriptsize $1\to 0.8$} \\[2pt]
\includegraphics[width=.145\linewidth]{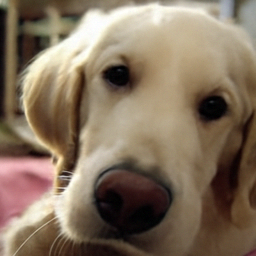} & \includegraphics[width=.145\linewidth]{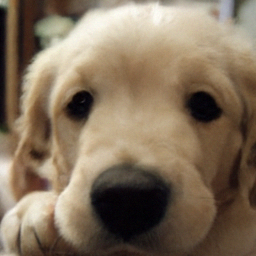} & \includegraphics[width=.145\linewidth]{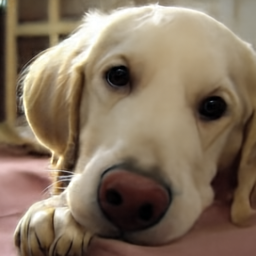} & \includegraphics[width=.145\linewidth]{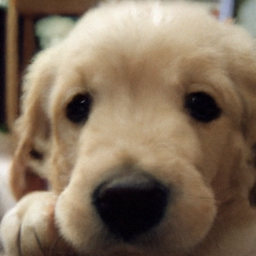} & \includegraphics[width=.145\linewidth]{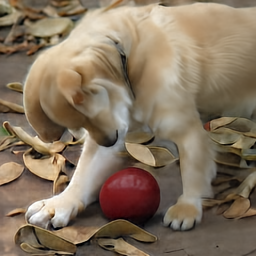} & \includegraphics[width=.145\linewidth]{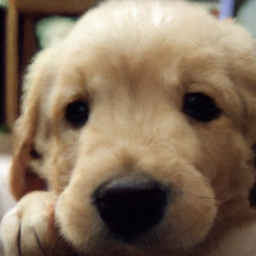} \\[3pt]
\includegraphics[width=.145\linewidth]{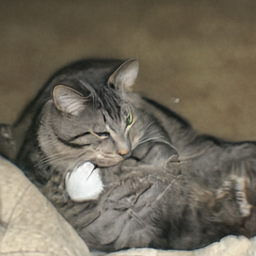} & \includegraphics[width=.145\linewidth]{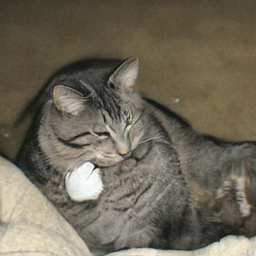} & \includegraphics[width=.145\linewidth]{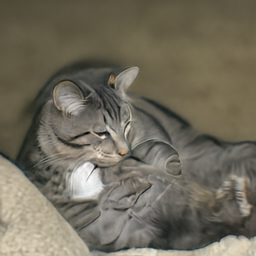} & \includegraphics[width=.145\linewidth]{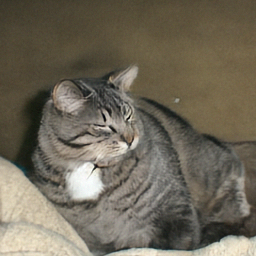} & \includegraphics[width=.145\linewidth]{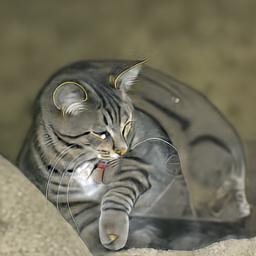} & \includegraphics[width=.145\linewidth]{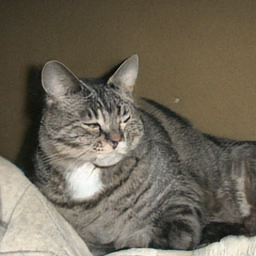} \\[3pt]
\includegraphics[width=.145\linewidth]{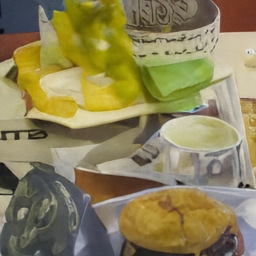} & \includegraphics[width=.145\linewidth]{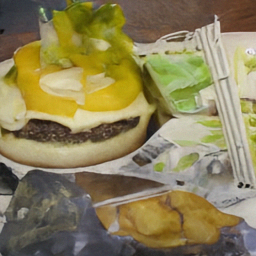} & \includegraphics[width=.145\linewidth]{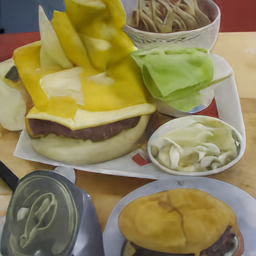} & \includegraphics[width=.145\linewidth]{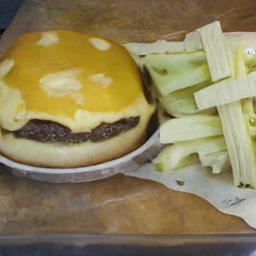} & \includegraphics[width=.145\linewidth]{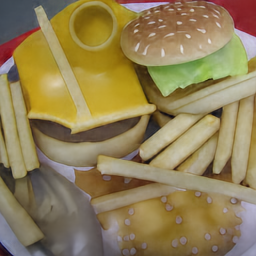} & \includegraphics[width=.145\linewidth]{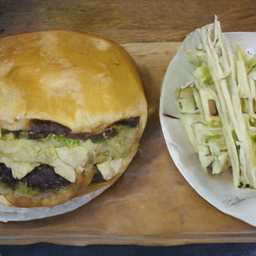} \\[3pt]
\end{tabular}
\endgroup

\caption[]{\textbf{Skipping early correction refines the baseline pose.} Gray boxes show the matched $K=0$ baseline; green, blue and orange mark early-only, skip-early and full-range correction, respectively.
\par\smallskip
\textbf{Restricting correction to early steps changes the generated structure.} Matched-noise comparisons of full-trajectory correction and the interval selected by FID8K for each $K$.}
\makeatletter
\begingroup
\protected@edef\@currentlabel{\thefigure(a)}
\label{fig:correction-interval-dog}
\label{fig:correction-interval}
\label{fig:correction-interval-k3}
\label{fig:correction-interval-k5}
\label{fig:correction-interval-k7}
\label{fig:correction-interval-k9}
\endgroup
\begingroup
\protected@edef\@currentlabel{\thefigure(b)}
\label{fig:correction-interval-overview}
\endgroup
\makeatother
\end{figure}

\clearpage
\begin{figure}[!htbp]
\centering
\captionsetup{font=footnotesize,skip=5pt}
{\small (a)}\par\vspace{2pt}
\includegraphics[width=.73\linewidth]{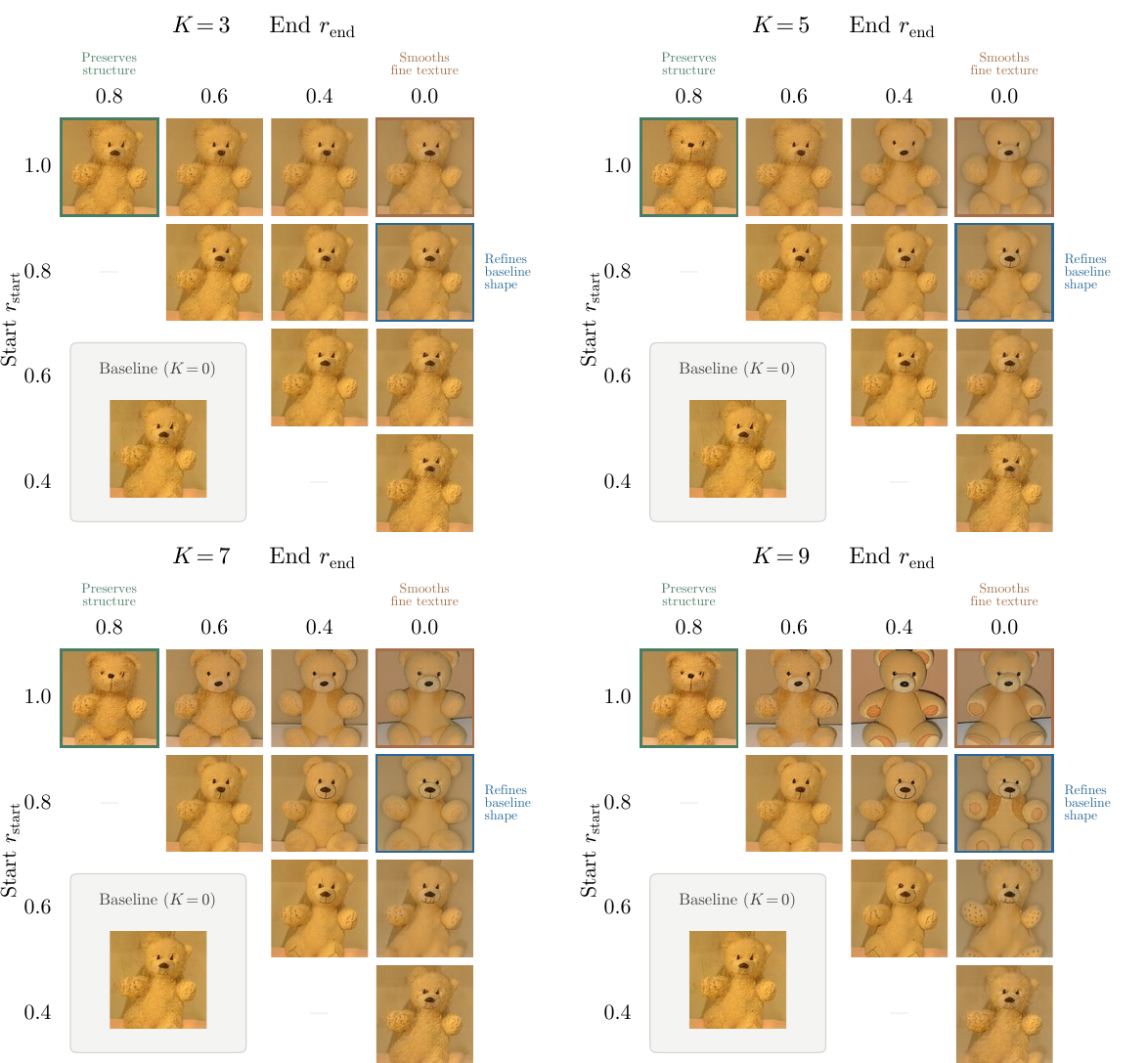}
\par\vspace{5pt}
{\small (b)}\par\vspace{2pt}
\includegraphics[width=.73\linewidth]{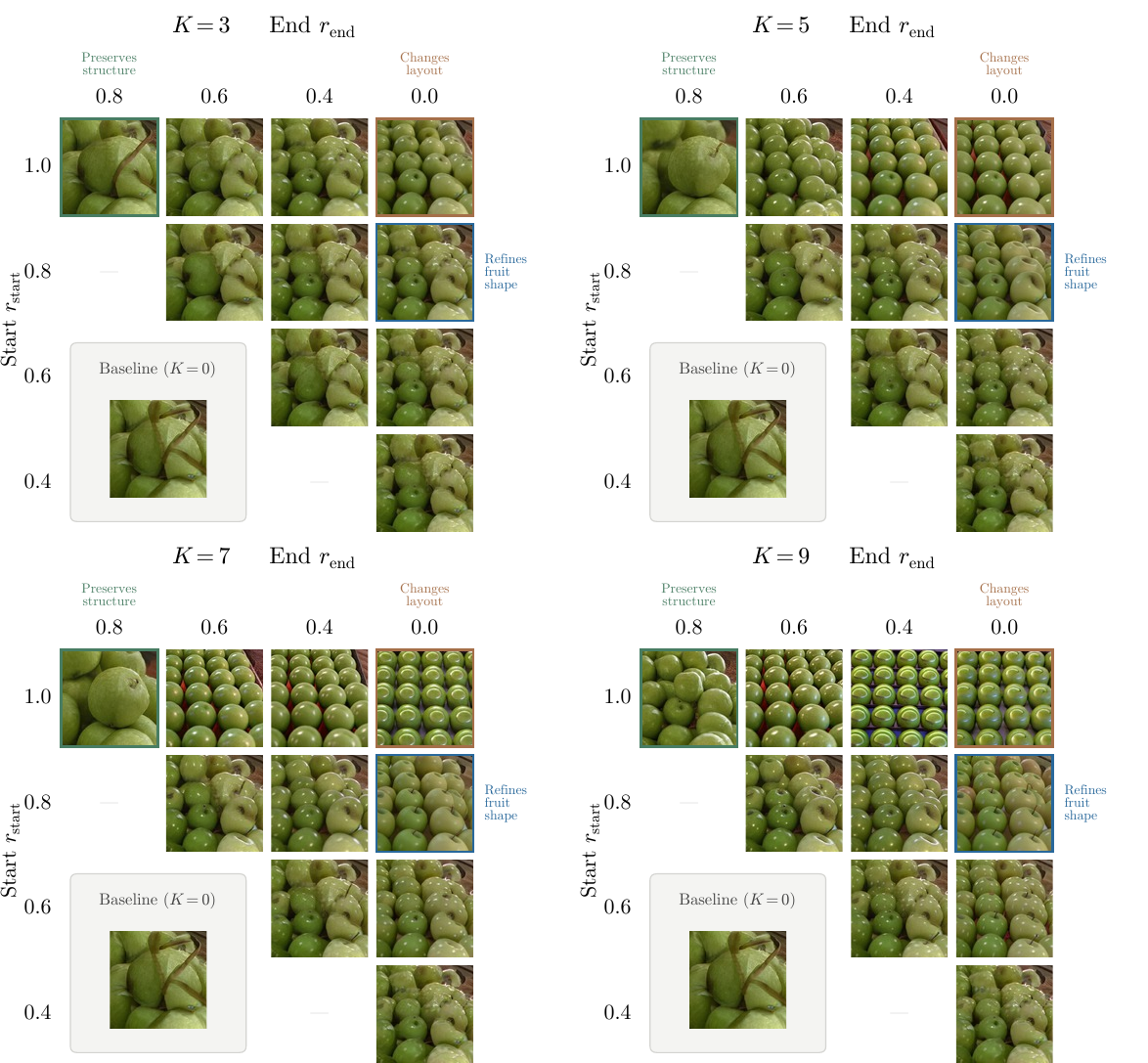}
\caption[]{\textbf{(a) Full-range correction changes fine texture and shape. (b) Full-range correction changes object layout and appearance.} Gray boxes show the matched $K=0$ baseline; green, blue and orange mark early-only, skip-early and full-range correction, respectively.}
\makeatletter
\begingroup
\protected@edef\@currentlabel{\thefigure(a)}
\label{fig:correction-interval-teddy}
\endgroup
\begingroup
\protected@edef\@currentlabel{\thefigure(b)}
\label{fig:correction-interval-apples}
\endgroup
\makeatother
\end{figure}
\par\ignorespaces

%% file: sections/appendix/implementation_details.tex
\newpage
\section{Implementation Details}
\label{app:imp}
\subsection{Training configurations}
\label{app:training-config}
\paragraph{Common backbone settings.} We retain each pretrained model's backbone and original conditioning. The separate corrector uses an independent copy of the denoiser backbone; latent-space models retain their original VAE, while JiT operates directly on RGB images. Training and quantitative sampling settings are summarized in Tables~\ref{tab:main-training-settings} and~\ref{tab:main-settings}.

\input{generated/main_training_settings}
\input{generated/sampling_settings_grouped_preview}

\newpage
\subsection{Corrector parameterization}
\label{app:different-parameterization}

Algorithm~\ref{alg:training} uses $x$-prediction. For an $\epsilon$- or $v$-predicting denoiser, we recommend the following two practical options.

\paragraph{Direct clean-sample prediction.}
The simplest starting point is to convert the denoiser's output into
a clean-sample proposal and train the corrector to predict a refined
clean sample:
\[
\hat{\bm{x}}^{(1)}
=C_{\bm{\phi}}\!\left(
\operatorname{sg}(\hat{\bm{x}}^{(0)}),t
\right).
\]
The result can then be converted into the representation required by
the sampler using $\bm{z}_t$, $t$, and the noise schedule.
These conversions are performed outside the corrector network;
$\bm{z}_t$ need not be supplied as an additional network input.
The training loss can still be computed in velocity or noise space.

Why convert the proposal to $x$ rather than retain the denoiser's
original output space?
In our small-scale experiments with a single correction ($K=1$)
and without additional noisy-state conditioning, $x$-space correction
performs best among the tested same-space input--output choices,
including with $\epsilon$- and $v$-predicting denoisers.
This makes clean-sample input and output a useful starting point,
especially when training the corrector from scratch.
These experiments do not establish a universally optimal
parameterization since it's small-scale run on $K=1$.  Directly correcting noise or velocity
remains an option.

\paragraph{Reusing pretrained output representations.}
When initializing the corrector from a pretrained denoiser, another
consideration is how much of its learned representation to retain.
For example, SiT~\citep{ma2024sit} is trained to predict velocity.
Changing its output to $x$ requires adapting the pretrained output
representation.
Direct $x$-prediction can still work and converged quickly in our
experiments, but retaining the original output parameterization
offers another way to preserve the representation of  pretrained models.
This motivated our use of clean-sample input with native
$\epsilon$- or $v$-output in the main pretrained-model experiments.

Retaining the native output also raises a practical question:
how should it be combined with a clean-sample input?
An exact conversion between $x$, $\epsilon$, and $v$ requires
$\bm{z}_t$ or another component in addition to the quantity being
converted.
Providing $\bm{z}_t$ as an additional condition is possible, but
its empirical benefits were mixed
(App.~\ref{app:xt-conditioning}), so we omit it by default.
Instead, we retain the base prediction in its native output space
and linearly combine it with the corrector's native-space output through
a learned gate.
This allows us to use a clean-sample input while retaining the
pretrained output space, without supplying $\bm{z}_t$ to the network
or requiring an explicit conversion inside it.

Let $\bm{q}$ denote the target in the chosen output space and
$\hat{\bm{q}}^{(0)}$ the base prediction.
Our full stop-gradient formulation is:
\begin{align}
\hat{\bm{q}}^{(1)}
&=\operatorname{sg}(\hat{\bm{q}}^{(0)})
+\bm{\gamma}\odot
\left[
C_{\bm{\phi}}^{q}\!\left(
\operatorname{sg}(\hat{\bm{x}}^{(0)}),t
\right)
-\operatorname{sg}(\hat{\bm{q}}^{(0)})
\right],
\label{eq:gated-correction}\\
\mathcal{L}_{\mathrm{corr}}
&=\mathbb{E}_{\bm{x},\bm{\epsilon},t}\!\left[
\lambda_q(t)
\left\|\hat{\bm{q}}^{(1)}-\bm{q}\right\|_2^2
\right],
\label{eq:gated-correction-loss}
\end{align}
where $\bm{q}$ is the target in the chosen output space, such as $v$ for SiT or $\epsilon$ for DiT, and $\lambda_q(t)$ preserves the model's loss weighting. The unconstrained gate is initialized to zero, preserving the base prediction at initialization. It is learned per channel, except for DiT and plain SiT, which use a scalar shared across channels. For better performance, one can extend it to timestep-dependent gating or 2D tensor gates, but we keep the (per-channel) scalar for simplicity.

\paragraph{Correction with a fixed base component.}
Gating is not the only way to combine different input and output
parameterizations.
We also tested a formulation that combines the corrector's output
with a fixed component of the base prediction.
Under our convention $v=\epsilon-x$, an $x$-space prediction is combined
with the base noise prediction, while an $\epsilon$-space prediction is
combined with the base clean-sample prediction:
\[
\hat{\bm{x}}^{(1)}
=\begin{cases}
\hat{\bm{x}},\\
\hat{\bm{\epsilon}}-\operatorname{sg}(\hat{\bm{v}}^{(0)}),\\
\operatorname{sg}(\hat{\bm{\epsilon}}^{(0)})-\hat{\bm{v}},
\end{cases}
\quad
\hat{\bm{\epsilon}}^{(1)}
=\begin{cases}
\hat{\bm{x}}+\operatorname{sg}(\hat{\bm{v}}^{(0)}),\\
\hat{\bm{\epsilon}},\\
\operatorname{sg}(\hat{\bm{x}}^{(0)})+\hat{\bm{v}},
\end{cases}
\quad
\hat{\bm{v}}^{(1)}
=\begin{cases}
\operatorname{sg}(\hat{\bm{\epsilon}}^{(0)})-\hat{\bm{x}},\\
\hat{\bm{\epsilon}}-\operatorname{sg}(\hat{\bm{x}}^{(0)}),\\
\hat{\bm{v}}.
\end{cases}
\]
These expressions illustrate alternative fixed-base-component
formulations for each output space.
Our small-scale experiments used the velocity-space formulation
on the right and minimized velocity-space MSE.

Here $\hat{\bm{x}}$, $\hat{\bm{\epsilon}}$, and $\hat{\bm{v}}$
denote the corrector's predictions.
The input can independently be chosen from the base clean, noise,
or velocity predictions.
All combinations minimize velocity-space MSE without a learned gate.
For $x$- and $\epsilon$-space predictions, this objective also requires
compensating for errors in the retained base component, so these predictions
are not directly supervised against the clean sample or noise alone.

By explicitly using the decomposition $v=\epsilon-x$, this formulation
may offer a more principled way to connect different input and output
parameterizations than a learned blend.
Our exploration was limited by time constraints, however:
we tested it only in small-scale experiments with a single correction
and did not extensively investigate repeated correction or larger-scale
settings.

\paragraph{Gradient paths.}
The correction loss can reach the denoiser through the
corrector's input $\hat{\bm{x}}^{(0)}$ and, when used,
the retained base prediction $\hat{\bm{q}}^{(0)}$.
Our default, \textbf{Full stop-gradient (SG)}, detaches both
(Eq.~\ref{eq:gated-correction}).
For separate networks, the denoiser is then updated only by
its denoising loss; it is not frozen.
We also compare \textbf{Proposal-only SG}, which detaches
only the corrector's input, and \textbf{No SG}, which
detaches neither.
Full SG achieves competitive FID and degrades less at
large correction counts in our comparisons
(App.~\ref{app:stop-gradient}).

\paragraph{Full-target versus residual prediction.}
Independently of the input and output spaces, the network can predict
either the full target or a residual relative to the base prediction.
This choice applies both to direct clean-sample prediction and to
native-space correction with gating.
Without gating, residual prediction takes the form:
\[
\hat{\bm{q}}^{(1)}
=\operatorname{sg}(\hat{\bm{q}}^{(0)})
+R_{\bm{\phi}}^{q}\!\left(
\operatorname{sg}(\hat{\bm{x}}^{(0)}),t
\right).
\]
With gating, the gate can instead scale the predicted residual.
Thus, residual prediction is not an alternative to gating;
the two can be used together.
The correction objective remains unchanged.
Zero-initializing the residual output projection preserves the base
prediction at initialization.
Although we do not present a dedicated comparison in this paper,
residual prediction also worked well in our experiments, sometimes
slightly better than full-target prediction.
We use full-target prediction in the main formulation for simplicity,
but residual prediction is another practical option within
correction training.

\paragraph{Choosing a formulation.}
For a first implementation, particularly when training from scratch,
we recommend clean-sample input and direct clean-sample output.
When reusing a pretrained denoiser, clean-sample input with gated
native-space prediction is another practical starting point.
We retained this form in the main pretrained-model experiments for
compatibility with the pretrained output representation and its
slightly better empirical performance.
Neither choice is required by correction training:
input space, output space, full-target versus residual prediction,
and loss space are distinct design choices.
Other combinations can also be used, provided that the information
needed for any conversion is available.
The network can also take noise or velocity proposals as input;
we use clean-sample input in the equations to illustrate our default.
Full-target versus residual prediction can be chosen independently.

\begin{table}[!htbp]
    \centering
    \small
    \setlength{\tabcolsep}{5pt}
    \renewcommand{\arraystretch}{1.15}
    \begin{tabular}{@{}>{\raggedright\arraybackslash}p{0.23\linewidth}>{\raggedright\arraybackslash}p{0.23\linewidth}>{\raggedright\arraybackslash}p{0.48\linewidth}@{}}
        \toprule
        Design choice & Option & Used in \\
        \midrule
        Output formulation & Direct $x$ & Table~\ref{tab:imagenet-results} --- JiT \\
        & Gated native $q$ & Table~\ref{tab:imagenet-results} --- DiT, SiT, SiT-REPA; Tables~\ref{tab:unconditional}--\ref{tab:diffusionbench-t2i} \\
        \addlinespace[3pt]
        Prediction type & Full target & Tables~\ref{tab:imagenet-results}--\ref{tab:diffusionbench-t2i} \\
        & Residual & Table~\ref{tab:ablation-timestep} --- timestep conditioning \\
        \bottomrule
    \end{tabular}
    \caption{\textbf{Corrector parameterization choices.}
    We distinguish the output formulation from full-target versus
    residual prediction, which can be used with either formulation.
    Here $q$ denotes the chosen output space.
    The equations use clean-sample input by default;
    input parameterization and loss space are separate choices.}
    \label{tab:corrector-parameterization}
\end{table}

\input{sections/figure_table_details}

%% file: generated/main_training_settings.tex
\begin{table}[!htbp]
\centering
\caption{\textbf{Iterative refinement uses correctors trained with a single correction pass.} Training settings for the main ImageNet and quantitative text-to-image comparisons. We follow each model's original training recipe as closely as possible. Updates count additional training, not pretraining; $D$ and $C$ denote separate denoiser and corrector networks. Sampling settings are reported separately.}
\label{tab:main-training-settings}
\begingroup
\footnotesize
\setlength{\tabcolsep}{3pt}
\renewcommand{\arraystretch}{1.16}
\begin{tabular}{@{}>{\raggedright\arraybackslash}p{0.20\linewidth}*{5}{>{\centering\arraybackslash}p{0.144\linewidth}}@{}}
\toprule
Setting & DiT-XL/2 & SiT-XL/2 & SiT$_{\mathrm{REPA}}$-XL/2\textsuperscript{$\dagger$} & JiT-L/16 & T2I\textsuperscript{$\ddagger$} (Sec.~\ref{subsec:text-to-image-generation})\\
\midrule
\multicolumn{6}{@{}l}{\textit{Data and initialization}} \\[-1pt]
Training data & ImageNet & ImageNet & ImageNet & ImageNet & BLIP3o mixture \\
Image resolution & \multicolumn{5}{c}{$256\times256$} \\
Denoising space & SD-VAE latents & SD-VAE latents & SD-VAE latents & RGB pixels & FLUX.2-VAE latents \\
Denoiser init. & Released & Released & Released & Released & 202K base \\
Corrector init. & \multicolumn{5}{c}{Independent copy of the pretrained denoiser} \\
Additional updates & 100K & 100K & 50K & 100K & 50K \\
\midrule
\multicolumn{6}{@{}l}{\textit{Optimization}} \\[-1pt]
Global batch size & 256 & 256 & 256 & 1,024 & 1,024 \\
Optimizer & \multicolumn{5}{c}{AdamW; weight decay $0$} \\
Adam $(\beta_1,\beta_2)$ & $(0.9,0.999)$ & $(0.9,0.999)$ & $(0.9,0.999)$ & $(0.9,0.95)$ & $(0.9,0.95)$ \\
Denoiser LR & $10^{-4}$ & $10^{-4}$ & $10^{-4}$ & $5\!\times\!10^{-5}$ & Linear decay \\
Corrector LR & $10^{-4}$ & $10^{-4}$ & $10^{-4}$ & $5\!\times\!10^{-5}$ & $10^{-4}$ \\
Warmup & None & None & 100 updates & 5 epochs & No new warmup \\
LR after warmup & Constant & Constant & Constant & Constant & $D$: linear; $C$: const. \\
EMA decay & $0.9999$ & $0.9999$ & $0.9999$ & $0.9999 / 0.9996$ & $0.9995$ \\
\midrule
\multicolumn{6}{@{}l}{\textit{Training objective and correction}} \\[-1pt]
Prediction ($D/C$) & $\epsilon/\epsilon$ & $v/v$ & $v/v$ & $x/x$ & $v/v$ \\
Corrector image input & \multicolumn{5}{c}{Detached clean-sample prediction; original timestep and conditioning retained} \\
Output combination & Gated $\epsilon$ & Gated $v$ & Gated $v$ & Direct $x$ & Gated $v$ \\
Gate initialization & $0$ & $0$ & $0$ & --- & $0$ \\
Correction passes & \multicolumn{5}{c}{$K_{\mathrm{train}}=1$} \\
Loss weights ($D:C$) & \multicolumn{5}{c}{$1:1$; correction loss has no gradient path into the separate denoiser} \\
\bottomrule
\end{tabular}
\par\smallskip
\parbox{\linewidth}{\raggedright\scriptsize
\textsuperscript{\(\dagger\)}\SiTREPA{} is initialized from the
\href{https://github.com/sihyun-yu/REPA}{publicly released SiT-XL/2 checkpoint}
trained with the REPA loss for 4M iterations;
the REPA loss is not used during additional training.
}
\parbox{\linewidth}{%
  \raggedright\scriptsize
  \textsuperscript{\(\ddagger\)}T2I denotes text-to-image generation
  (Sec.~\ref{subsec:text-to-image-generation}).
  We use the \href{https://github.com/End2End-Diffusion/diffusion-bench}{DiffusionBench codebase}
  and its \href{https://huggingface.co/diffusion-bench/diffusion-bench/blob/main/t2i/ckpts/t2i-ddt-en28d1152hd72-dn2d2048hd128-flux2-vpred-t4-v0-200k.pt}{T2I checkpoint},
  trained in the FLUX.2 VAE latent space using velocity prediction.\par
}
\endgroup
\end{table}

%% file: generated/sampling_settings_grouped_preview.tex
\begin{table}[p]
\centering
\begingroup
\centering
\caption{\textbf{The main quantitative comparisons use approximately 250 network evaluations per image.} Sampling settings for Tables~\ref{tab:imagenet-results} and~\ref{tab:diffusionbench-t2i}. $K$ is the number of corrections per active step; intervals use noise time $t$ ($1$: noise, $0$: data). NFE includes correction and both CFG branches. Guided T2I applies correction before CFG.}
\label{tab:main-settings}
\label{app:cfg-selection}
\begingroup
\footnotesize
\setlength{\tabcolsep}{4pt}
\renewcommand{\arraystretch}{1.08}
\begin{tabular*}{\linewidth}{@{\extracolsep{\fill}}lcccrrr@{}}
\toprule
Method & CFG $w$ & $K$ & CFG interval & Correction interval & Steps & NFE \\
\midrule
\multicolumn{7}{@{}l}{\textit{DiT-XL/2; DDPM}} \\
Baseline & 1 & 0 & --- & --- & 250 & 250 \\
CFG & 1.5 & 0 & $1\to0$ & --- & 125 & 250 \\
Correction & 1 & 2 & --- & $1\to0$ & 83 & 249 \\
\midrule
\multicolumn{7}{@{}l}{\textit{SiT-XL/2; Euler--Maruyama SDE}} \\
Baseline & 1 & 0 & --- & --- & 250 & 250 \\
CFG & 1.5 & 0 & $1\to0$ & --- & 125 & 250 \\
Correction & 1 & 2 & --- & $1\to0$ & 83 & 249 \\
\midrule
\multicolumn{7}{@{}l}{\textit{SiT$_{\mathrm{REPA}}$-XL/2; Euler--Maruyama SDE}} \\
Baseline & 1 & 0 & --- & --- & 250 & 250 \\
CFG & 1.8 & 0 & $0.7\to0$ & --- & 148 & 250 \\
Correction & 1 & 5 & --- & $0.7\to0.5$ & 123 & 248 \\
\midrule
\multicolumn{7}{@{}l}{\textit{JiT-L/16; Heun ODE (final Euler step)}} \\
Baseline & 1 & 0 & --- & --- & 125 & 249 \\
CFG & 2.4 & 0 & $0<t<0.9$ & --- & 63 & 250 \\
Correction & 1 & 9 & --- & $1\to0.8$ & 45 & 242 \\
\midrule
\multicolumn{7}{@{}l}{\textit{T2I (DiT + DDT); Euler ODE}} \\
Baseline & 1 & 0 & --- & --- & 250 & 250 \\
Correction & 1 & 6 & --- & $1\to0$ & 35 & 245 \\
CFG & 4 & 0 & $1\to0$ & --- & 125 & 250 \\
CFG + Correction & 4 & 2 & $1\to0$ & $1\to0$ & 41 & 246 \\
\bottomrule
\end{tabular*}
\endgroup

\endgroup
\par\vspace{10pt}
\begingroup
    \centering
    \small
    \setlength{\tabcolsep}{3pt}
    \caption{\textbf{Unconditional ImageNet settings.} Sampling settings for Table~\ref{tab:unconditional}, using SiT$_{\mathrm{REPA}}$-XL/2.}
    \label{tab:uncond-settings}
    \begin{tabular*}{\linewidth}{@{\extracolsep{\fill}}lcrrr@{}}
        \toprule
        \textbf{Method} & \textbf{Setting} & \textbf{Train steps} & \textbf{Denoising steps} & \textbf{NFE} \\
        \midrule
        Baseline & -- & \UncondBaseStep & \UncondBaseSamplerSteps & \UncondBaseNFE \\
        Correction & $K=1$ & \UncondCorrectionOneStep & \UncondCorrectionOneSamplerSteps & \UncondCorrectionOneNFE \\
        Correction & $K=\UncondCorrectionBestK$ & \UncondCorrectionBestStep & \UncondCorrectionBestSamplerSteps & \UncondCorrectionBestNFE \\
        \midrule
        \textcolor[HTML]{777777}{Autoguidance} & \textcolor[HTML]{777777}{$w=\UncondAGBestW$} & \textcolor[HTML]{777777}{\UncondBaseStep} & \textcolor[HTML]{777777}{\UncondAGBestSamplerSteps} & \textcolor[HTML]{777777}{\UncondAGBestNFE} \\
        \bottomrule
    \end{tabular*}

\endgroup
\par\vspace{10pt}
\begingroup
\centering
\begingroup
\centering
\fontsize{8.5}{10}\selectfont
\setlength{\tabcolsep}{1pt}
\renewcommand{\arraystretch}{1.12}
\begin{minipage}[t]{0.485\linewidth}
\vspace{0pt}\centering
\textit{With CFG} ($w=2$)\par\vspace{2pt}
\begin{tabular*}{\linewidth}{@{\extracolsep{\fill}}lrr@{}}
\toprule
Method & GenEval $\uparrow$ & DPG-Bench $\uparrow$ \\
\midrule
Baseline & 54.57 & 80.11 \\
\textbf{Correction} ($K=2$) & 61.17 & 80.82 \\
\textbf{Correction} ($K=4$) & 63.61 & \textbf{81.11} \\
\textbf{Correction} ($K=6$) & \textbf{63.74} & 80.79 \\
\bottomrule
\end{tabular*}
\end{minipage}\hfill%
\begin{minipage}[t]{0.485\linewidth}
\vspace{0pt}\centering
\textit{With CFG} ($w=6$)\par\vspace{2pt}
\begin{tabular*}{\linewidth}{@{\extracolsep{\fill}}lrr@{}}
\toprule
Method & GenEval $\uparrow$ & DPG-Bench $\uparrow$ \\
\midrule
Baseline & 71.70 & 84.44 \\
\textbf{Correction} ($K=2$) & \textbf{72.65} & \textbf{84.55} \\
\textbf{Correction} ($K=4$) & 71.52 & 84.38 \\
\textbf{Correction} ($K=6$) & 69.53 & 83.42 \\
\bottomrule
\end{tabular*}
\end{minipage}
\par
\captionsetup{font=footnotesize,skip=5pt,justification=raggedright,singlelinecheck=false}
\caption{\textbf{Correction's alignment gains depend on guidance strength and iteration count.} Additional guidance scales use the same checkpoint as Table~\ref{tab:diffusionbench-t2i}, but fix 50 Euler steps rather than NFE. Correction is applied before CFG. Bold marks the best score within each block.}
\label{tab:diffusionbench-guidance-additional}
\endgroup

\endgroup
\end{table}

%% file: sections/figure_table_details.tex
\subsection{Figure and Table Details}
\label{app:figure-table-details}

\paragraph{Unconditional generation (Table~\ref{tab:unconditional}; Figure~\ref{fig:unconditional}).}
We evaluate SiT$_{\mathrm{REPA}}$-XL/2 with the null class after \UncondBaseStep{} updates, using EMA weights, 50,000 images, and the native Euler--Maruyama sampler. Correction detaches both the proposal and the base prediction. We sweep $K=0,\ldots,10$ using $\lfloor250/(K+1)\rfloor$ denoising steps, and autoguidance $w=1,\ldots,10$ using the released pretrained guide at 250 NFE. The table reports the lowest-FID settings from these sweeps, in addition to $K=1$. Sampling settings are summarized in Table~\ref{tab:uncond-settings}.

\paragraph{Text-to-image examples (Figure~\ref{fig:diffusionbench-samples}).}
We compare $K=0$ and $K=6$ using the same 50K-update EMA checkpoint, without CFG, and 50 Euler steps (50 and 350 NFE). Paired images share their prompt and recorded rank seed. Full prompts are listed in Table~\ref{tab:diffusionbench-full-prompts}.

\label{app:diffusionbench-full-prompts}

\input{generated/diffusionbench_full_prompts}

\paragraph{Guidance and correction (Figure~\ref{fig:diversity-control}).}
We compare $w\in\{1,6\}$ and $K\in\{0,6\}$ at 50 Euler steps. The selected prompt is \emph{a photo of a bench}, with all eight seeds retained per setting. CFG is applied before correction.

\paragraph{Parameter sharing (Table~\ref{tab:sit-shared-ladder}).}
We compare plain SiT-XL/2 models after 100K updates, without CFG, using FID50K. The shared model uses separate input embedders and shares the transformer blocks and output head. Both separate and shared correctors detach the proposal and base prediction, but shared parameters receive gradients from both losses. The baseline is a separate denoising-only continuation. Baseline, separate-corrector, and shared-corrector settings use $(K,\text{steps})=(0,250),(2,83),(3,62)$, respectively.

\paragraph{Distillation (Table~\ref{tab:distillation}).}
We freeze the separate SiT-XL/2 teacher at EMA100K and initialize the student from its denoiser. The student minimizes velocity MSE against the teacher's $K=2$ prediction, without a ground-truth loss, for \DistillKTwoStudentTrainingSteps{} updates on class-conditional ImageNet. We sample Gaussian noise and uniform time, using batch size 256, AdamW with learning rate $10^{-4}$, 100 warmup updates, zero weight decay, and EMA decay 0.9999. We evaluate the final EMA with FID@50K using the native SiT Euler--Maruyama sampler and ADM evaluator.

\paragraph{Power spectra (Figure~\ref{fig:power-spectrum}).}
We average radial power over 100 ImageNet-256 samples per setting for $K=0,\ldots,10$ at approximately 250 NFE. The panels show $\log_{10}$ power and $\log_{10}(P_{\mathrm{sample}}/P_{\mathrm{real}})$ over normalized frequencies 0.1--0.5.

%% file: generated/diffusionbench_full_prompts.tex
\begin{table}[htbp]
\centering
\caption{\textbf{Full prompts for the text-to-image examples in Figure~\ref{fig:diffusionbench-samples}.}}
\label{tab:diffusionbench-full-prompts}
\small
\renewcommand{\arraystretch}{1.15}
\begin{tabular}{@{}p{0.17\textwidth}p{0.79\textwidth}@{}}
\toprule
Figure column & Full prompt \\
\midrule
1. Green guitar & In the center of a dimly lit stage, a vintage green guitar with a glossy finish leans gently against a classic black amplifier. The polished wooden flooring of the stage reflects the warm, golden hue from a single spotlight above, which casts an inviting glow over the instrument. A microphone stand with a chrome finish stands to the left of the guitar, poised for the evening's performance. \\
\addlinespace[0.7em]
2. Vintage car & a vintage blue Porsche 356 is captured mid-turn on a winding asphalt road, its polished surface reflecting the bright sunlight. The car's classic design is accentuated by its rounded headlights and sleek bodywork. Alongside the road, a low stone wall partially covered in moss provides a contrast to the well-maintained vehicle. \\
\addlinespace[0.7em]
3. Sunlit kitchen & A spacious kitchen with a large wooden table at the center, cluttered with an assortment of pots, pans, and cooking utensils. The table is surrounded by matching wooden chairs, and the walls are lined with shelves filled with spices and kitchenware. Sunlight streams in through a window, casting a warm glow on the array of cookware. \\
\addlinespace[0.7em]
4. Three violins & Three exquisitely crafted violins, with their elegant F-shaped sound holes and smooth, glossy wooden surfaces, rest gently beside a grand piano. The piano's polished ebony finish reflects the soft, gentle curves of the violins that lie on a velvet-lined case. Behind them, the piano's open lid reveals its intricate strings and hammers, inviting a symphony of sounds to be played. \\
\addlinespace[0.7em]
5. Rabbit and seashell & On the soft, warm sand of the beach, a fluffy white rabbit with rounded ears is caught in a curious moment, gently placing its paw on the ribbed surface of a pink scallop shell. The scallop, slightly open, reveals its smooth interior contrasting with its coarse outer texture, while hues of pink and orange from the setting sun reflect off its surface. There's a tranquil ocean backdrop with the gentle ebbing of the tide, and the fading daylight casts a golden glow over the scene, highlighting the rabbit's soft fur and the shell's subtle color. \\
\addlinespace[0.7em]
6. Victorian library & A voluminous library, bathed in soft, muted blue hues, stretches majestically across a spacious Victorian-style living room. Intricately carved wooden bookshelves, filled with an array of books of various sizes and colors, line the walls and a grand fireplace sits as the room's focal point. The vray render showcases hyperrealistic textures, from the plush, patterned rug beneath the polished wooden tables, to the delicate fabric of the antique armchairs scattered throughout the space. \\
\bottomrule
\end{tabular}
\end{table}

%% file: sections/appendix/CFG.tex
\section{Additional Results}
\label{app:cfg-correction}
\label{app:human-evaluation}

\subsection{Comparing Correction and CFG}
\label{app:cfg-comparison}

\begin{figure}[htbp]
\centering
\includegraphics[width=\linewidth]{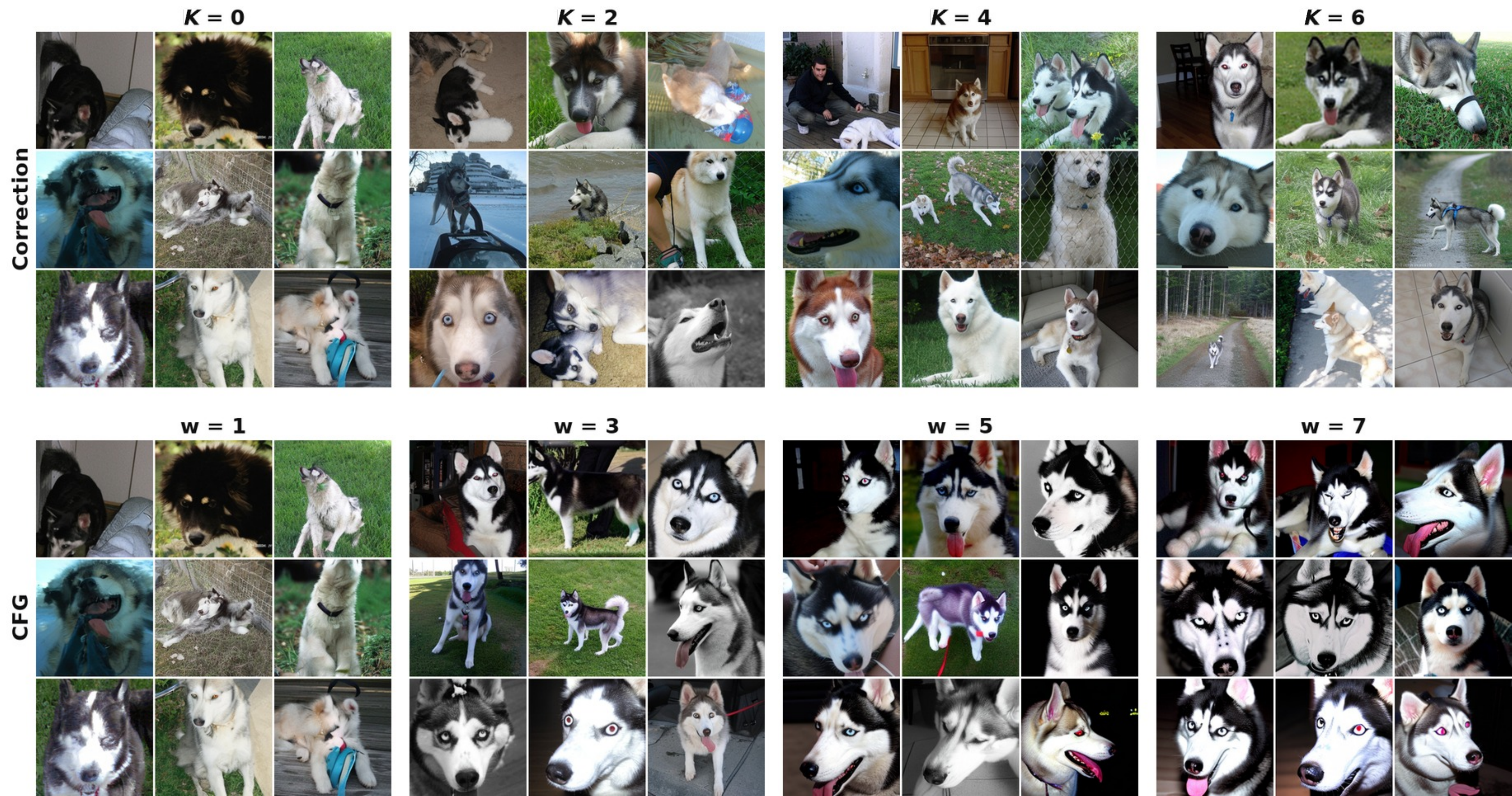}
\caption{\textbf{Correction and guidance produce different visual trade-offs.} Matched-noise samples from SiT$_{\mathrm{REPA}}$-XL/2 as correction depth $K$ (top) or guidance scale $w$ (bottom) increases, at approximately matched NFE.}
\label{fig:oversaturation}
\end{figure}

Fig.~\ref{fig:oversaturation} compares correction and CFG on ImageNet using the same checkpoint and initial noise. In these selected examples, increasing guidance produces more similar close-up compositions and stronger colors, whereas correction retains more varied poses and backgrounds. These observations concern the displayed samples; they do not by themselves establish distributional mode collapse.

\begin{figure}[htbp]
\centering
\includegraphics[width=\linewidth]{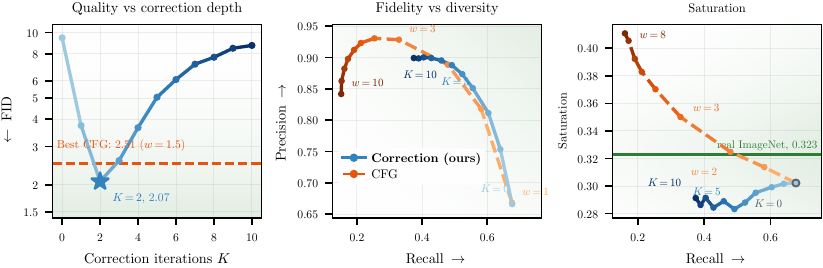}
\caption{\textbf{Correction improves FID with a different precision--recall and saturation trade-off.} DiT-XL/2 at approximately matched NFE. The dashed orange line marks the best evaluated CFG FID; the green line marks real-image saturation.}
\label{fig:cfg-pareto}
\end{figure}

Fig.~\ref{fig:cfg-pareto} quantifies the trade-offs on DiT. At approximately matched sampling compute, correction reaches a lower minimum FID than the best evaluated CFG setting. 
Following \citet{sadat2024eliminating}, we also measure mean HSV saturation against a real-image reference. Stronger CFG raises saturation above the reference, whereas correction remains near or below the unguided level. This ImageNet measurement complements the text-to-image examples; the latter illustrate refinement at weaker guidance rather than measuring a reduction in saturation caused by correction.

\input{sections/appendix/correction_guidance_samples}

\subsection{Combining CFG and Correction}
\paragraph{Combination order.}
CFG and correction can be applied in either order. Applying CFG first combines the conditional and null predictions before the corrector refines the guided prediction using the conditional text. Applying correction first refines both branches independently and then combines them with CFG. We use CFG $\rightarrow$ correction for the guided qualitative examples and the main human-preference comparison because we prefer its visual appearance, while correction $\rightarrow$ CFG is used for the guided alignment results in Tab.~\ref{tab:diffusionbench-t2i}.

Table~\ref{tab:human-preference-sweep} compares quality and diversity preferences at matching $w,K$ across the two orders. Among the evaluated guided settings, correction before CFG more often yields higher quality preference, while correction after CFG more often yields higher diversity preference. Thus neither order consistently leads on both criteria.

\input{generated/human_preference_sweep}

\subsection{Human evaluation}
\label{app:human-evaluation}
We evaluate our method with human raters, using \href{https://www.rapidata.ai/}{Rapidata} to collect human-feedback. It enables high-throughput collection of diverse human judgments, roughly takes few minutes on  thousands of human responses. Annotator qualification and reliability filtering are applied to control response quality. We additionally used language filter ``English" and age filter ``adult". 

Table~\ref{tab:human-evaluation} reports unweighted preference votes on all \HumanStudyPromptCount{} GenEval single-object prompts, with \HumanStudyVotesPerPrompt{} judgments per prompt (\HumanStudyVotesPerMetric{} per metric). Quality uses one image per method and diversity uses four-image sets. We first apply CFG and then correct the guided prediction using the conditional text. Both methods use 50 Euler steps.
Table~\ref{tab:human-preference-sweep} provides the full results for both correction orders over $w\in\{1,2,4,6\}$ and $K\in\{2,4,6,8\}$, highlighting the setting reported in the main text. Each setting is compared separately against the same released baseline. At $w=1$, the two orders use identical images, so differences reflect repeated human ratings. Figure~\ref{fig:human-study-example} illustrates the questions and stimuli for the main table's CFG-then-correction setting ($w=\HumanStudyCFG$, $K=\HumanStudyK$).
\input{generated/human_study_teddy}

%% file: sections/appendix/correction_guidance_samples.tex
\paragraph{Additional Correction and Guidance Comparisons}
\label{app:correction-guidance-samples}

\begin{figure}[!htp]
\centering
\captionsetup{font=footnotesize,skip=5pt}
\includegraphics[width=.85\linewidth]{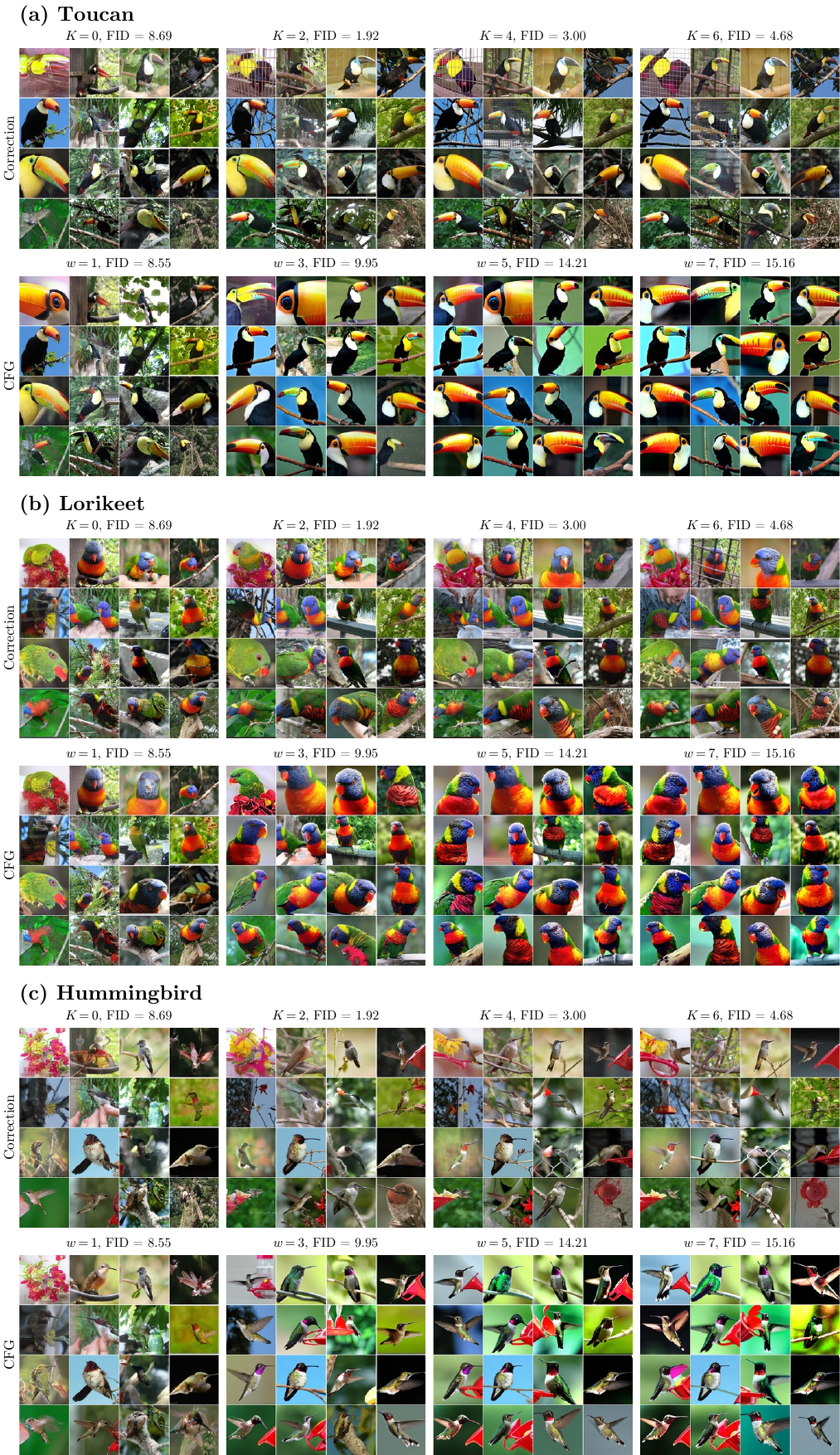}
\caption{\textbf{Correction and guidance comparisons.} \textbf{(a)} toucan, \textbf{(b)} lorikeet, and \textbf{(c)} hummingbird samples from SiT-XL/2. For each class, the upper row varies correction depth $K=0,2,4,6$ without CFG; the lower row varies guidance scale $w=1,3,5,7$ in a separately trained denoising-only model. Both models use EMA weights after 100K updates and approximately 250 NFE. All 16 fixed seeds appear in the same row-major order in every panel. FID labels use 50K samples across ImageNet classes.}
\label{fig:correction-guidance-class-96}
\label{fig:correction-guidance-class-90}
\label{fig:correction-guidance-class-94}
\end{figure}

\begin{figure}[!htp]
\centering
\captionsetup{font=footnotesize,skip=5pt}
\includegraphics[width=.85\linewidth]{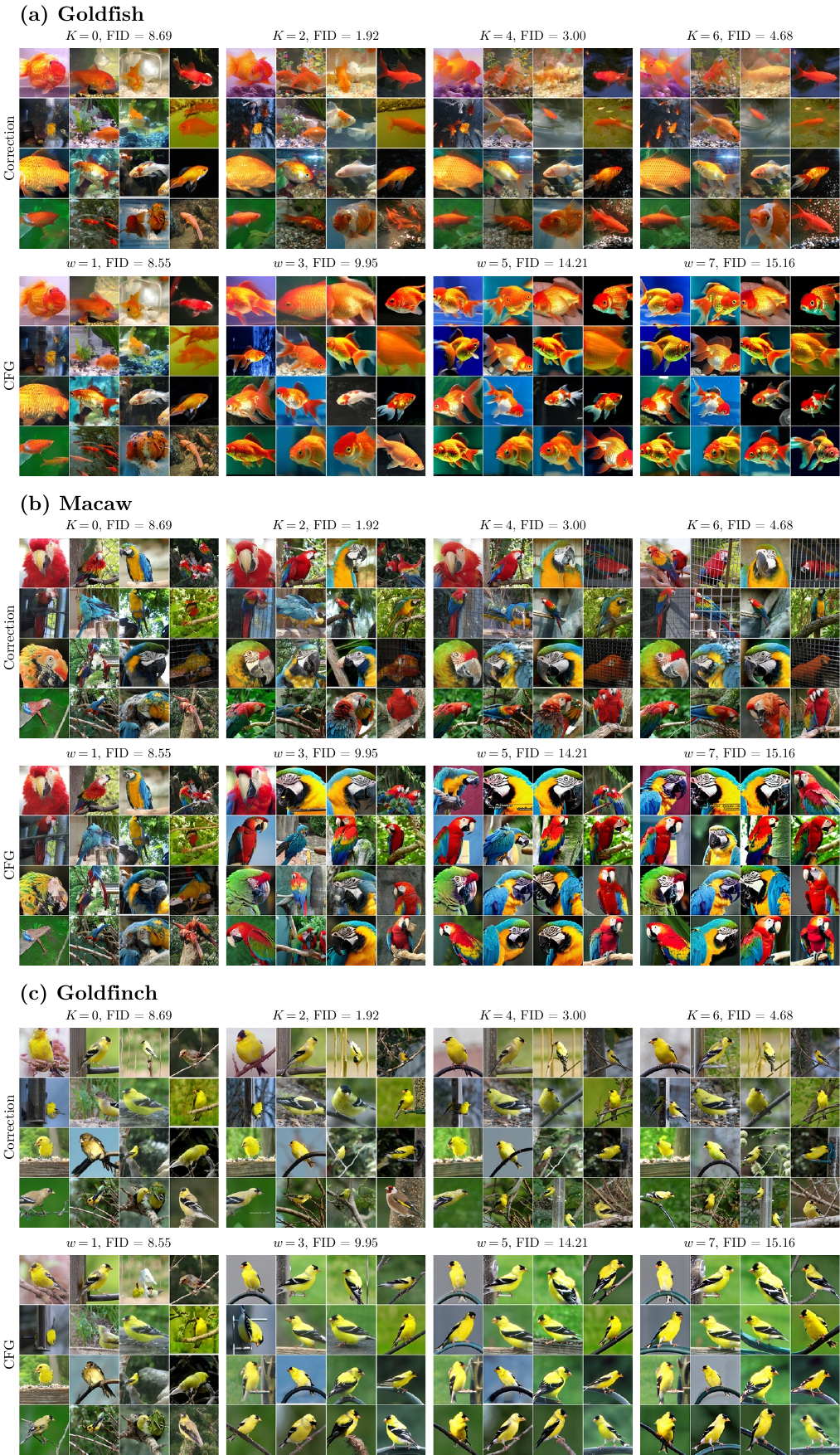}
\caption{\textbf{Correction and guidance comparisons.} \textbf{(a)} goldfish, \textbf{(b)} macaw, and \textbf{(c)} goldfinch samples from SiT-XL/2. For each class, the upper row varies correction depth $K=0,2,4,6$ without CFG; the lower row varies guidance scale $w=1,3,5,7$ in a separately trained denoising-only model. Both models use EMA weights after 100K updates and approximately 250 NFE. All 16 fixed seeds appear in the same row-major order in every panel. FID labels use 50K samples across ImageNet classes.}
\label{fig:correction-guidance-class-1}
\label{fig:correction-guidance-class-88}
\label{fig:correction-guidance-class-11}
\end{figure}

\begin{figure}[!htp]
\centering
\captionsetup{font=footnotesize,skip=5pt}
\includegraphics[width=.85\linewidth]{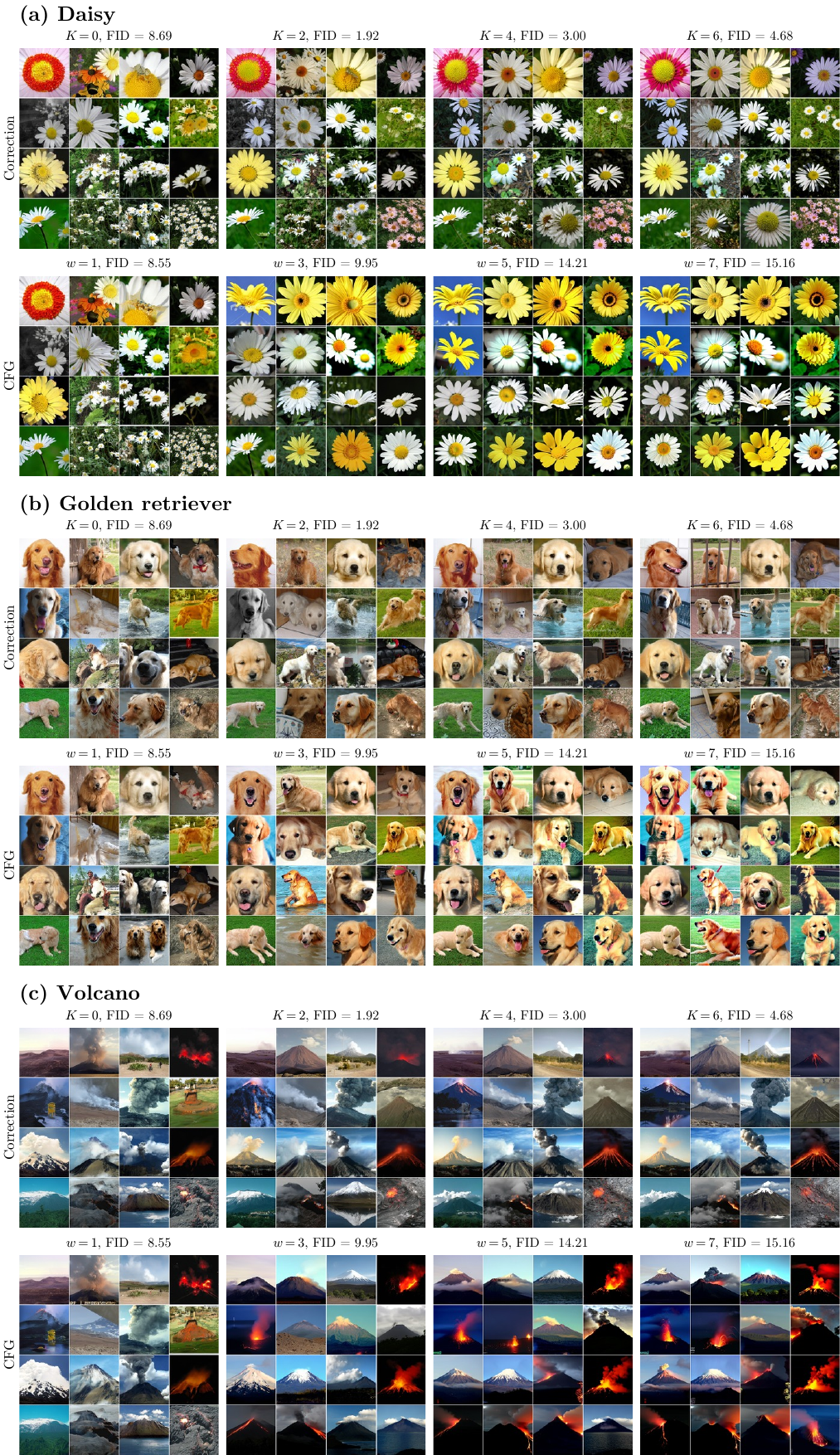}
\caption{\textbf{Correction and guidance comparisons across additional classes.}
\textbf{(a)} Daisy, \textbf{(b)} golden retriever, and \textbf{(c)} volcano.
SiT-XL/2 with the same checkpoints, sampling settings, and 16 fixed seeds as
Figs.~\ref{fig:correction-guidance-class-96} and~\ref{fig:correction-guidance-class-1}.
Top: correction $K=0,2,4,6$ without CFG. Bottom: CFG $w=1,3,5,7$.
FID labels are ImageNet-wide FID@50K.}
\label{fig:correction-guidance-class-985}
\label{fig:correction-guidance-class-207}
\label{fig:correction-guidance-class-980}
\end{figure}

Figures~\ref{fig:correction-guidance-class-96},~\ref{fig:correction-guidance-class-1}, and~\ref{fig:correction-guidance-class-985} retain all 16 fixed noise seeds (0--15) for nine classes selected from an exploratory 24-class comparison. No samples are omitted within these classes. The correction and guidance rows use separately trained SiT-XL/2 models at the same 100K-update budget, with EMA weights. Correction uses no CFG. All settings use approximately 250 network evaluations (249--252). Panels share the same initial Gaussian noise; their denoising grids and subsequent stochastic trajectories differ. FID labels are existing ImageNet-wide FID@50K results, rather than FID estimates for the displayed class. These examples illustrate visual trade-offs and do not estimate the number of modes covered by either method.

\clearpage

%% file: generated/human_preference_sweep.tex
\begin{table}[!htbp]
\centering
\small
\setlength{\tabcolsep}{7pt}
\caption{\textbf{Neither correction order consistently leads in both quality and diversity.} Full sweep of candidate vote shares (\%) against released CFG ($w=6$, $K=0$); higher is better. Bold marks the setting reported in Table~\ref{tab:human-evaluation}. Each entry contains 800 responses across 80 prompts. The two orders were rated separately against the baseline.}
\label{tab:human-preference-sweep}
\begin{tabular}{ccrrrr}
\toprule
& & \multicolumn{2}{c}{CFG $\rightarrow$ Correction} & \multicolumn{2}{c}{Correction $\rightarrow$ CFG} \\
\cmidrule(lr){3-4}\cmidrule(lr){5-6}
$w$ & $K$ & Quality & Diversity & Quality & Diversity \\
\midrule
1 & 2 & 33.38 & 55.38 & 37.25 & 54.50 \\
1 & 4 & 49.38 & 57.25 & 47.50 & 59.38 \\
1 & 6 & 55.75 & 55.75 & 52.38 & 60.75 \\
1 & 8 & 54.13 & 56.38 & 56.75 & 63.88 \\
\addlinespace[3pt]
2 & 2 & 48.50 & 60.50 & 55.38 & 59.13 \\
2 & 4 & 55.25 & 59.25 & 65.00 & 59.00 \\
2 & 6 & 57.25 & 58.75 & 64.25 & 61.50 \\
2 & 8 & 60.25 & 59.38 & 65.75 & 59.50 \\
\addlinespace[3pt]
4 & 2 & 54.63 & 61.25 & 61.63 & 52.25 \\
4 & 4 & 58.25 & 59.50 & 67.00 & 57.50 \\
4 & 6 & 61.75 & 55.38 & 66.88 & 55.63 \\
4 & 8 & \textbf{60.75} & \textbf{60.25} & 60.50 & 56.63 \\
\addlinespace[3pt]
6 & 2 & 57.13 & 54.38 & 59.50 & 49.63 \\
6 & 4 & 59.63 & 52.75 & 61.00 & 53.75 \\
6 & 6 & 57.38 & 58.38 & 63.63 & 54.00 \\
6 & 8 & 58.63 & 58.38 & 59.75 & 58.13 \\
\bottomrule
\end{tabular}
\end{table}

%% file: generated/human_study_teddy.tex
\begin{figure}[!htbp]
\centering
\includegraphics[width=\linewidth]{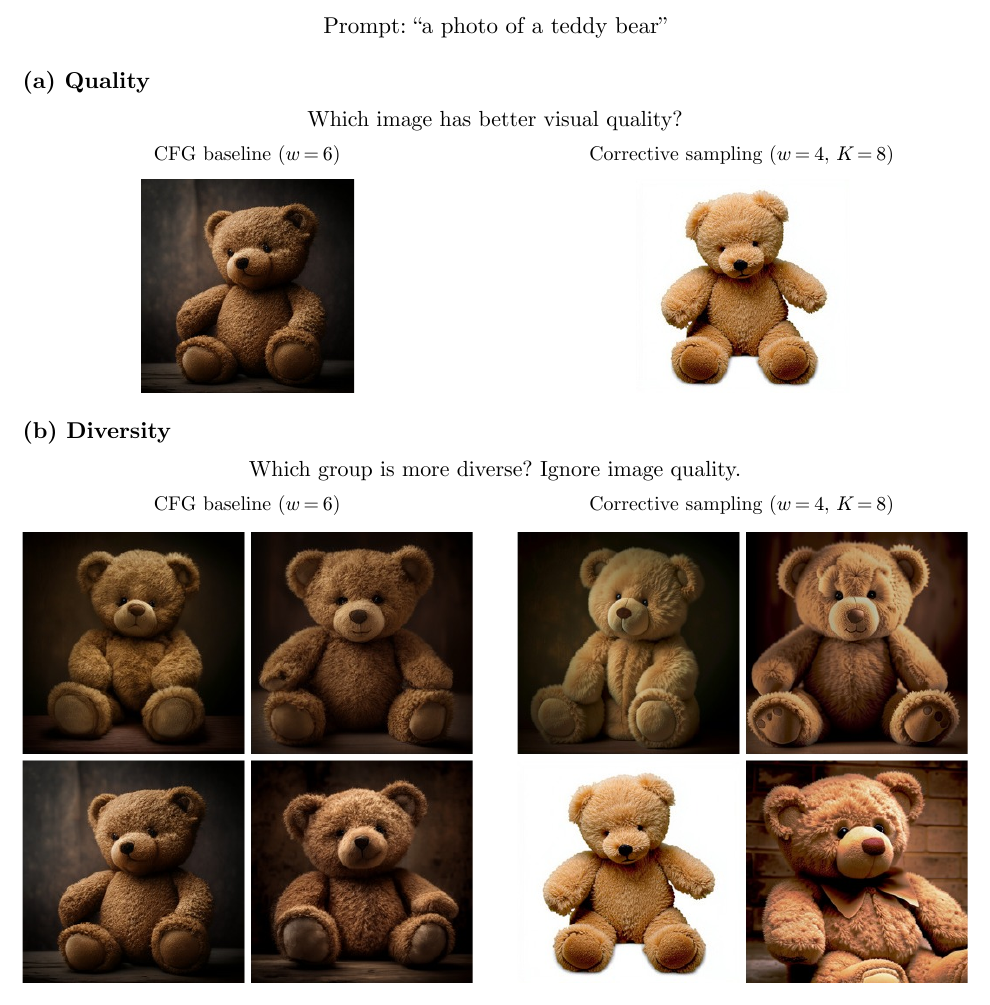}
\caption{\textbf{Quality and diversity are assessed separately.} Actual teddy-bear stimuli and questions for Table~\ref{tab:human-evaluation}, using CFG followed by correction ($w=4$, $K=8$): single images for quality and four-image sets for diversity.}
\label{fig:human-study-example}
\end{figure}

%% file: sections/appendix/discussion.tex
\section{Discussion}
\label{app:dis}

\subsection{Training and capacity controls}

\label{subsec:discussion-k-1}
\label{gradient-variance-corrector-training}
\input{generated/repa_learning_variance_results}

\begin{figure}[!htbp]
    \centering
    \includegraphics[width=\linewidth]{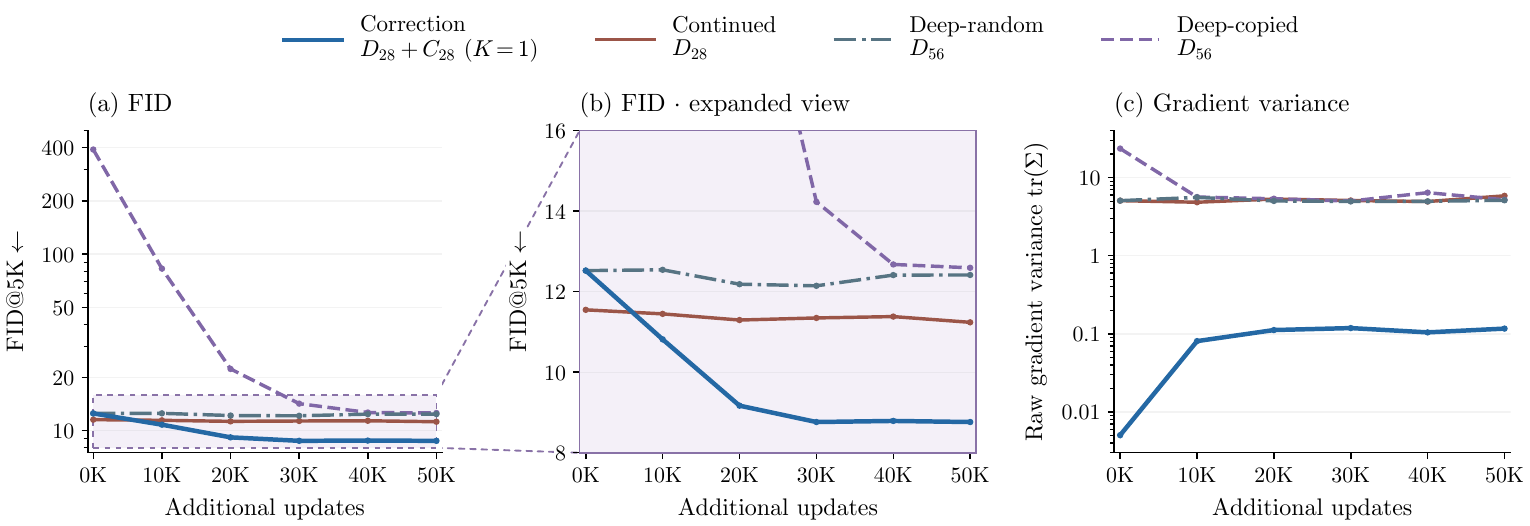}
    \input{captions/repa_learning_variance}
    \label{fig:convergence-comparison}
\end{figure}

\paragraph{Additional-training and capacity controls.} 
We examine how quickly the corrector converges by comparing correction training with three baselines: continued training of the base denoiser and two deeper denoisers with different initialization schemes. We use the pretrained \SiTREPA-XL/2 denoiser with 28 transformer layers from Tab.~\ref{tab:imagenet-results}. For the deeper denoisers, we append 28 transformer layers before the final unembedding layer, matching the total number of transformer layers in the denoiser--corrector pair. The added layers are either randomly initialized or copied from the pretrained denoiser, with the latter matching our corrector initialization. We train all configurations for 50K steps and report their progress in Fig.~\ref{fig:convergence-comparison}.
\yb{We use a single correction pass ($K=1$) to isolate its benefit from the additional gains of iterative refinement.}

As can be seen in Fig.~\ref{fig:convergence-comparison} (a) and (b), correction improves FID substantially faster over the 50K-update training window we used, whereas continued training and both deeper-denoiser controls improve much less. Importantly, this gain appears with only a \textit{single correction pass}: despite using the same total sequential transformer depth as the 56-layer denoisers, the $D_{28}+C_{28}$ correction model achieves substantially better generation quality.
\yb{These results suggest that \textit{additional training or model depth alone cannot explain the gains}, pointing to an advantage of learning to refine an already structured prediction over learning directly from noisy observations.}

\paragraph{Iterative correction improves quality at matched effective depth.}
\label{app:matched-effective-depth}
\input{generated/matched_effective_depth_allocation}
One might argue that the gain from repeating correction simply comes from increasing the effective depth within each denoising step. To test whether this is the case, we compare the 28-layer monolithic denoiser ($D_{28}$) with a 7-layer denoiser and a 7-layer corrector ($D_7+C_7$), using $K=3$ corrections to match the effective depth of the monolithic denoiser ($7+3\times7=28$; Fig.~\ref{fig:matched-effective-depth-allocation}). For this comparison, we train both configurations from scratch for 80K updates. Both use exactly 240 denoising steps and 6,720 transformer-layer evaluations per sample. Surprisingly, despite using roughly half as many parameters (346M versus 675M), the shallow denoiser--corrector pair achieves a lower FID@50K (4.21 versus 4.88; Tab.~\ref{tab:deeper-denoiser}). This highlights that iterative correction does more than improve prediction accuracy through additional denoiser layers. 

\FloatBarrier
\subsection{Iterative correction visualizations}
\label{app:correction-qualitative}

\paragraph{Analytic target.}
Let $\mathcal{B}_y=\{\bm{x}_i\}_{i=1}^{N_y}$ be a bank of clean images from class $y$, with a uniform empirical prior.
For $\bm{z}_t=(1-t)\bm{x}+t\bm{\epsilon}$, where $\bm{\epsilon}\sim\mathcal{N}(\bm{0},\bm{I})$ and $0<t\leq1$, the posterior mean under this prior is
\begin{equation}
\label{eq:analytic}
\begin{aligned}
\bm{x}_{\mathrm{analytic}}(\bm{z}_t,t,y)
&=\frac{\sum_{i=1}^{N_y}w_i(\bm{z}_t,t)\,\bm{x}_i}
         {\sum_{i=1}^{N_y}w_i(\bm{z}_t,t)},\\
w_i(\bm{z}_t,t)
&=\exp\!\left(-\frac{\|\bm{z}_t-(1-t)\bm{x}_i\|_2^2}{2t^2}\right).
\end{aligned}
\end{equation}

\input{sections/appendix/discussion_new}

\input{sections/appendix/correction_qualitative}

\paragraph{Correction within a denoising step.}
\label{app:correction-visualizations}
Fig.~\ref{fig:iterative-correction} provides a qualitative view of how correction changes intermediate clean-sample predictions during sampling. Without correction, structural errors in an intermediate prediction can persist into subsequent denoising steps because the prediction is immediately used to update the noisy state. Correction instead refines the prediction at fixed $t$ before this update. In the examples in Fig.~\ref{fig:iterative-correction}, successive corrections sharpen object structure and suppress visible artifacts within a single denoising step, so the subsequent sampler state is constructed from an improved prediction.

To illustrate what correction does at each step, Fig.~\ref{fig:correction-full-dog} shows the individual correction passes for three examples from \SiTREPA-XL/2. For each example, the baseline and correction trajectories start from the same initial noise and use the same 15 Euler denoising steps, with $K=8$ for correction. In the dog example (Fig.~\ref{fig:correction-full-dog}), already at early-to-middle stages of sampling ($t=0.73$--$0.60$), successive corrections reduce structural artifacts and reveal a coherent dog shape. These improvements carry through to the final image, while the baseline retains structural artifacts even after completing all denoising steps.

\input{sections/appendix/correction_trajectories}
\input{generated/joint_frozen_selected_appendix}

\FloatBarrier
\subsection{Power spectrum analysis}
\label{app:power-spectrum-analysis}

\paragraph{Frequency-domain effects of correction.}
To further characterize the changes induced by correction, we examine the frequency content of generated samples. Prior work~\citep{yu2025frequency,kobayashi2026spectral} has reported lower subband energy in reverse-process samples than in forward-corrupted real data. This motivates examining whether correction recovers frequency components that are underrepresented in baseline samples.

We compare the power spectra of samples generated with $K=0,\ldots,10$ correction iterations against those of real images. As shown in Fig.~\ref{fig:power-spectrum}, the baseline samples exhibit a spectral deficit, particularly at high frequencies. Increasing the number of correction iterations reduces this deficit, with $K=6$ yielding a spectrum that approximately matches the real-image reference. Further iterations can raise both low- and high-frequency power above the reference, indicating that the effect depends on the correction count.
\input{generated/correction_power_spectrum}

%% file: generated/repa_learning_variance_results.tex
\providecommand{\RepaLearningCorrectionFID}{8.76}
\providecommand{\RepaLearningContinuedFID}{11.24}
\providecommand{\RepaLearningRandomFID}{12.42}
\providecommand{\RepaLearningCopiedFID}{12.59}
\providecommand{\RepaLearningUpdates}{50K}

%% file: captions/repa_learning_variance.tex
\caption{\textbf{Correction improves generation quality beyond continued denoiser training or parameter-matched deeper denoisers.}}

%% file: generated/matched_effective_depth_allocation.tex
\begin{figure}[!ht]
    \centering
    \includegraphics[width=\linewidth]{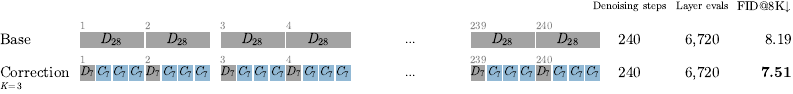}
    \caption{\textbf{Matched effective depth.} Correction outperforms a deeper denoiser at matched effective depth per denoising step and inference budget, using fewer parameters.}
    \label{fig:matched-effective-depth-allocation}
\end{figure}

%% file: sections/appendix/discussion_new.tex
\begin{figure}[!htbp]
    \centering
    \captionsetup{font=footnotesize,skip=5pt}
    \includegraphics[width=.79\linewidth]{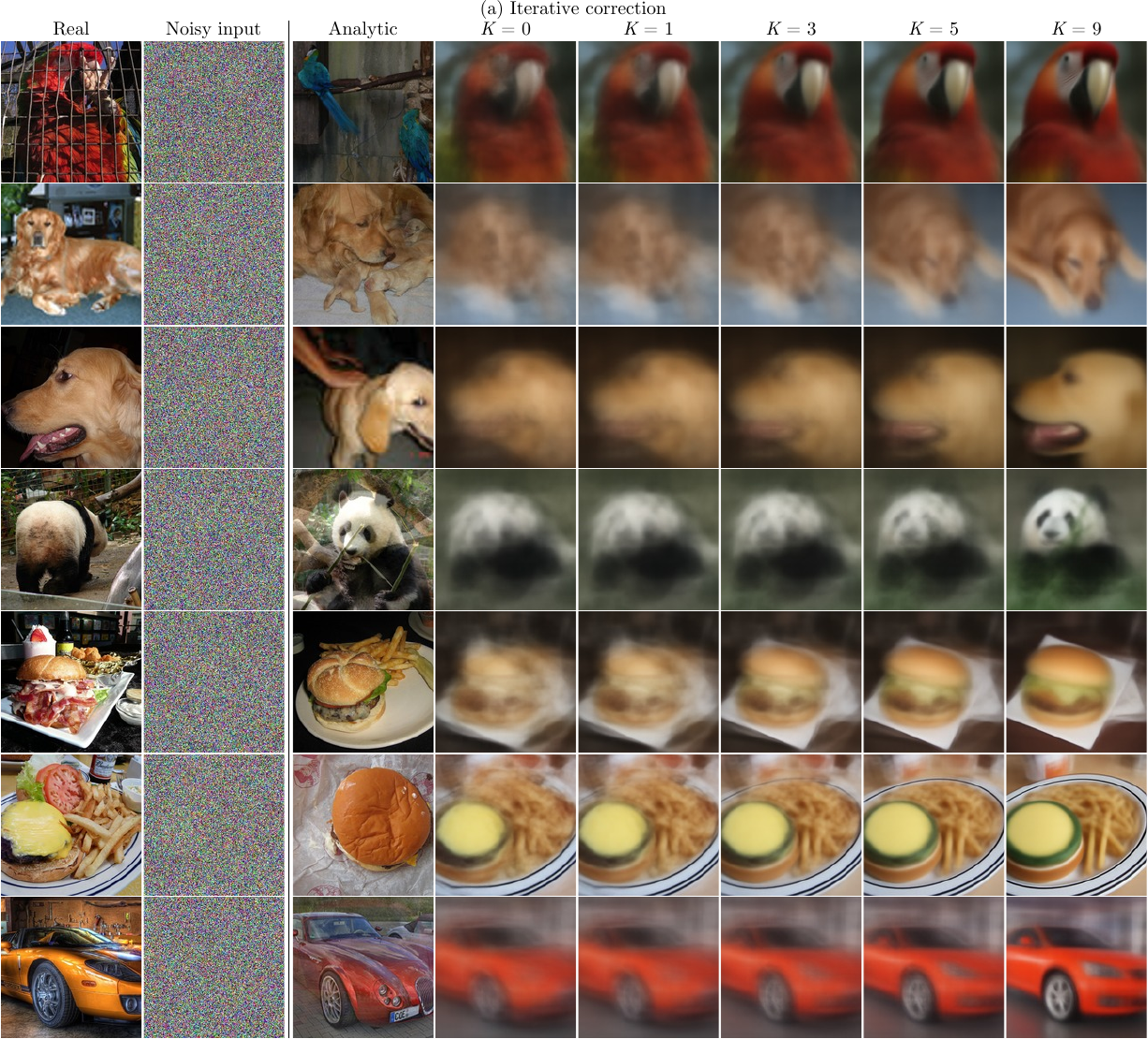}
    \par\vspace{2pt}
    \includegraphics[width=.79\linewidth]{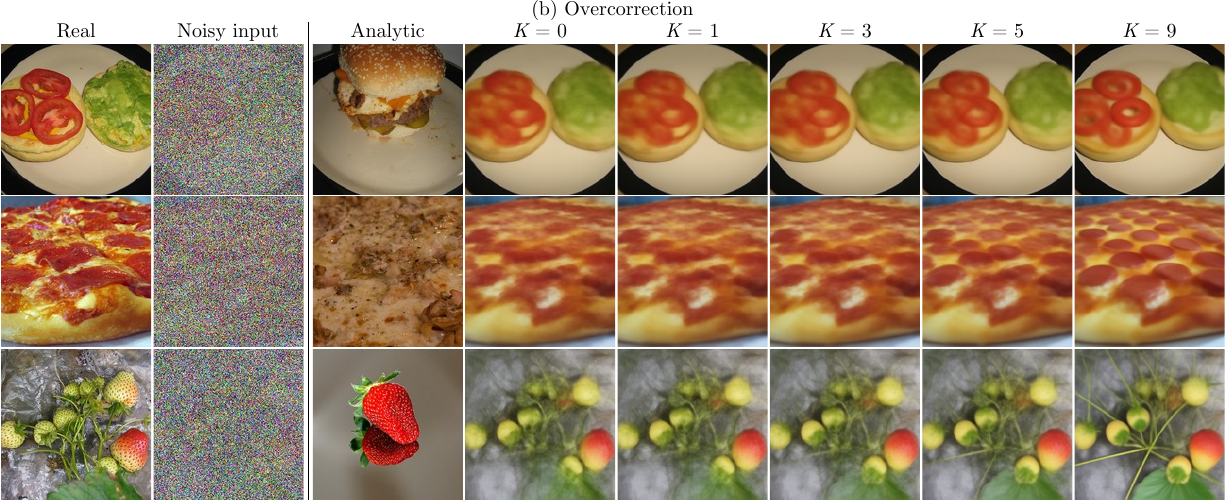}
    \caption[]{\textbf{Additional iterative-correction examples.}
    The seven selected examples not included in Fig.~\ref{fig:repeated-correction}(a), using the same column layout and fixed-input protocol.
    Repetition strengthens object contours and details, while the predictions need not align with the original clean image or the finite-bank analytic estimate.
\par\smallskip
\textbf{Additional overcorrection examples.}
    The three selected examples not included in Fig.~\ref{fig:repeated-correction}(b).
    Repeated correction can simplify irregular textures or amplify thin structures into unnatural patterns.}
\makeatletter
\begingroup
\protected@edef\@currentlabel{\thefigure(a)}
\label{fig:correction-qualitative-additional}
\endgroup
\begingroup
\protected@edef\@currentlabel{\thefigure(b)}
\label{fig:overcorrection-additional}
\endgroup
\makeatother
\end{figure}

\paragraph{Additional iterative-correction examples.}
Figures~\ref{fig:correction-qualitative-additional} and~\ref{fig:overcorrection-additional} show the remaining selected examples from the visualization in Fig.~\ref{fig:repeated-correction}, without repeating its six rows.
All use the same JiT-L/16 checkpoint after 100K training updates and its second EMA weights, with class conditioning and CFG scale $1$.
For each row, the denoiser receives a forward input $\bm{z}_t=(1-t)\bm{x}+t\bm{\epsilon}$, and the corrector is repeatedly applied to the clean prediction at that fixed timestep.
The selected examples use noise levels $t\in\{0.8,0.9,0.95,0.975\}$, where $1$ denotes noise and $0$ denotes data.
The Analytic column uses the same formula (Eq.~\ref{eq:analytic}), with a uniform empirical prior over a finite same-class image bank excluding the original query cohort.
The examples were selected for qualitative illustration and are not a random-sample quality estimate.

\begin{figure}[!htp]
\centering
\captionsetup{font=footnotesize,skip=5pt}
\includegraphics[width=.76\linewidth]{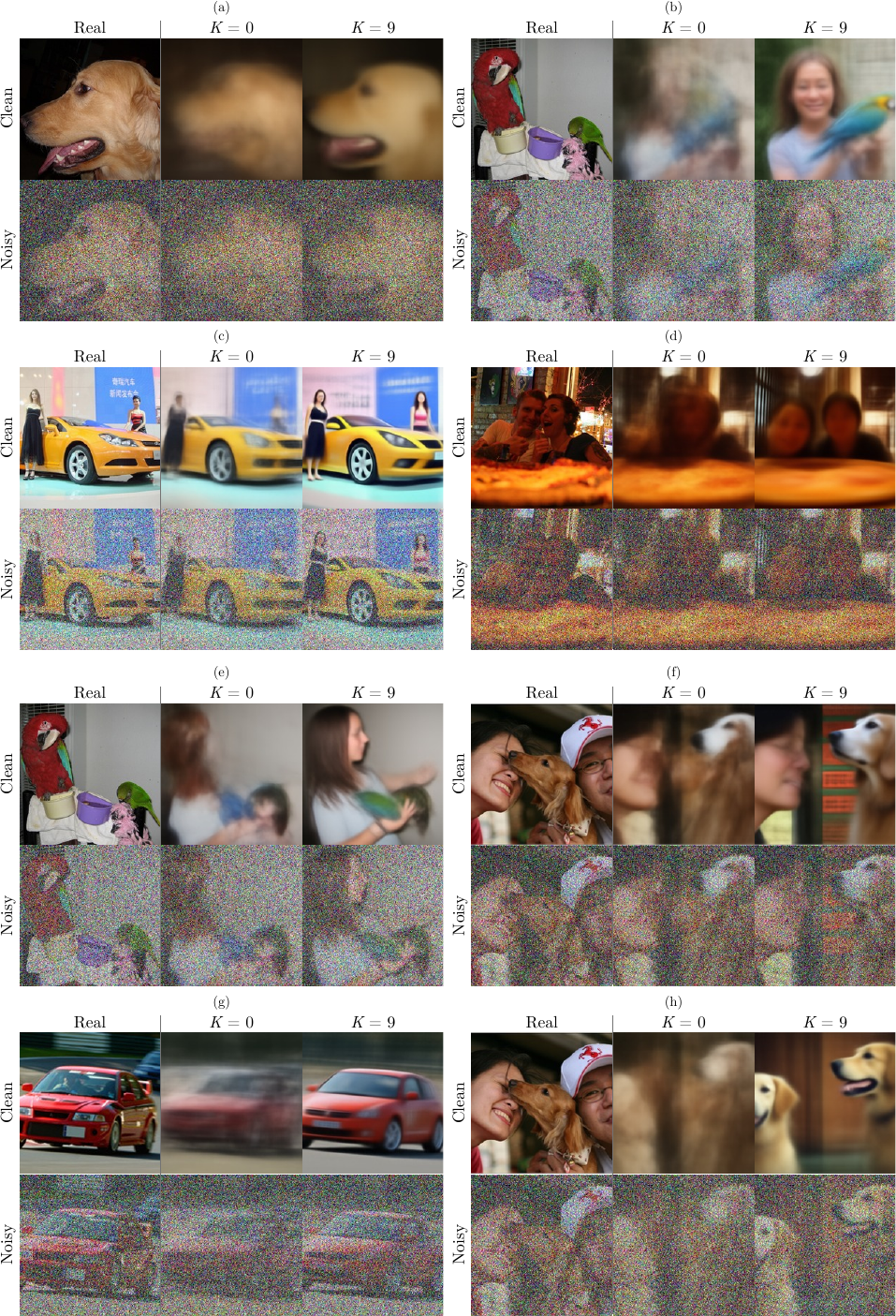}
\caption{\textbf{Shared-input noisy-state comparisons.} Clean images (top) and interpolated noisy states (bottom), comparing \emph{Real}, $K=0$, and $K=9$. Panels \textbf{(a,b,g,h)} use $t=0.95\to s=0.55$; panels \textbf{(c,d,e,f)} use $t=0.9\to s=0.5$. Source queries (classes), in panel order, are 264649 (207), 114599 (88), 1047154 (817), 1233725 (963), 114599 (88), 265339 (207), 1047733 (817), and 265339 (207). Repeated query IDs are shown at different noise levels.}
\label{fig:noised-repeated-correction-additional}
\label{fig:noised-repeated-correction-additional-2}
\end{figure}

\paragraph{Interpolation of corrected predictions toward noise.}
Figure~\ref{fig:noised-repeated-correction-additional} shows all eight selected noisy-state comparisons.
They use $t=0.9\to s=0.5$ or $t=0.95\to s=0.55$, with the same interval $t-s=0.4$ throughout.
For each example and every $K$, we hold the original forward input $\bm{z}_t$ fixed and construct $\bm{z}_s^{(K)}=(s/t)\bm{z}_t+(1-s/t)\hat{\bm{x}}^{(K)}$.
The real reference is $(1-s)\bm{x}+s\bm{\epsilon}$ with the same original Gaussian.
Interpolation is applied to unclipped predictions; all cells use the same display range, without contrast normalization.
This is an illustrative interpolation to a lower noise level rather than a single step from the sampling grid.
The corrected predictions need not reconstruct the original image: the comparison concerns the plausibility of visible structure after interpolation, rather than paired reconstruction accuracy.

%% file: sections/appendix/correction_qualitative.tex
\relax

%% file: sections/appendix/correction_trajectories.tex
\begin{figure}[p]
\centering
\includegraphics[width=.98\linewidth]{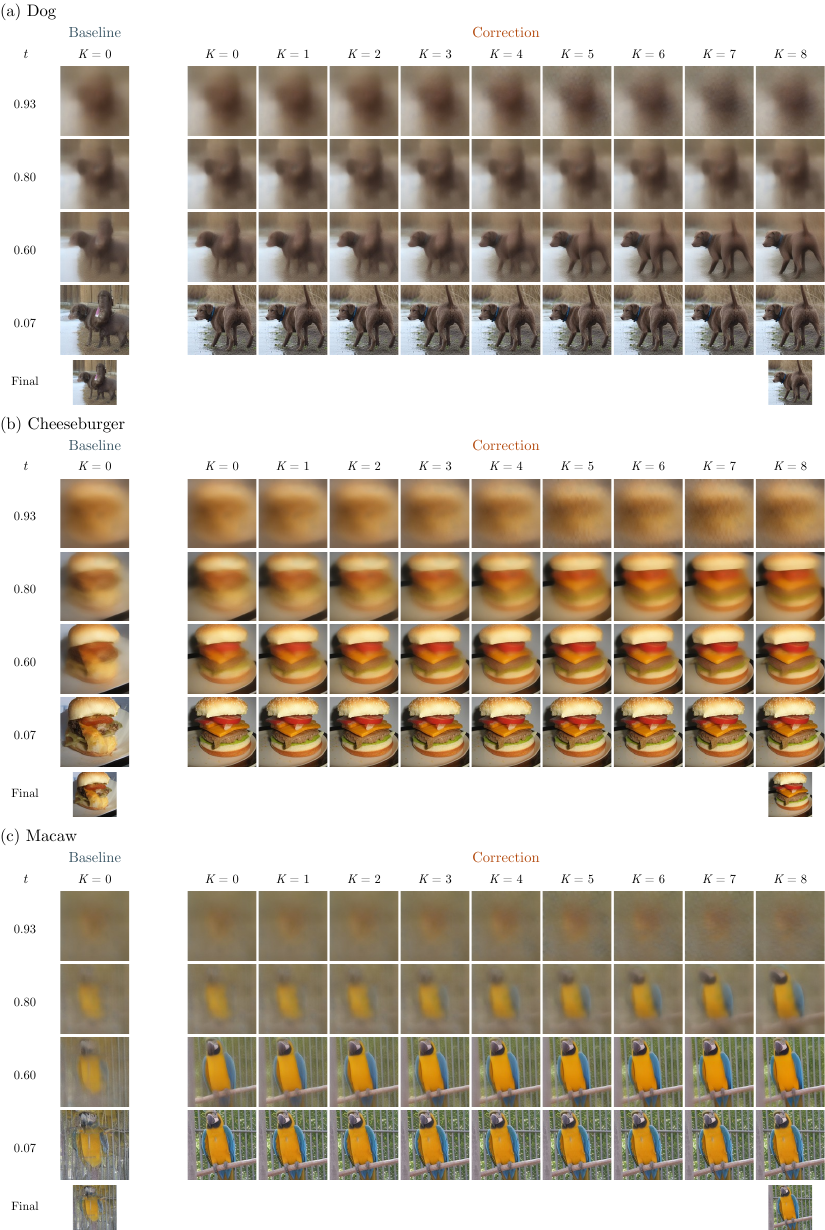}
\caption{\textbf{Correction refines intermediate predictions before denoising continues.} Baseline and correction start from the same noise and use 15 denoising steps. Panels (a--c) show dog, cheeseburger, and macaw examples at selected timesteps $t\in\{0.93,0.80,0.60,0.07\}$. Each prediction row shows all eight correction passes at fixed $t$; the final row of each panel shows the actual final samples.}
\label{fig:correction-full-dog}
\label{fig:correction-full-hamburger}
\label{fig:correction-full-macaw}
\end{figure}
\clearpage

%% file: generated/joint_frozen_selected_appendix.tex
\clearpage
\begin{figure}[p]
\centering
\includegraphics[width=.99\linewidth]{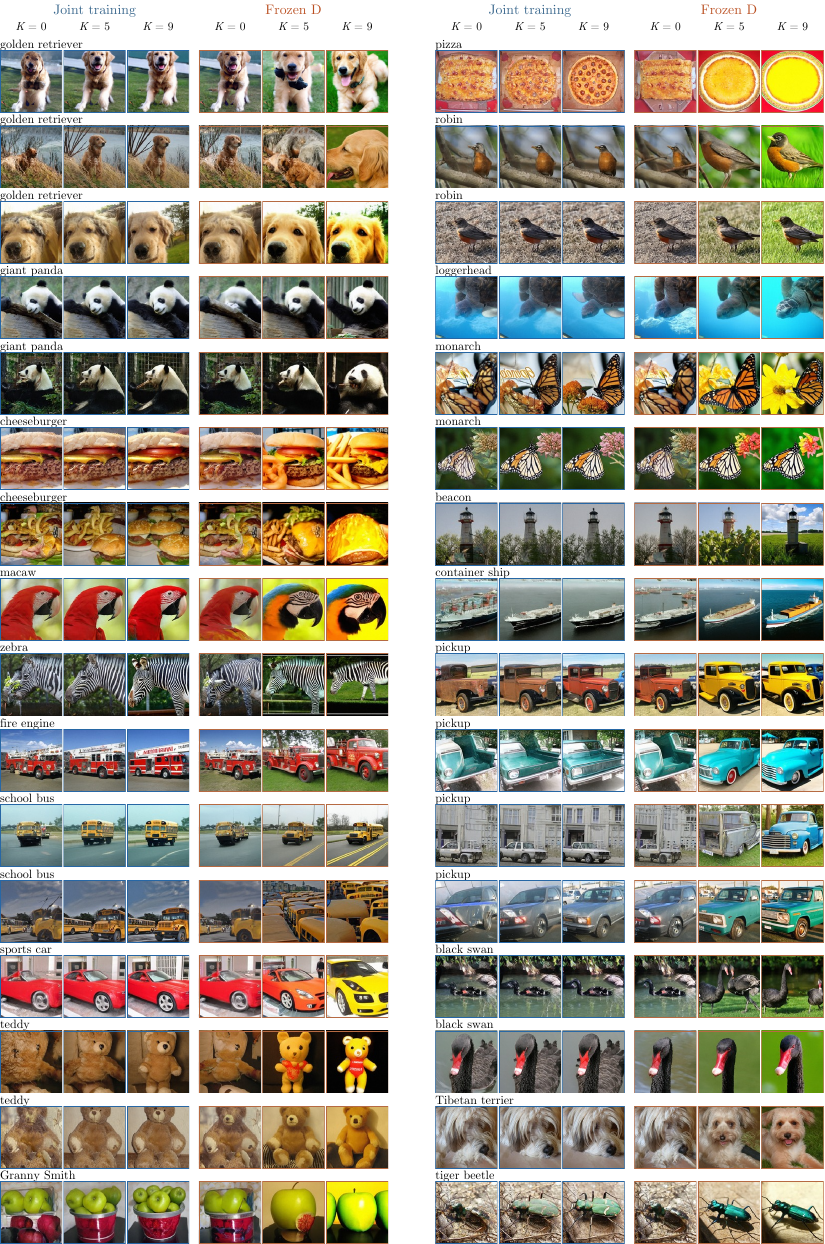}
\caption{\textbf{Additional examples with joint and frozen denoisers.} Each row compares joint training (left triplet) and a frozen denoiser (right triplet) at $K=0,5,9$, starting from the same noise. Selected JiT-L/16 samples use 100K checkpoints, EMA weights, no CFG, and 45 Heun steps; larger $K$ adds correction compute.}
\label{fig:joint-frozen-selected-1}
\label{fig:joint-frozen-selected-2}
\end{figure}
\clearpage

%% file: generated/correction_power_spectrum.tex
\begin{figure}[!htbp]
    \centering
    \captionsetup{font=footnotesize,skip=3pt,justification=raggedright,singlelinecheck=false}
    \includegraphics[width=0.82\linewidth]{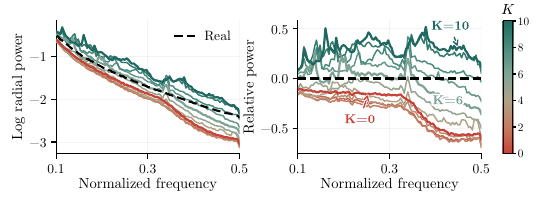}
    \caption{\textbf{Correction reduces the high-frequency deficit of generated samples.} Left: $\log_{10}P$. Right: $\log_{10}(P/P_{\mathrm{real}})$. Dashed lines indicate real-image references.}
    \label{fig:power-spectrum}
\end{figure}

%% file: sections/more_samples.tex
\section{Additional Samples}
\label{app:more-samples}

\input{sections/appendix/more_samples_paired}

\subsection{Unconditional Generation}

\begin{figure}[!htbp]
    \centering
    \includegraphics[width=\linewidth]{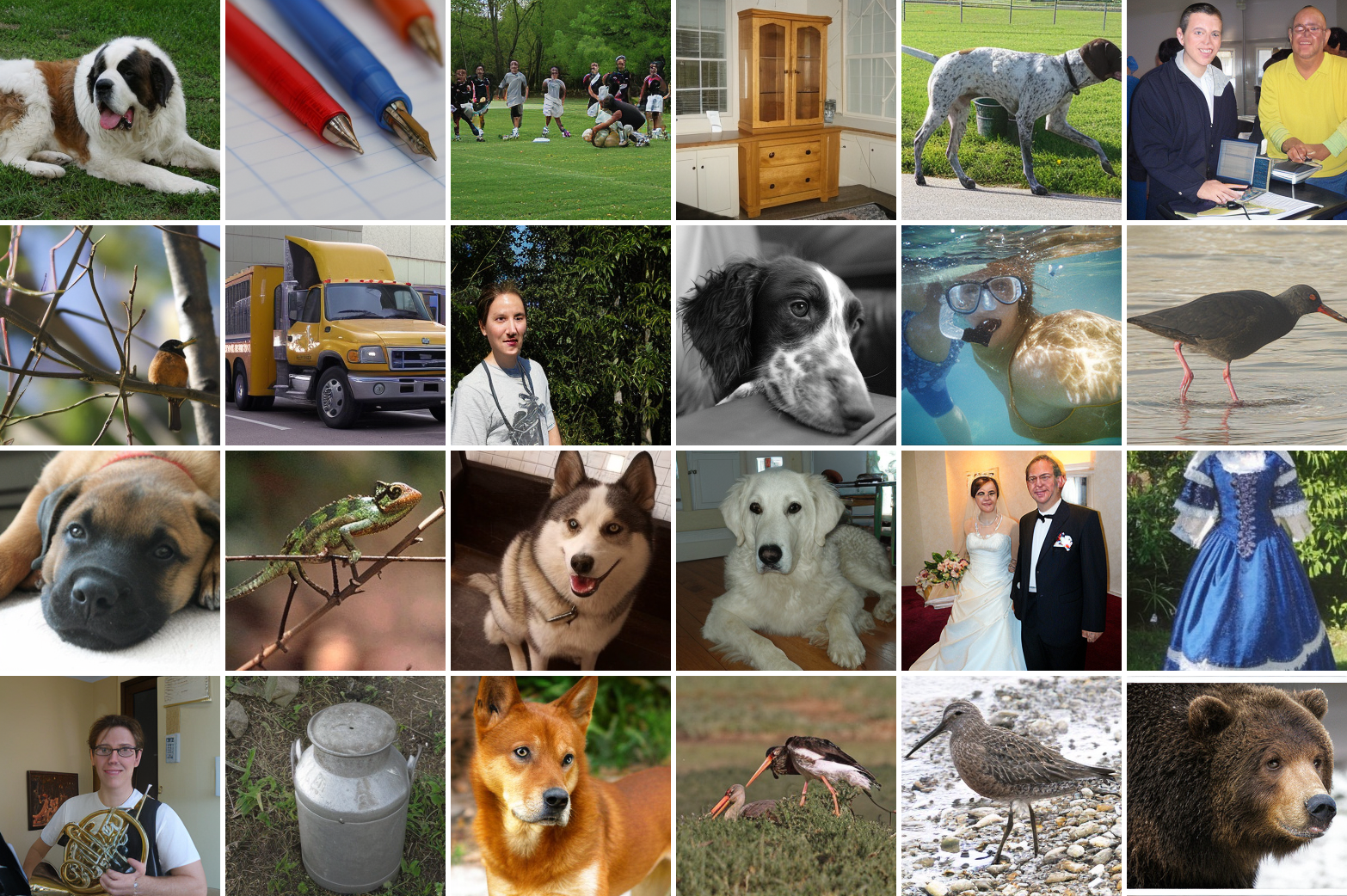}
    \caption{\textbf{High-quality unconditional generation without guidance.} Selected ImageNet samples from SiT$_{\mathrm{REPA}}$-XL/2 with iterative correction.}
    \label{fig:imagenet-samples-uncond}
\end{figure}

%% file: sections/appendix/more_samples_paired.tex
\subsection{Effect of Correction Depth}
\label{app:more-samples-depth}
\label{app:conditional-samples}

We show the same class and initial noise as correction depth increases, followed by paired galleries over randomly drawn ImageNet classes. All examples use class conditioning without CFG.

\begin{figure}[!ht]
    \centering
    \includegraphics[width=\linewidth]{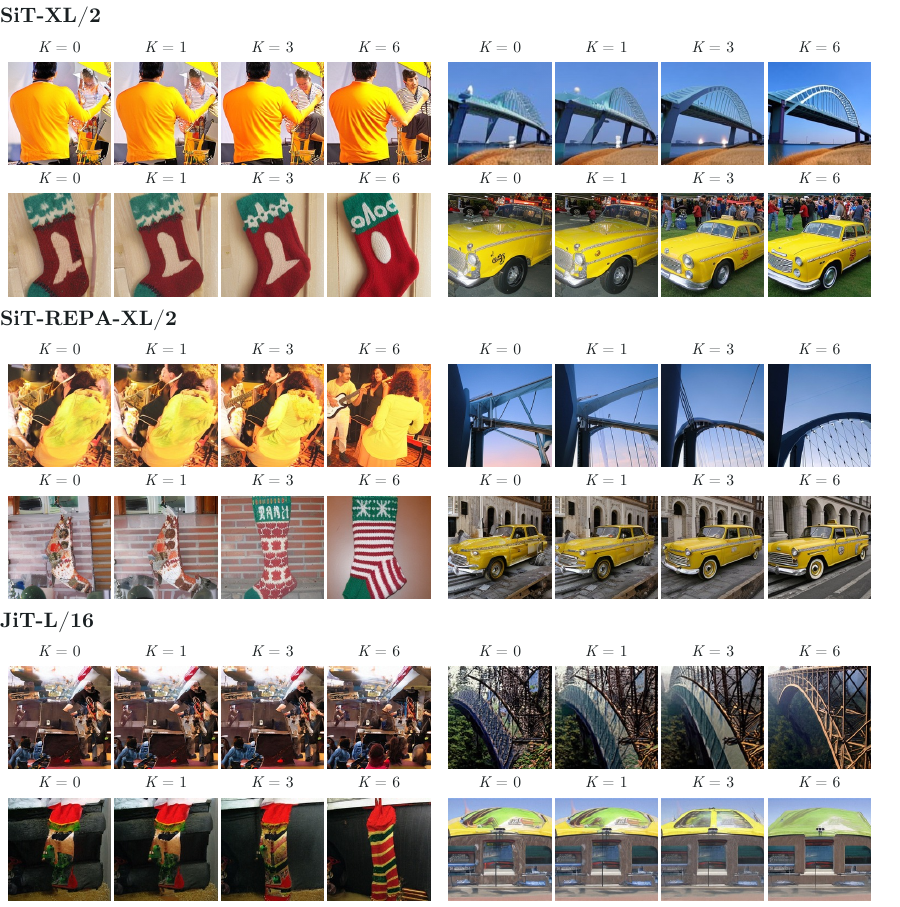}
    \caption{\textbf{\textit{Uncurated} samples: correction depth changes both structure and detail.} Each strip follows one sample as $K$ increases, with the initial noise and denoising grid fixed.}
    \label{fig:more-samples-k-sweep}
\end{figure}

\clearpage
\subsection{SiT: Random-Class Generation}

\begin{figure}[!ht]
    \centering
    \includegraphics[width=\linewidth]{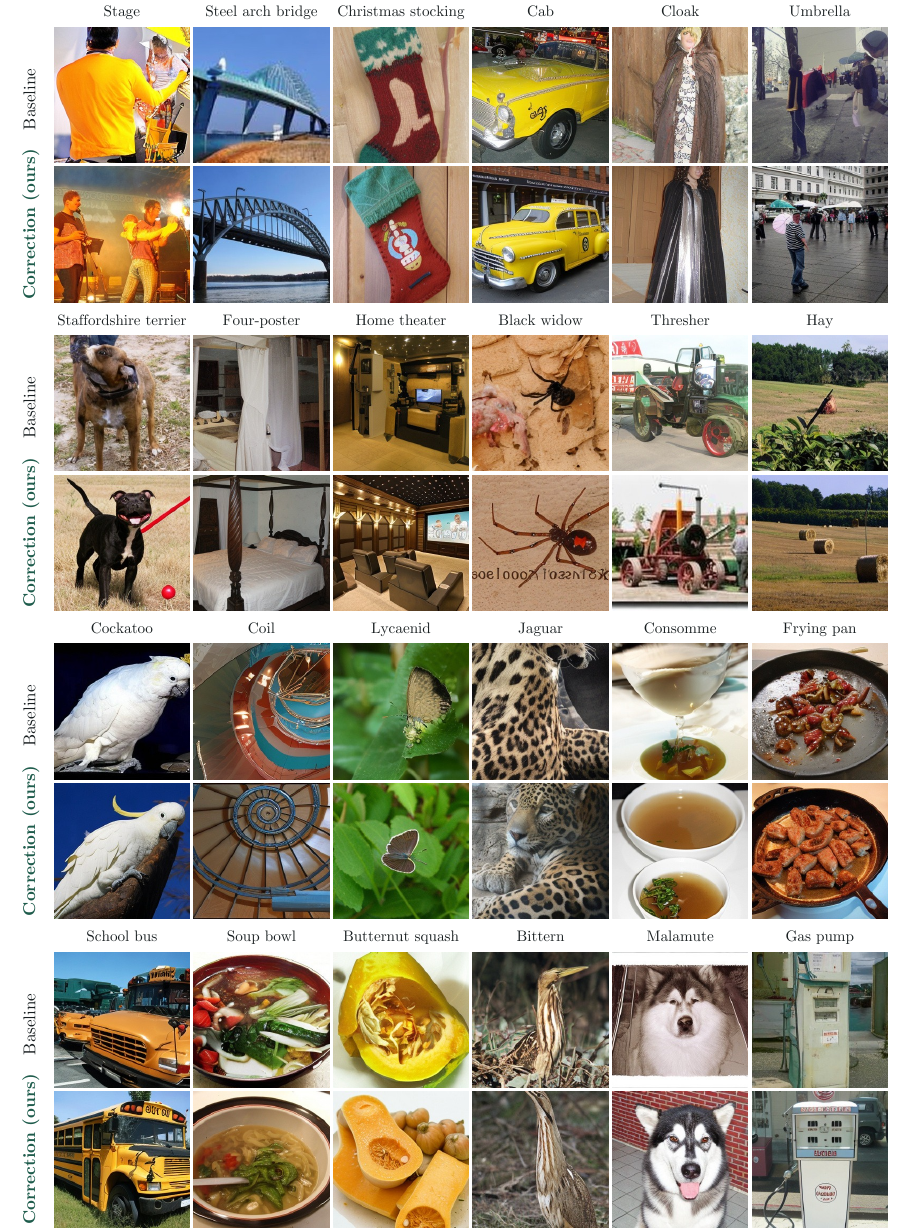}
    \caption{\textbf{\textit{Uncurated} samples: correction refines unguided SiT samples across random classes.} Within each pair of rows, baseline ($K=0$) and correction ($K=6$) share the same class and initial noise at approximately matched NFEs.}
    \label{fig:more-samples-random-sit}
\end{figure}

\clearpage
\subsection{SiT$_{\mathrm{REPA}}$: Random-Class Generation}

\begin{figure}[!ht]
    \centering
    \includegraphics[width=\linewidth]{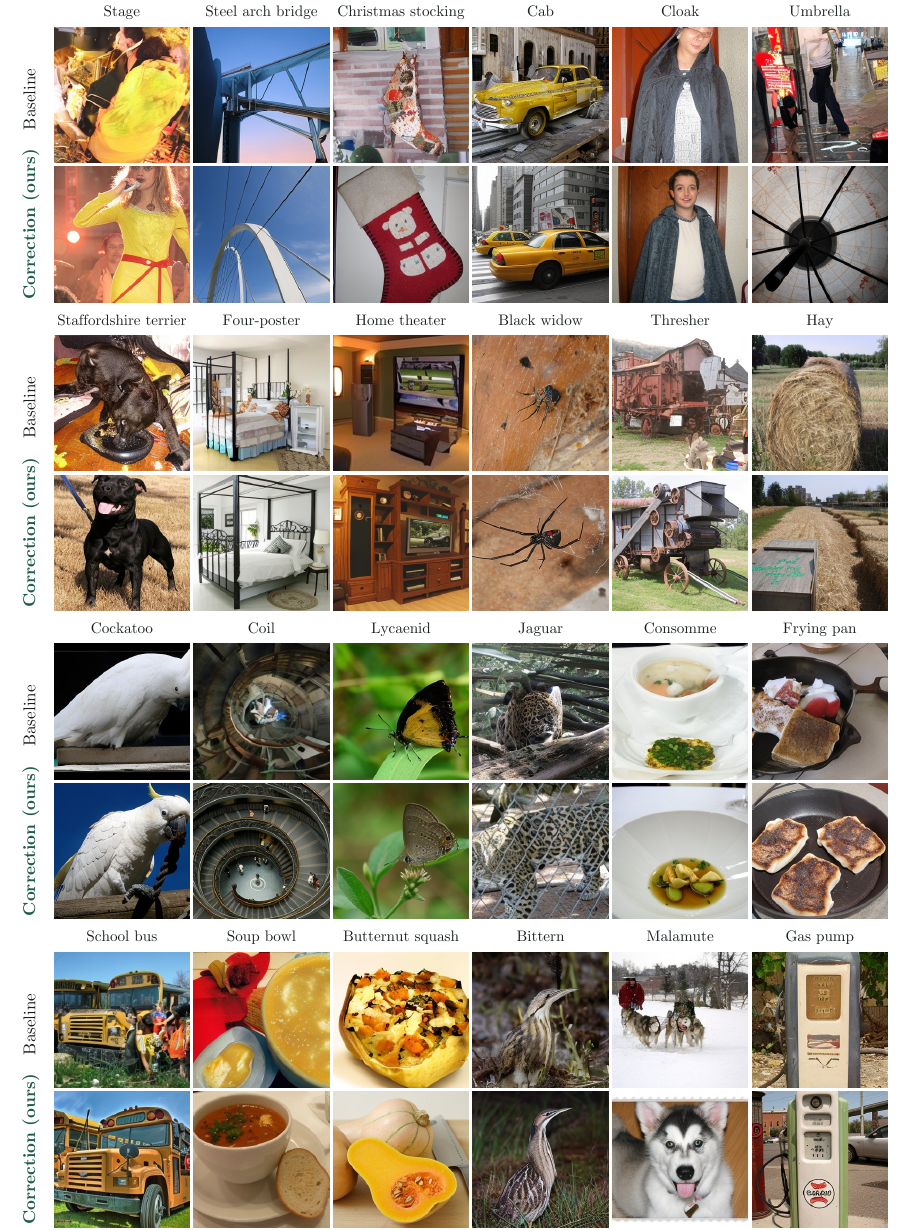}
    \caption{\textbf{\textit{Uncurated} samples: correction refines unguided SiT$_{\mathrm{REPA}}$ samples across random classes.} Within each pair of rows, baseline ($K=0$) and correction ($K=6$) share the same class and initial noise at approximately matched NFEs.}
    \label{fig:more-samples-random-repa}
\end{figure}

\clearpage
\subsection{JiT: Random-Class Generation}

\begin{figure}[!ht]
    \centering
    \includegraphics[width=\linewidth]{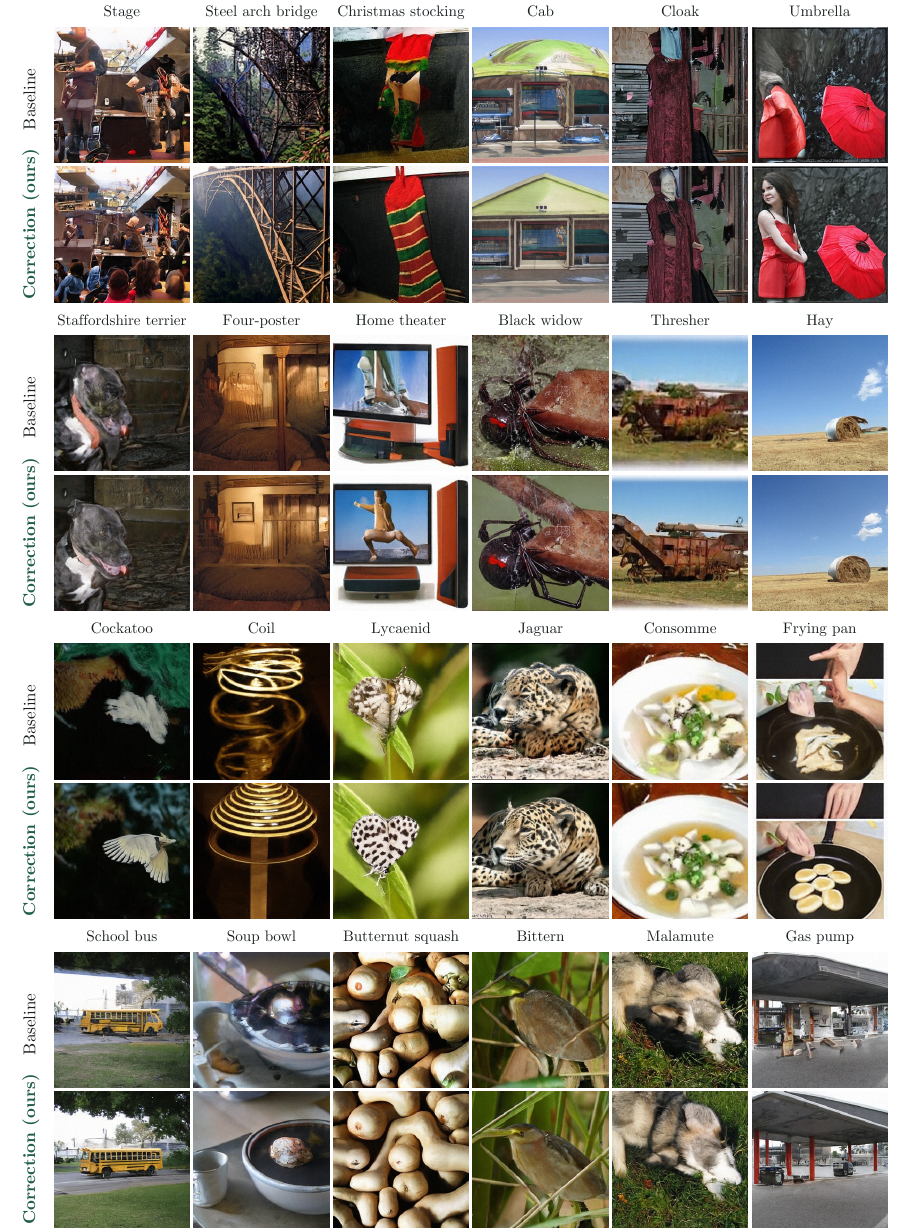}
    \caption{\textbf{\textit{Uncurated} samples: early correction changes structure as well as appearance in JiT.} Within each pair of rows, baseline ($K=0$) and correction ($K=6$) share the same class and initial noise at matched NFEs.}
    \label{fig:more-samples-random-jit}
\end{figure}

\clearpage

%% file: generated/diffusionbench_unconditional.tex
\subsection{Unconditional Generation from a Text-to-Image Model}
\label{app:t2i-unconditional}

\begin{figure}[!ht]
\centering
\includegraphics[width=\linewidth]{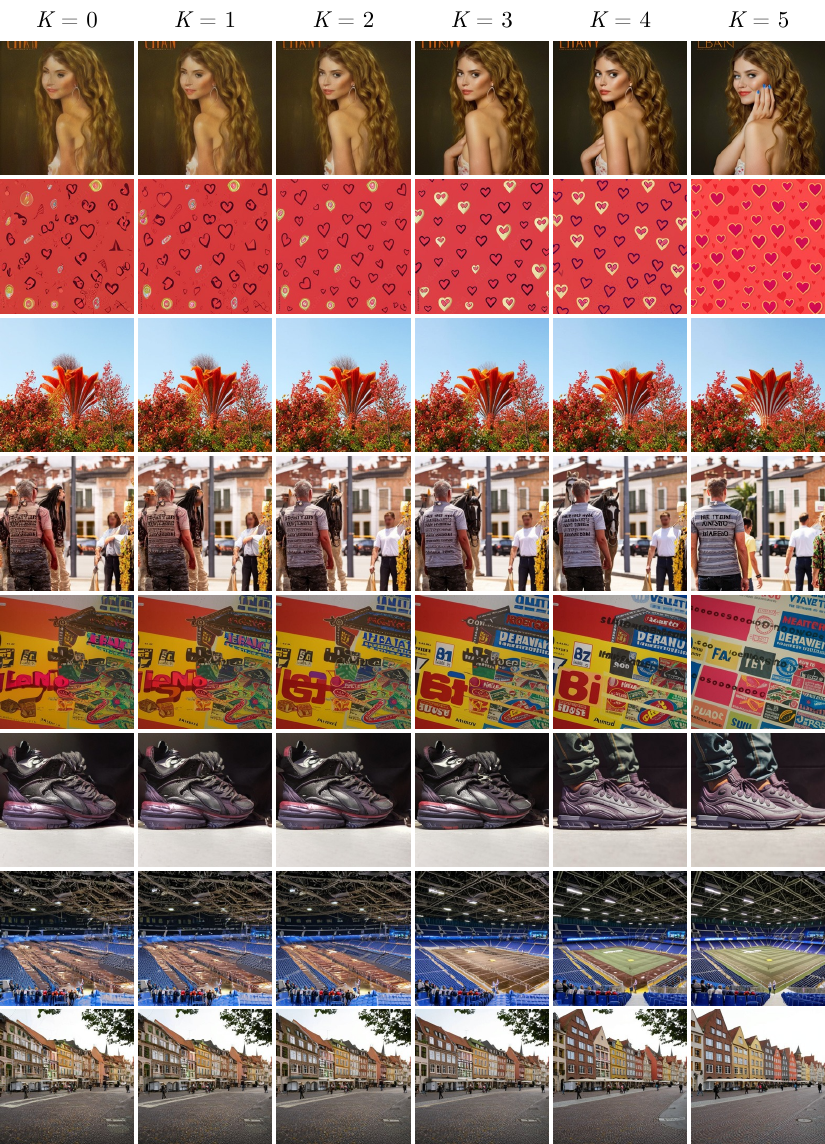}
\caption{\textbf{\textit{Uncurated} samples: unconditional generation supports iterative correction without text or CFG.} Each row shares its initial noise across $K=0,\ldots,5$. The checkpoint and denoising steps are fixed; larger $K$ adds correction compute. All seeds are shown without selection.}
\label{fig:t2i-unconditional-one}
\end{figure}

\clearpage

\begin{figure}[!ht]
\centering
\includegraphics[width=\linewidth]{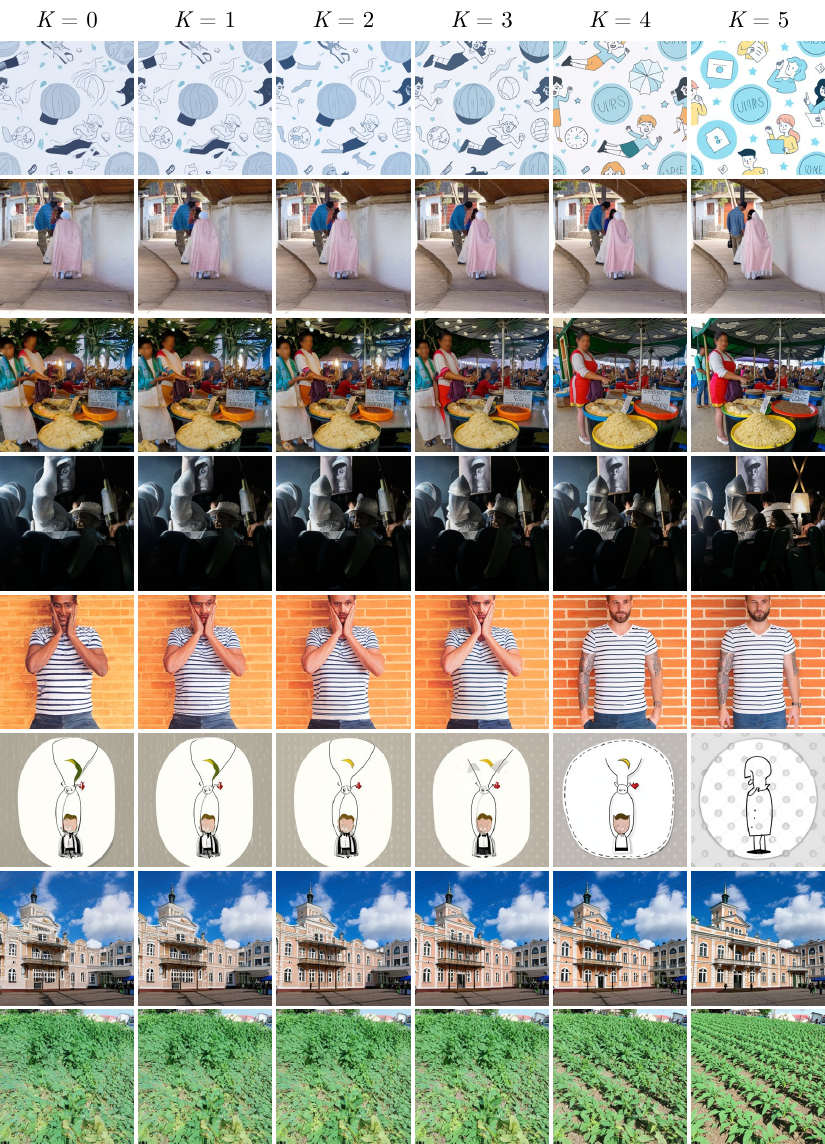}
\caption{\textbf{\textit{Uncurated} samples: unconditional generation supports iterative correction without text or CFG.} Each row shares its initial noise across $K=0,\ldots,5$. The checkpoint and denoising steps are fixed; larger $K$ adds correction compute. All seeds are shown without selection.}
\label{fig:t2i-unconditional-two}
\end{figure}

\clearpage

\begin{figure}[!ht]
\centering
\includegraphics[width=\linewidth]{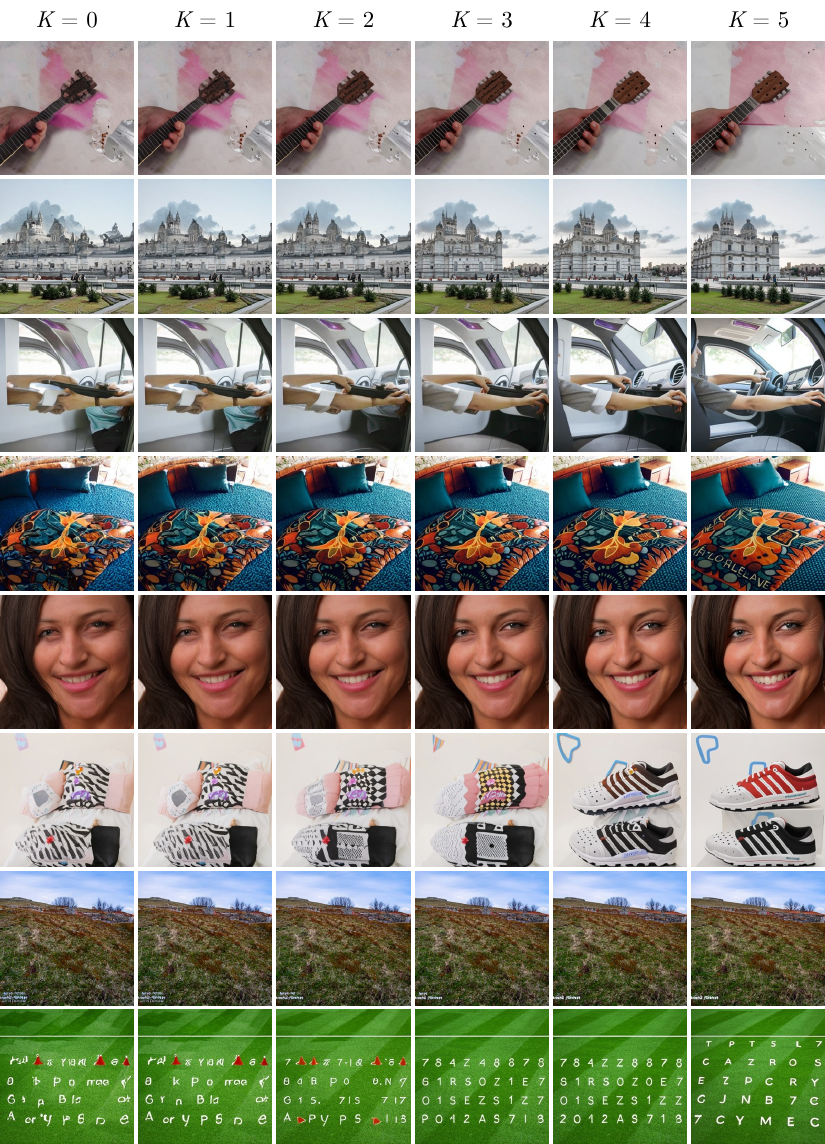}
\caption{\textbf{\textit{Uncurated} samples: unconditional generation supports iterative correction without text or CFG.} Each row shares its initial noise across $K=0,\ldots,5$. The checkpoint and denoising steps are fixed; larger $K$ adds correction compute. All seeds are shown without selection.}
\label{fig:t2i-unconditional-three}
\end{figure}

\clearpage

\begin{figure}[!ht]
\centering
\includegraphics[width=\linewidth]{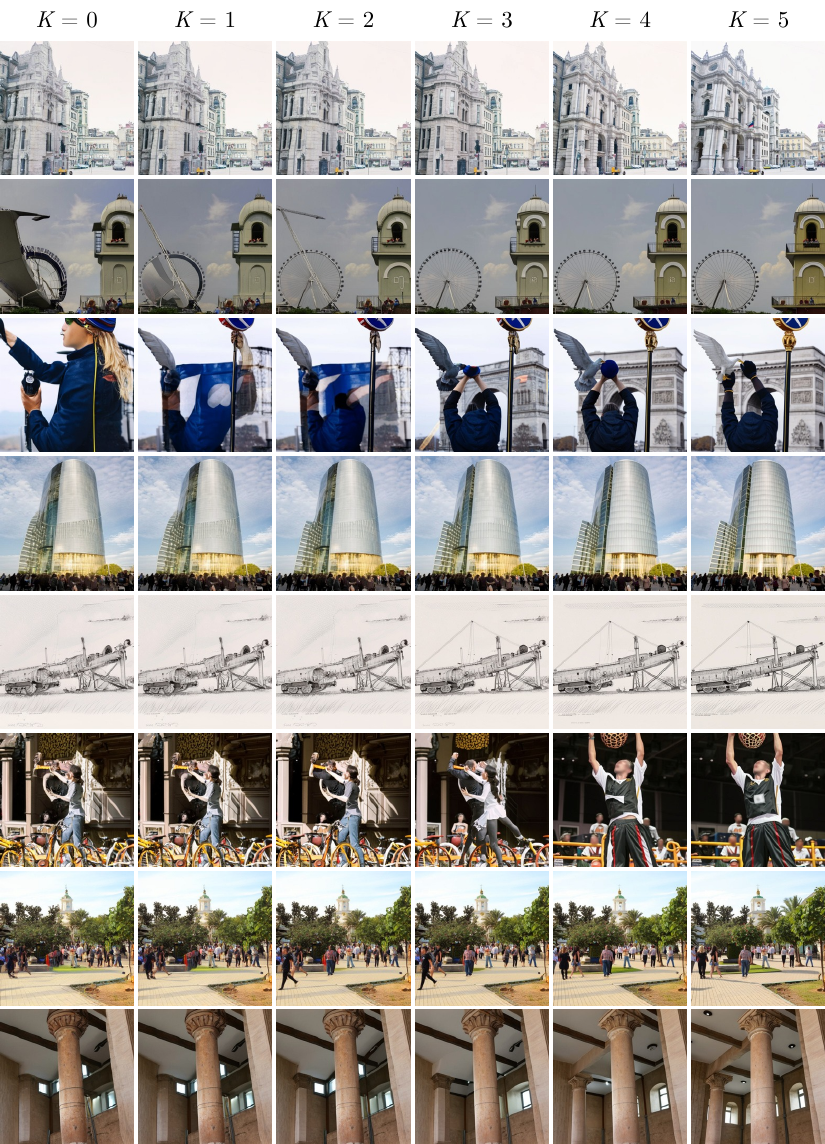}
\caption{\textbf{\textit{Uncurated} samples: unconditional generation supports iterative correction without text or CFG.} Each row shares its initial noise across $K=0,\ldots,5$. The checkpoint and denoising steps are fixed; larger $K$ adds correction compute. All seeds are shown without selection.}
\label{fig:t2i-unconditional-four}
\end{figure}